\documentclass[journal]{IEEEtran}
\usepackage{tikz}
\usetikzlibrary{arrows.meta, positioning}
\usepackage[utf8]{inputenc}
\usepackage{booktabs}
\usepackage{tikz}
\usepackage{xcolor}
\usepackage{tikz}
\usepackage{pgfplots}
\usetikzlibrary{pgfplots.groupplots}  
\pgfplotsset{compat=1.18}      
\usepackage{lipsum}  % For dummy text
\usepackage{amsmath,amssymb}
\usepackage{ragged2e}
\usepackage{tabularx}
\usepackage{adjustbox}

\renewcommand{\arraystretch}{1.1} % Adjust row spacing

\usepackage[
  colorlinks=false,
  linkbordercolor={0 1 0},   % RGB values for green
  citebordercolor={0 1 0},
  urlbordercolor={0 1 0}
]{hyperref}

\ifCLASSINFOpdf
\else
\fi
\begin{document}
\title{Reasoning Beyond Limits: Advances and Open Problems for LLMs}

\author{
\IEEEauthorblockN{Mohamed~Amine~Ferrag\IEEEauthorrefmark{1}\IEEEauthorrefmark{5}, Norbert Tihanyi\IEEEauthorrefmark{2}\IEEEauthorrefmark{3}, and  Merouane Debbah\IEEEauthorrefmark{4}}
 \\\IEEEauthorblockA{\IEEEauthorrefmark{1}United Arab Emirates University, UAE}
 \\\IEEEauthorblockA{\IEEEauthorrefmark{2}Technology Innovation Institute, UAE}
\\\IEEEauthorblockA{\IEEEauthorrefmark{3}Eötvös Loránd University, Hungary }
\\\IEEEauthorblockA{\IEEEauthorrefmark{4}Khalifa University of Science and Technology, UAE}
\\\IEEEauthorblockA{\IEEEauthorrefmark{5}Corresponding author: mohamed.ferrag@uaeu.ac.ae}
 }


\markboth{ }%
{Shell \MakeLowercase{\textit{et al.}}: Bare Demo of IEEEtran.cls for IEEE Journals}



\maketitle

\begin{abstract}
\textcolor{black}{Recent breakthroughs in generative reasoning have fundamentally reshaped how large language models (LLMs) address complex tasks, enabling them to dynamically retrieve, refine, and organize information into coherent, multi-step reasoning chains. Techniques such as inference-time scaling, reinforcement learning, supervised fine-tuning, and distillation have been effectively applied to state-of-the-art models, including DeepSeek-R1, OpenAI’s o1 and o3, GPT-4o, Qwen-32B, and various Llama variants, significantly enhancing their reasoning capabilities. In this paper, we present a comprehensive review of the top 27 LLMs released between 2023 and 2025, such as Mistral AI Small 3 24B, DeepSeek-R1, Search-o1, QwQ-32B, and Phi-4, and analyze their core innovations and performance improvements.}

\textcolor{black}{We also provide a detailed overview of recent advancements in multilingual large language models (MLLMs), emphasizing methods that improve cross-lingual reasoning and address the limitations of English-centric training. In parallel, we present a comprehensive review of progress in State Space Model (SSM)-based architectures, including models like Mamba, which demonstrate improved efficiency for long-context processing compared to Transformer-based approaches. Our analysis covers training strategies such as general optimization techniques, mixture-of-experts (MoE) configurations, retrieval-augmented generation (RAG), chain-of-thought prompting, self-improvement methods, and test-time compute scaling and distillation frameworks.}

\textcolor{black}{Finally, we identify key challenges for future research, including enabling multi-step reasoning without human supervision, improving robustness in chained task execution, balancing structured prompting with generative flexibility, and enhancing the integration of long-context retrieval and external tools.}

\end{abstract}


% \fi
\IEEEpeerreviewmaketitle
\section*{List of Abbreviations}
\begin{tabular}{ll}
APO   & Anchored Preference Optimization \\
AI    & Artificial Intelligence \\
CGPO  & Constrained Generative Policy Optimization \\
CoI   & Chain-of-Ideas \\
CoT   & Chain-of-Thought \\
DIA   & Dynamic Intelligence Assessment \\
DFT   & Supervised Fine-Tuning \\
DPO   & Direct Preference Optimization \\
EP    & Expert Parallelism \\
ESFT  & Expert-Specialized Fine-Tuning \\
GRPO  & Group Relative Policy Optimization \\
GQA   & Generalized Query Attention \\
\end{tabular}
\begin{tabular}{ll}
HLE   & Humanity's Last Exam \\
IPO   & Identity Preference Optimisation \\
KTO   & Kahneman-Tversky Optimisation \\
LAVR  & LLM-Aided Visual Reasoning \\
LLM   & Large Language Model \\
LRM   & Large Reasoning Models \\
LoRA  & Low-Rank Adaptation \\
M-ICL & Multimodal In-Context Learning \\
M-CoT & Multimodal Chain-of-Thought \\
MCTS  & Monte Carlo Tree Search \\
MLLM  & Multimodal Large Language Models \\
MLA   & Multihead Latent Attention \\
MMLU  & Measuring Massive Multitask Language Understanding \\
M-RoPE& Multimodal Rotary Position Embedding \\
MTP   & Multi-Token Prediction \\
NDR   & Naive Dynamic Resolution \\
NLP   & Natural Language Processing \\
ORM   & Observational Reward Model (ORM) \\
PEFT  & Parameter-Efficient Fine-Tuning \\
PPO   & Proximal Policy Optimization \\
PRM   & Process Reward Model \\
RAG   & Retrieval-Augmented Generation \\
RFT   & Reinforcement Fine-Tuning \\
RL    & Reinforcement Learning \\
RLHF  & Reinforcement Learning from Human Feedback \\
RLAIF & Reinforcement Learning from AI Feedback \\
RoPE  & Rotary Positional Embeddings \\
SKD   & Speculative Knowledge Distillation \\
SLM   & Small Language Model \\
SSM   & State Space Model \\
SoS   & Stream of Search \\
TPO   & Thought Preference Optimization \\
VLM   & Vision-Language Models \\
\end{tabular}

\begin{figure}
    \centering
    \includegraphics[width=0.5\textwidth]{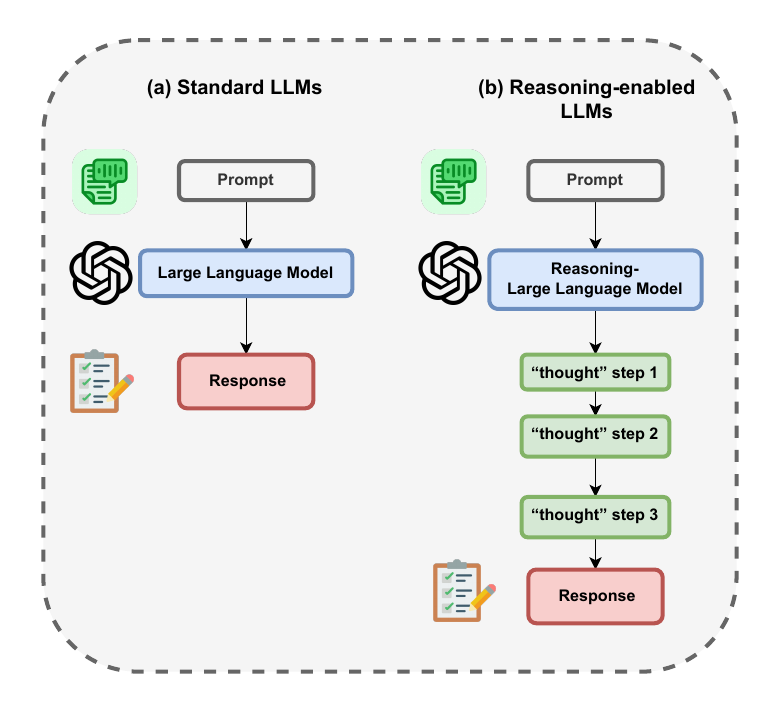}
    \caption{\textcolor{black}{Standard LLMs process an input prompt and immediately generate a response. Reasoning-enabled LLMs, by contrast, interpose a chain of internal “thought” steps. Each step builds on the last to work through the problem methodically before producing the final answer.}}
    \label{fig:figr1}
\end{figure}

\begin{figure*}
    \centering
    \includegraphics[width=0.85\textwidth]{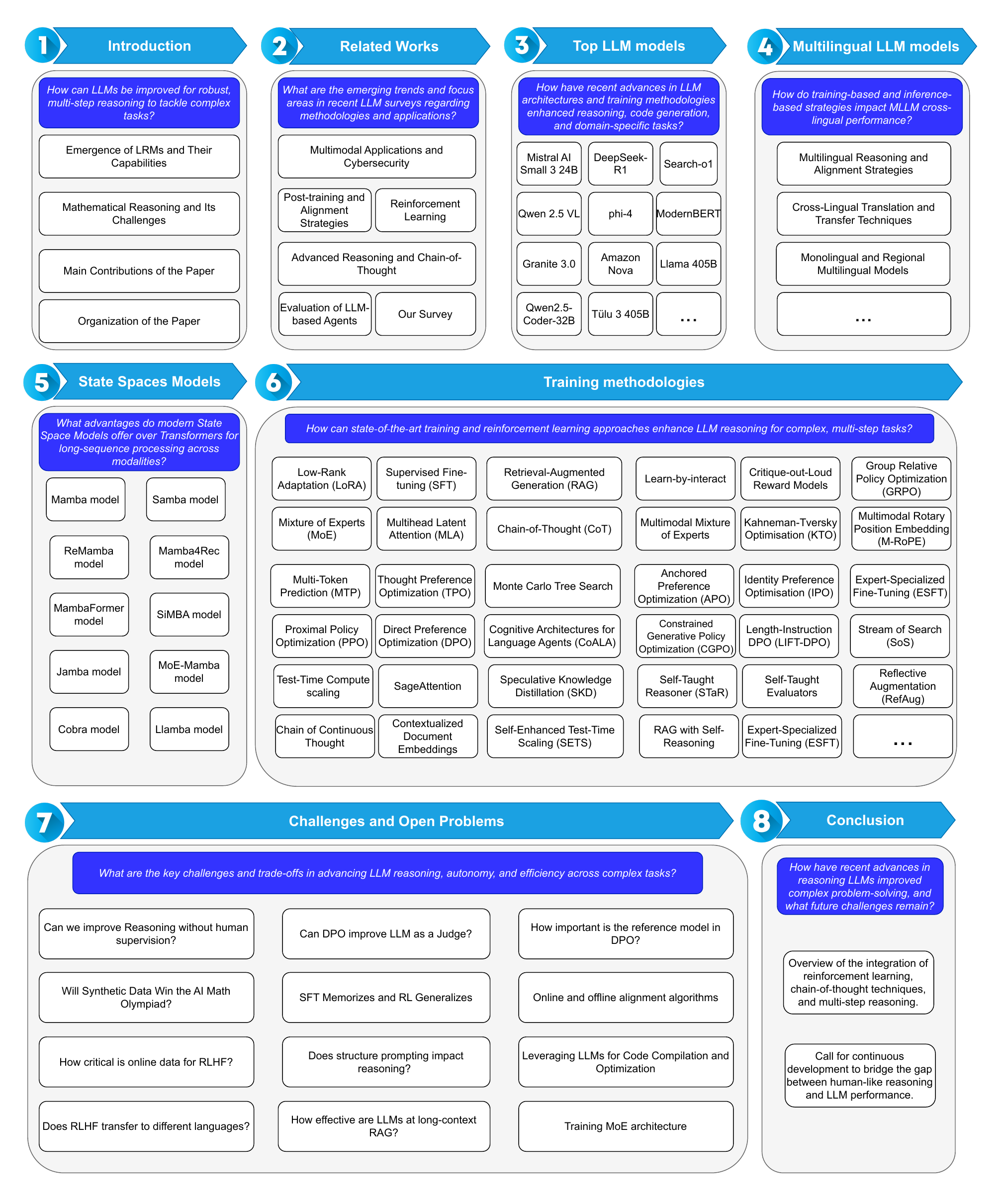}
    \caption{\textcolor{black}{Survey Structure.}}
    \label{fig:survey}
\end{figure*}

\section{Introduction}

The recent emergence of large reasoning models (LRMs) such as OpenAI’s o1 \cite{jaech2024openai}, QwenQwQ \cite{bai2025qwen2}, and DeepSeek-R1 \cite{guo2025deepseek} marks a significant milestone in AI research \cite{saoud2024artificial}. These models leverage large-scale reinforcement learning to achieve impressive long-sequence, stepwise reasoning capabilities. Their ability to break down complex problems into extended chains of thought has inspired numerous foundational efforts to replicate and broaden these reasoning patterns across various model architectures \cite{zhang2025lessons}. A key characteristic of these o1-like reasoning patterns is their promoting a slower, more deliberate thinking process. By decomposing intricate problems into stepwise reasoning chains, LRMs can enhance logical coherence and interpretability. However, this extended reasoning process also introduces challenges such as overthinking and the risk of propagating errors through knowledge gaps, which may disrupt the reasoning chain \cite{li2025search,muhammad2024pre,mohammadi2023ai,ajabani2025enhancing}.

Mathematical reasoning remains a formidable challenge in AI, with applications ranging from automated theorem proving to scientific discovery \cite{zhang2024llama,ahmad2025liu,shahid2025blockchain}. Recent advances, such as those seen in GPT-4, have improved performance in arithmetic and geometric tasks, yet complex problems, especially at Olympiad-level benchmarks, continue to test these models \cite{aops_aime,he2024olympiadbench}. Researchers have explored breaking down solutions into step-by-step reasoning paths using methods like Chain-of-Thought \cite{wei2022chain}, Self-Refine \cite{madaan2023self}, and Reflexion to enhance problem-solving \cite{shinn2023reflexion}, although these approaches sometimes face challenges related to feedback inefficiencies and local optima. Simultaneously, effective evaluation methods, including outcome and process reward models \cite{uesato2022solving,liu2025efficient}, struggle with inconsistent labeling and varied scoring standards. Innovations integrating techniques like Monte Carlo Tree Search with reinforcement learning-based evaluations offer promising avenues to improve the efficiency and reliability of mathematical reasoning in AI \cite{christiano2017deep,akhund2024comprehensive}.

Large language models (LLMs) have achieved impressive performance on complex reasoning tasks through prompting methods that induce chain-of-thought (CoT) reasoning. Still, recent findings reveal that by simply modifying the decoding procedure, specifically by inspecting alternative top-k tokens, LLMs can naturally elicit CoT reasoning without additional prompts or tuning \cite{wang2024chain,chung2024scaling,cobbe2021training,kousar2024mlhs}. Therefore, text often conceals meaning in its implications, and while language models (LMs) improve with reasoning, traditional methods like curated QA datasets (as used in Self-Taught Reasoner, or STaR) \cite{zelikman2022star,yang2024optimizing} limit their scalability and generality. Quiet-STaR \cite{zelikman2024quiet,shaikh2024new,shaikh2014towards} extends this approach by training LMs on a vast internet corpus using REINFORCE-based rewards, enabling them to learn diverse, implicit reasoning skills and "think before speaking" without relying solely on manually curated data.

Traditional LLMs lack inherent "Thinking then Responding" abilities because their training data often omits explicit thinking process annotations. \textcolor{black}{Fig. \ref{fig:figr1} highlights the key difference between “regular” and “reasoning” LLMs. A user’s question is passed straight into the model in the standard setup, emitting an answer in one step. By contrast, a reasoning-enabled LLM breaks the problem into a series of internal “thought” stages, each building on the previous one before arriving at its final response. This multi-step reasoning process makes the model’s decision path more transparent and helps it tackle complex, multi-stage tasks with greater reliability}. Recent efforts address this gap by generating synthetic thinking data using techniques like Monte Carlo Tree Search \cite{guan2025rstar}, GPT-4o demonstrations \cite{chen2024huatuogpt}, and model distillation \cite{muennighoff2025s1} or by employing reinforcement learning \cite{team2025kimi} to enable autonomous reasoning path learning. Despite strong benchmark performances, LLMs still face challenges in advanced reasoning tasks like mathematical problem-solving and code generation \cite{team2024gemini}. Techniques such as chain-of-thought prompting \cite{wei2022chain}, fine-tuning with CoT examples \cite{cobbe2021training}, and RLHF \cite{ouyang2022training} have enhanced reasoning capabilities, yet complex multi-step problems remain problematic. While decoding strategies like self-consistency and verifiers help, reward models especially Process Reward Models offering step-by-step feedback prove more effective but are hindered by costly, manual supervision \cite{lightman2023let}. To address this, OmegaPRM \cite{luo2024improve} introduces a divide-and-conquer Monte Carlo Tree Search algorithm that automates the collection of over 1.5 million high-quality process annotations, significantly boosting performance on benchmarks such as MATH500 and GSM8K.

This paper investigates strategies for enhancing the reasoning capabilities of LLMs by integrating methods such as inference-time scaling, pure reinforcement learning, and combined supervised fine-tuning with reinforcement learning, as well as distillation. We examine how models like DeepSeek-R1, OpenAI’s o1, GPT-4o, Qwen-32B, and Llama variants are refined to generate transparent, multi-step reasoning processes that improve performance on complex tasks ranging from advanced mathematics to coding challenges. The main contributions of this study are:

\begin{itemize}
    \item We comprehensively analyze the recent top 27 LLM models, particularly those released between 2023 and 2025 (e.g., Mistral AI Small 3 24B, DeepSeek-R1, Search-o1, QwQ-32B, phi-4, ...etc.).
    \item \textcolor{black}{We present a comprehensive overview of recent advancements in multilingual large language models (MLLMs).}
    \item \textcolor{black}{We present a comprehensive review of recent advancements in State Space Model (SSM)-based architectures.}
    \item We provide a comprehensive overview of training methodologies for LLMs, categorizing them into groups that span general training approaches, mixture-of-experts (MoE) and architectural innovations, retrieval-augmented generation (RAG), chain-of-thought and self-improvement techniques, test-time compute scaling and distillation methods, as well as reinforcement learning (RL) methods.
    \item We present a discussion of key challenges in advancing LLM capabilities. These include improving multi-step reasoning without human supervision, overcoming limitations in chained tasks, balancing structured prompts with flexibility, and enhancing long-context retrieval and external tool integration for optimized performance.
\end{itemize}

Fig. \ref{fig:survey} illustrates this survey's structure. Section \ref{sec:2} presents the related works. Section \ref{sec:3} analyzes the top 27 LLM models released from 2023 to 2025.\textcolor{black}{Section \ref{sec:r1} presents a comprehensive overview of recent advancements in multilingual large language models.}\textcolor{black}{Section \ref{sec:r2} presents a comprehensive overview of recent advancements in State Space Model (SSM)-based architectures.} Section \ref{sec:4} overviews training methodologies for LLMs, categorizing them into general approaches, MoE and architectural innovations, RAG, chain-of-thought and self-improvement, test-time scaling and distillation, and RL methods. Section \ref{sec:7} discusses key challenges in advancing LLM capabilities. Finally, Section \ref{sec:8} concludes the paper.

\begin{table*}[ht]
\centering
\scriptsize
\caption{Related works.}
\label{tab:relatedworks}
\resizebox{\textwidth}{!}{%
\begin{tabular}{p{2.2cm} p{0.8cm} p{2.2cm} p{3.5cm} p{4.2cm} p{1.4cm} p{1.6cm} p{1.6cm}}
\hline
\textbf{Reference} & \textbf{Year} & \textbf{Focus} & \textbf{Techniques/Methods} & \textbf{Key Contributions} & \textbf{LLM Models Analysis} & \textbf{Training Methodologies} & \textbf{Key Challenges} \\[5pt]
\hline
Yin et al. \cite{Yin_2024} 
  & 2024 
  & Multimodal LLMs 
  & Architecture details, training strategies, M-ICL, M-CoT, LAVR 
  & Overview of MLLMs with novel capabilities (e.g., narrative generation, OCR-free math reasoning) 
  & Partial 
  & Yes 
  & Partial \\[8pt]
\hline
Plaat et al. \cite{plaat2024reasoning} 
  & 2024 
  & LLM Reasoning 
  & In-context learning, chain-of-thought prompting 
  & Develops a taxonomy of reasoning strategies with insights into multi-step inference challenges 
  & No 
  & Partial 
  & Yes \\[8pt]
\hline
Wang et al. \cite{wang2024reinforcement} 
  & 2024 
  & RL Integration with LLMs 
  & RL-enhanced architectures, reward modeling (RLHF, RLAIF, DPO) 
  & Reviews integration challenges and reward modeling strategies for LLMs 
  & No 
  & Yes 
  & Yes \\[8pt]
\hline
Wang et al. \cite{wang2024enhancing} 
  & 2024 
  & RL for Code Generation 
  & RL in compiler optimization, resource allocation, and code generation frameworks 
  & Optimizes code generation and system efficiency via RL-based approaches 
  & No 
  & Yes 
  & Partial \\[8pt]
\hline  
Cao et al. \cite{cao2024survey} 
  & 2024 
  & LLMs in RL 
  & Roles: information processor, reward designer, decision-maker, generator 
  & Enhances RL via LLM integration; addresses multi-task learning, sample efficiency, and high-level task planning 
  & No 
  & Partial 
  & Partial \\[8pt]
\hline
Wang et al. \cite{wang2024comprehensive} 
  & 2024 
  & LLM Alignment Techniques 
  & RLHF, RLAIF, PPO, DPO 
  & Categorizes alignment methods to improve the reliability and safety of LLM outputs 
  & No 
  & Yes 
  & Partial \\[8pt]
\hline
Kumar et al. \cite{kumar2025llm} 
  & 2025 
  & Post-training for LLMs 
  & Fine-tuning, reinforcement learning, testing-time scaling 
  & Enhances reasoning, factual accuracy, and adaptability; tackles issues like catastrophic forgetting 
  & No 
  & Yes 
  & Yes \\[8pt]
\hline
Tie et al. \cite{tie2025survey} 
  & 2025 
  & Post-training Methodologies 
  & Fine-tuning, alignment, reasoning, efficiency, integration/adaptation 
  & Provides a structured taxonomy of post-training evolution 
  & No 
  & Yes 
  & Yes \\[8pt]
\hline
Yigit et al. \cite{yigit2025generative} 
  & 2025 
  & Cybersecurity \& LLMs 
  & Evaluation benchmarks for cybersecurity, agentic AI for proactive defense 
  & Secures critical infrastructures by addressing trust, privacy, and resilience 
  & No 
  & No 
  & Yes \\[8pt]
\hline
Chen et al. \cite{chen2025towards} 
  & 2025 
  & Long Chain-of-Thought Reasoning 
  & Novel taxonomy of reasoning paradigms, deep reasoning, reflective processes 
  & Differentiates Long CoT from traditional approaches and discusses integration with multimodal reasoning 
  & No 
  & Partial 
  & Yes \\[8pt]
\hline
Yehudai et al. \cite{yehudai2025surveyevaluationllmbasedagents} 
  & 2025
  & Evaluation of LLM-based Agents 
  & Systematic analysis of evaluation benchmarks and frameworks across agent capabilities, application-specific benchmarks, generalist agents, and evaluation methodologies 
  & First comprehensive survey on evaluation methodologies for LLM-based agents; highlights trends toward more realistic, challenging evaluations and identifies gaps in cost-efficiency, safety, and scalability 
  & No 
  & No 
  & Yes \\[8pt]
\hline
Ferrag et al. \cite{ferrag2025generative} 
  & 2025 
  & Cybersecurity and Generative AI 
  & LLM applications across hardware security, intrusion detection, software engineering, design verification, cyber threat intelligence, malware, and phishing detection
  & Provides a comprehensive review on integrating generative AI and LLMs into cybersecurity frameworks
  & Partial 
  & Partial  
  & No \\[8pt]
\hline

\textbf{Our Survey} 
  & 2025 
  & Comprehensive LLM Analysis \& Methodologies 
  & Analysis of 27 LLM models (2023--2025), categorization of training methods, and discussion on key challenges 
  & Fully covers all key aspects: Recent LLM models analysis, training methodologies (MoE, RAG, chain-of-thought, RL, etc.), and critical challenges in advancing LLM capabilities 
  & Yes 
  & Yes 
  & Yes \\[8pt]
\hline
\end{tabular}%
}
\end{table*}

\section{Related Works}\label{sec:2} 

Recent LLM surveys have increasingly sought to encapsulate the rapidly evolving landscape of methodologies and applications. For example, several studies from 2024 have focused on specialized domains such as multimodal LLMs, reasoning strategies, and reinforcement learning integrations, while surveys from 2025 have extended these analyses to include cybersecurity applications and long chain-of-thought reasoning, as presented in Table. \ref{tab:relatedworks}. These surveys systematically address various aspects, including architectural innovations, training strategies (e.g., fine-tuning, reinforcement learning, and in-context learning), and evaluation benchmarks. 

\subsection{Multimodal Applications and Cybersecurity}
Yin et al. \cite{Yin_2024} provide an in-depth overview of the rapidly evolving field of multimodal large language models (MLLMs), exemplified by cutting-edge systems like GPT-4V. The authors lay a solid foundation by describing the core formulation of MLLMs, including their architectures, training strategies, data usage, and evaluation metrics. They emphasize the unique capabilities of these models such as generating narratives from images and performing OCR-free mathematical reasoning which are seldom observed in traditional multimodal approaches and hint at the potential trajectory toward artificial general intelligence. The review further explores avenues for enhancing MLLMs by extending support to additional granularities, modalities, languages, and application scenarios. It also tackles critical issues like multimodal hallucination and introduces innovative techniques, including Multimodal In-Context Learning (M-ICL), Multimodal Chain-of-Thought (M-CoT), and LLM-Aided Visual Reasoning (LAVR), to improve the models' reasoning capabilities.

The review by Yigit et al. \cite{yigit2025generative} provides a comprehensive examination of how generative AI and LLMs can be harnessed to bolster the cybersecurity of critical national infrastructures such as energy grids, water systems, transportation networks, and communication frameworks. It begins by assessing the reliability of these infrastructures using established evaluation benchmarks tailored for cybersecurity applications of LLMs, and it then delves into fundamental issues like trust, privacy, resilience, and securability. Notably, the paper highlights the emerging role of agentic AI in developing proactive defense mechanisms, thereby offering innovative pathways for detecting and mitigating cyber threats. By synthesizing current challenges with advanced AI methodologies, the authors propose a strategic roadmap that reinforces existing protection measures and sets the stage for future research in safeguarding critical infrastructures.

The survey by Ferrag et al. \cite{ferrag2025generative} offers a comprehensive review of integrating generative AI and large language models (LLMs) within cybersecurity frameworks. It systematically examines how LLMs, including state-of-the-art models such as GPT-4, GPT-3.5, Mixtral-8x7B, BERT, Falcon2, and LLaMA, can be applied across various cybersecurity domains from hardware design security and intrusion detection to software engineering and malware as well as phishing detection. The study not only highlights the evolution and current state of LLMs but also critically assesses their vulnerabilities, including prompt injection, insecure output handling, data poisoning, DDoS attacks, and adversarial instructions. Furthermore, it evaluates the performance of 42 distinct LLM models regarding cybersecurity knowledge and hardware security, offering valuable insights into their strengths and limitations.

\subsection{Reinforcement Learning}
Cao et al. \cite{cao2024survey} comprehensively reviews how LLMs can enhance reinforcement learning (RL) by integrating extensive pre-trained knowledge into the learning process. It systematically categorizes the contributions of LLMs into four distinct roles: information processor, reward designer, decision-maker, and generator, each designed to address specific challenges in RL, such as multi-task learning, sample efficiency, and high-level task planning. By framing its analysis within the classical agent-environment interaction paradigm, the survey not only highlights the methodologies underpinning each role but also contrasts these innovations with conventional RL approaches. Furthermore, the paper presents a comparative analysis that elucidates potential applications in complex domains like robotics, autonomous driving, and energy systems, and it outlines future research directions aimed at accelerating progress in the field.

Wang et al. \cite{wang2024reinforcement} provides a comprehensive overview of how reinforcement learning (RL) can be integrated with LLMs to significantly enhance their performance, as illustrated by models like DeepSeek-R1. The authors detail the complexities involved in this integration, including developing sophisticated algorithms, reward modeling strategies, and optimization techniques critical for practical RL implementations. The survey systematically reviews the fundamentals of RL, introduces leading RL-enhanced LLM architectures, and critically examines two primary reward model-based methods: reinforcement learning from human feedback (RLHF) and reinforcement learning from AI feedback (RLAIF), as well as direct preference optimization (DPO), which leverages human preference data directly. By identifying the current challenges and gaps in existing methodologies, the paper lays out a strategic agenda for future research, aiming to foster deeper understanding and further advancements in the field.

Wang et al. \cite{wang2024enhancing} examine the integration of reinforcement learning (RL) techniques with LLMs for code generation and optimization. It highlights how RL has become pivotal in enhancing compiler optimization, where sophisticated algorithms are employed to improve efficiency and resource utilization, particularly in tasks like register allocation and overall system optimization. In addition, the paper discusses RL's role in advancing resource allocation strategies and strengthening the development of robust code generation frameworks and tools. By providing a detailed review of these applications, the survey is a comprehensive resource for researchers and practitioners seeking to leverage RL to refine and innovate code generation methodologies, ultimately addressing current challenges and potential future directions in the field.

\subsection{Post-training and Alignment Strategies}
Kumar et al. \cite{kumar2025llm} delves into post-training techniques for LLMs, emphasizing that while pre-training establishes a robust linguistic foundation, it is subsequent refinement, through methods such as fine-tuning, reinforcement learning, and testing-time scaling, that truly enhances the models' reasoning abilities, factual accuracy, and overall adaptability. The work systematically examines these post-training strategies, focusing on their potential to address critical challenges such as catastrophic forgetting, reward hacking, and the trade-offs encountered during inference. Furthermore, it highlights emerging avenues in model alignment and scalable adaptation, underscoring the importance of these techniques in ensuring that LLMs understand language and perform effectively across a diverse range of real-world tasks.

The survey by Wang et al. \cite{wang2024comprehensive} offers a systematic overview of various alignment techniques developed to enhance the reliability of LLMs. It emphasizes that despite LLMs achieving remarkable capabilities through massive pretraining, the variability in data quality can still lead to suboptimal or undesired outputs. To mitigate this, the paper categorizes and examines a range of methods, including reinforcement learning from human feedback (RLHF), reinforcement learning from AI feedback (RLAIF), proximal policy optimization (PPO), and direct preference optimization (DPO), detailing how each approach contributes to aligning model outputs with human expectations. The survey fills a critical gap in the literature by organizing these techniques into distinct topics. It provides a comprehensive framework that not only aids in understanding current alignment strategies but also guides future research toward improving the performance and safety of LLMs.

The survey by Tie et al. \cite{tie2025survey} presents a thorough synthesis of post-training methodologies to overcome the inherent limitations of pre-trained large language models. It critically addresses challenges such as constrained reasoning abilities, ethical ambiguities, and performance issues in specialized domains, underscoring the need for advanced post-training strategies. The authors systematically categorize the evolution of post-training language models into five core paradigms: fine-tuning to enhance task-specific accuracy, alignment to better meet human expectations, reasoning to facilitate multi-step inference despite reward design challenges, efficiency to optimize resource use amid growing model complexity, and integration and adaptation to extend capabilities across diverse modalities. By mapping progress from early alignment techniques exemplified by ChatGPT to the innovative reasoning approaches of models like DeepSeek-R1, the survey establishes a structured taxonomy that not only consolidates recent advancements but also charts a strategic path for future research toward developing LLMs with improved precision, ethical robustness, and domain adaptability.

\subsection{Advanced Reasoning and Chain-of-Thought}
The survey by Plaat et al. \cite{plaat2024reasoning} examines the evolving reasoning landscape with LLMs as they scale to billions of parameters, enabling advanced in-context learning and few-shot capabilities. It highlights the breakthrough performance of these models on traditional "System 1" tasks such as translation, summarization, and question-answering. It extends the discussion to "System 2" reasoning enabled by chain-of-thought prompting. The authors present a comprehensive taxonomy that categorizes various strategies for generating, evaluating, and controlling multi-step reasoning processes. In doing so, the paper offers an in-depth review of current methodologies and identifies key open challenges, including integrating sequential decision processes and reinforcement learning. Furthermore, it explores the potential for self-improvement and metacognitive abilities through prompt-based techniques while noting that LLMs' ultimate shift toward autonomous reasoning remains a promising avenue for future research.

Chen et al. \cite{chen2025towards} provide a unified perspective on the emerging paradigm of long chain-of-thought (Long CoT) reasoning within LLMs. Building on the impressive capabilities demonstrated by models such as OpenAI-O1 and DeepSeek-R1 in complex domains like mathematics and coding, the authors distinguish Long CoT from traditional short chain-of-thought approaches by introducing a novel taxonomy of reasoning paradigms. The paper highlights key characteristics of Long CoT deep reasoning, extensive exploration, and reflective processes that enable models to solve more intricate tasks and yield more coherent outcomes. It also examines overthinking and test-time scaling phenomena, offering insights into how these behaviors manifest in practice. By identifying significant research gaps and proposing future directions, including integrating multi-modal reasoning and efficiency improvements, the survey sets a strategic roadmap for advancing logical reasoning capabilities in artificial intelligence.

\subsection{Evaluation of LLM-based Agents}
Yehudai et al. \cite{yehudai2025surveyevaluationllmbasedagents} provide a comprehensive review of evaluation methodologies for LLM-based agents, marking a significant paradigm shift in autonomous AI systems. These agents, capable of planning, reasoning, tool usage, and maintaining memory in dynamic environments, require robust evaluation frameworks to gauge their performance accurately. The paper systematically dissects evaluation benchmarks and methodologies across four critical dimensions: assessing fundamental capabilities such as planning, self-reflection, tool utilization, and memory; domain-specific benchmarks for areas like web interactions, software engineering, scientific inquiry, and conversational tasks; evaluation standards for generalist agents; and overarching frameworks that integrate these diverse facets. Notably, the survey highlights emerging trends toward more realistic, challenging evaluations with continuously updated benchmarks while identifying key gaps, especially regarding cost-efficiency, safety, robustness, and scalability that must be addressed. This work thus not only maps the rapidly evolving landscape of agent evaluation and sets forth a clear research agenda for advancing future methodologies.

\subsection{Our Survey}

Our study makes a series of novel contributions compared to related works as presented in Table. \ref{tab:relatedworks}. We analyze the top 27 LLM models released between 2023 and 2025, including models such as Mistral AI Small 3 24B, DeepSeek-R1, Search-o1, QwQ-32B, and phi-4, providing detailed insights into their architectures, performance, and innovations. In addition, we offer a thorough overview of training methodologies by categorizing approaches into general training techniques, mixture-of-experts (MoE) innovations, retrieval-augmented generation (RAG), chain-of-thought and self-improvement methods, as well as test-time compute scaling, distillation, and reinforcement learning (RL) strategies.

Furthermore, our work extends beyond model analysis and training techniques by providing the key challenges in advancing LLM capabilities, including improving multi-step reasoning without human supervision, overcoming limitations in chained tasks, balancing structured prompts with flexibility, and enhancing long-context retrieval alongside external tool integration. These integrated contributions uniquely position our work as a holistic reference for current trends and future directions in LLM research.

%==================================================
%                TABLE 1
%==================================================
\begin{table*}[htbp]
\centering
\scriptsize
\caption{Summary of Top LLM Models (Part I)}
\label{tab:llm_models_part1}
\begin{adjustbox}{max width=\textwidth}
% Define first three columns with fixed widths, and the last three as flexible X columns
\begin{tabularx}{\textwidth}{@{}%
>{\raggedright\arraybackslash}p{3.2cm}%  <-- Model Name
>{\centering\arraybackslash}p{1.0cm}%   <-- Year
>{\raggedright\arraybackslash}p{2.8cm}%<-- Parameters
>{\raggedright\arraybackslash}X%        <-- Key Metrics
>{\raggedright\arraybackslash}X%        <-- Innovations
>{\raggedright\arraybackslash}X@{}}%     <-- Domain/Notes
\toprule
\textbf{Model Name} & \textbf{Year} & \textbf{Parameters} & \textbf{Key Metrics/Benchmarks} & \textbf{Innovations/Techniques} & \textbf{Observations} \\
\midrule
UI-TARS \cite{qin2025ui} & 2025 & 2B, 7B, 72B 
& OSWorld: 24.6 (50-step), 22.7 (15-step); AndroidWorld: 46.6 (vs. GPT-4’s 34.5)
& Native GUI agent using screenshot inputs; end-to-end design; System-2 reasoning; unified action modeling; iterative training
& GUI agent outperforming comparable models (e.g., Claude) \\
\midrule
Tülu 3 405B \cite{lambert2025tulu3pushingfrontiers} & 2025 & 405B
& Surpasses Llama 3.1-Instruct, Qwen 2.5, Mistral, GPT-4o-mini, Claude 3.5-Haiku
& Transparent, reproducible post-training using SFT, DPO, and RLVR
& Built on Llama 3.1 base; open-source pipeline \\
\midrule
Mistral AI Small 3 24B \cite{mistral_small_3} & 2025 & 24B
& 81\% accuracy on MMLU; 150 tokens/s; 3$\times$ faster on same hardware
& Fewer layers for lower latency; optimized for instruction-following; no RL or synthetic data
& Efficient LLM for generative AI tasks \\
\midrule
DeepSeek-R1 \cite{deepseekai2025} & 2025 & 1.5B--70B 
& FP8 quantization cuts memory by 75\%; MTP yields 2--3$\times$ faster token generation
& Uses Mixture of Experts (MoE), Multihead Latent Attention (MLA), and Multi-Token Prediction (MTP)
& Matches OpenAI’s o1-1217 in reasoning tasks \\
\midrule
Qwen 2.5 VL \cite{qwen2.5_vl} & 2024 & 3B, 7B, 72B
& 72B version matches GPT-4o and Claude 3.5-Sonnet on multimodal benchmarks
& Naive Dynamic Resolution (NDR) for adaptive visual tokenization; M-RoPE; unified image--video processing
& Advanced vision--language integration \\
\midrule
HuatuoGPT-o1 \cite{chen2024huatuogpt} & 2024 & 70B
& $\sim$80\% improvement on MedQA and PubMedQA; complex reasoning chains $>$700 tokens
& Two-stage training: 40K verifiable problems then PPO-based RL with a robust verifier
& Focused on enhancing medical reasoning \\
\midrule
Search-o1 \cite{li2025search} & 2025 & --
& Outperforms base models by 4.7\% and traditional RAG systems by up to 29.6\% on multi-hop questions
& Retrieval-augmented generation with a “Reason-in-Documents” module
& Enhances long stepwise reasoning in LRMs \\
\midrule
InternLM2 \cite{cai2024internlm2} & 2024 & 8B (up to 128k context)
& 83.0\% on MATH-500; 20.0\% on AIME2024; trained on 4T tokens
& COOL RLHF; multi-phase pre-training (4k $\rightarrow$ 32k tokens) for long-context learning
& Open-source LLM with 75\% cost savings \\
\midrule
MiniCPM-o 2.6 \cite{yao2024minicpm} & 2024 & 8B
& Visual: 70.2 on OpenCompass; encodes 1.8M pixels in 640 tokens
& On-device multimodal processing; real-time speech and OCR
& Designed for on-device applications with flexible voice customization \\
\midrule
KaLM-Embedding \cite{hu2025kalm} & 2025 & $<$1B
& Average score: 64.53 on MTEB benchmark
& Persona-based synthetic data generation; ranking consistency filtering; Matryoshka Representation Learning
& Multilingual embedding model with flexible dimensions \\
\midrule
rStar-Math \cite{guan2025rstar} & 2025 & 7B
& 90.0\% on MATH; 53.3\% on AIME
& Monte Carlo Tree Search with code-augmented chain-of-thought; self-evolution; improved PRM training
& Specialized for advanced mathematical reasoning \\
\midrule
phi-4 \cite{abdin2024phi} & 2024 & 14B
& Trained on 9.8T tokens; 16K token context; comparable to 70B models
& High-quality synthetic data via multi-agent self-revision; SFT and DPO for post-training
& STEM-focused QA model (Microsoft) \\
\midrule
DeepSeek-V3 \cite{liu2024deepseek} & 2024 & 671B total (37B active)
& Throughput: 60 tokens/s; 2.788M H800 GPU hrs (\$5.5M)
& MoE, MLA, DeepSeekMoE; auxiliary-loss-free load balancing; multi-token prediction; FP8 training
& Massive LLM for advanced reasoning \\
\midrule
DeepSeekMath 7B \cite{shao2024deepseekmath} & 2024 & 7B
& 51.7\% on MATH (60.9\% with self-consistency)
& Group Relative Policy Optimization (GRPO); tailored data selection pipeline
& Specialized for mathematical reasoning \\
\midrule
Qwen 2.5 \cite{yang2024qwen2} & 2024 & 0.5B--72B
& 72B-Instruct outperforms many larger models on multiple benchmarks
& Enhanced pre-training (7T $\rightarrow$ 18T tokens), extensive SFT, offline RL (DPO) and online RL (GRPO); modern features (GQA, SwiGLU, RoPE)
& Comprehensive general-purpose LLM series \\
\bottomrule
\end{tabularx}
\end{adjustbox}
\end{table*}

%==================================================
%                TABLE 2
%==================================================
\begin{table*}[htbp]
\centering
\scriptsize
\caption{Summary of Top LLM Models (Part II)}
\label{tab:llm_models_part2}
\begin{adjustbox}{max width=\textwidth}
% Same column definitions to ensure consistent layout
\begin{tabularx}{\textwidth}{@{}%
>{\raggedright\arraybackslash}p{3.2cm}%  <-- Model Name
>{\centering\arraybackslash}p{1.0cm}%   <-- Year
>{\raggedright\arraybackslash}p{2.8cm}%<-- Parameters
>{\raggedright\arraybackslash}X%        <-- Key Metrics
>{\raggedright\arraybackslash}X%        <-- Innovations
>{\raggedright\arraybackslash}X@{}}%     <-- Domain/Notes
\toprule
\textbf{Model Name} & \textbf{Year} & \textbf{Parameters} & \textbf{Key Metrics/Benchmarks} & \textbf{Innovations/Techniques} & \textbf{Observations} \\
\midrule
ModernBERT \cite{warner2024smarter} & 2024 & 139M (base), 395M (large)
& Extended context: 8,192 tokens; trained on 2T tokens; 2--4$\times$ faster than peers
& Flash Attention 2; RoPE embeddings; alternating attention mechanism
& Next-generation encoder for classification and retrieval \\
\midrule
Gemini 2.0 Flash \cite{googleblog2024gemini} & 2024 & --
& 51.8\% on SWE-bench Verified
& Multimodal Live API for real-time audio/video streaming; native tool use; supports 8 voices and 109 languages
& Developed by Google DeepMind for enhanced developer applications \\
\midrule
Llama 3.3 70B-Instruct \cite{meta_llama2023llama33} & 2023 & 70B
& $\sim$4\% overall improvement; +9\% on MATH; +8\% on HumanEval
& Improved post-training techniques without altering the architecture
& Matches performance of high-end models (e.g., GPT-4o, Anthropic) \\
\midrule
Llama 3.1 \cite{llama31} & 2024 & 8B, 70B, 405B
& 405B variant matches/exceeds GPT-4o benchmarks; supports 128k token context
& Trained on $>$15T tokens with 25M human/synthetic samples; optimized quantization (FP8, AWQ, GPTQ)
& Offered in Instruct and Base versions; commercial-friendly \\
\midrule
Amazon Nova \cite{aws2023amazonNova} & 2023 & Multiple variants
& Context up to 300k tokens; $>$200 languages; Pricing: Micro (\$0.035 input, \$0.14 output), Lite (\$0.06/\$0.24), Pro (\$0.80/\$3.20)
& Suite includes text-only, multimodal, and specialized models (Canvas for image, Reel for video); watermarking
& Deployed on AWS (US); integrates with Amazon Bedrock \\
\midrule
QwQ-32B-Preview \cite{qwenlm2023qwq32b} & 2023 & 32.5B; 32,768 token context
& GPQA: 65.2\%; AIME: 50.0\%; MATH-500: 90.6\%; LiveCodeBench: 50.0\%
& Open-weight reasoning model based on Qwen2.5; experimental with noted limitations
& Released under Apache 2.0 on Hugging Face \\
\midrule
Qwen2.5-Coder-32B \cite{qwenlm2023qwen25} & 2023 & 32B; 128K token context
& HumanEval: 92.7 (vs. 92.1); EvalPlus: 86.3 (vs. 85.9); Aider (code repair): 73.7
& Code-focused architecture with extended system prompts; multilingual support ($>$40 languages)
& Optimized for diverse coding benchmarks \\
\midrule
Hunyuan-Large \cite{sun2024hunyuan} & 2024 & 389B total (52B active); up to 256K tokens
& Uses 1.5T synthetic tokens (of 7T total)
& Largest open-source MoE with 16 experts; mixed expert routing; KV cache compression; expert-specific learning rates
& Outperforms Llama 3.1-70B; competitive with 405B models \\
\midrule
Granite 3.0 \cite{ibm2023granite} & 2023 & 8B \& 2B Dense; 3B \& 1B MoE
& OpenLLM Leaderboard: 37.6 (vs. Llama 3.1 8B: 37.3)
& Speculator Head accelerates token processing up to 220\%; multilingual, coding, function-calling features
& Lightweight foundation for enterprise/on-device; Apache 2.0 licensed \\
\midrule
Pyramid Flow SD3 \cite{jin2024pyramidal} & 2024 & 2B (DiT model)
& Videos up to 10 sec at 768p, 24 FPS; Training: 20.7k A100 GPU hrs
& Unified pyramidal flow matching; autoregressive temporal pyramid for spatiotemporal compression
& First robust open-source text-to-video and image-to-video model \\
\midrule
Molmo model \cite{deitke2024molmo} & 2024 & 7B MoE (A1B), dual 7B, 72B
& Molmo 72B outperforms Llama 3.2 90B and Pixtral 12B; matches proprietary models (GPT-4o, Gemini 1.5 Pro, Claude Sonnet 3.5)
& Built entirely from scratch using the PixMo dataset; uses OpenAI CLIP in the 72B variant
& Open-weight vision-language model family \\
\midrule
Moshi \cite{defossez2024moshi} & 2024 & 7B
& Theoretical latency: 160\,ms (200\,ms on L4 GPU); 12.5\,Hz at 1.1\,kbps
& Full-duplex spoken dialogue via a Temporal Transformer; “Inner Monologue” for time-aligned text prefix to audio tokens
& Speech--text model for real-time, on-device dialogue (cc-by-4.0) \\
\bottomrule
\end{tabularx}
\end{adjustbox}
\end{table*}

\section{Top LLM models}\label{sec:3}
Recent advances in LLMs during 2023–2025 have demonstrated remarkable progress in architecture and training methodologies. For example, UI-TARS \cite{qin2025ui} and Tülu 3 \cite{lambert2025tulu3pushingfrontiers} exemplify novel approaches in multimodal processing and transparent post-training pipelines, respectively, that have enabled significant improvements in reasoning, code generation, and domain-specific tasks. These developments leverage innovations such as reinforcement learning with verifiable rewards, retrieval-augmented generation, and large-scale fine-tuning, addressing critical scalability, efficiency, and task adaptability challenges. The concentrated efforts during this period underscore a broader trend toward enhancing performance through high-quality synthetic data and advanced optimization techniques.

In parallel, models such as HuatuoGPT-o1 \cite{chen2024huatuogpt} and Search-o1 \cite{li2025search} highlight a focused commitment to refining reasoning capabilities in specialized domains, including medicine and long-step problem-solving. In addition, breakthroughs in vision language integration and context management have been achieved by systems such as Qwen2.5 VL \cite{qwen2.5_vl} and InternLM2 \cite{cai2024internlm2}, further broadening the scope of applications for LLM. These contributions, emerging in a concentrated timeframe between 2024 and 2025, reinforce the rapid pace of innovation in the field and provide a robust foundation for subsequent research and development in artificial intelligence.

This section comprehensively analyzes recent advancements in LLMs, particularly those released between 2023 and 2025. By systematically examining these state-of-the-art models, we aim to elucidate the underlying innovations in architecture, training methodologies, and application domains that collectively contribute to their enhanced performance. The models discussed herein represent breakthroughs in multimodal integration, reasoning, and code generation and underscore the trend toward increased transparency and reproducibility in AI research. This detailed survey serves as a foundation for understanding the current landscape and identifying potential avenues for future investigation in artificial intelligence. Tables \ref{tab:llm_models_part1}, \ref{tab:llm_models_part2} present the recent top 27 LLM models, particularly those released between 2023 and 2025.

\subsection{UI-TARS model}

UI-TARS \cite{qin2025ui} is a novel native GUI agent model that exclusively utilizes screenshot inputs to execute human-like interactions such as mouse and keyboard operations. Unlike conventional frameworks that rely on heavily wrapped commercial models and meticulously designed prompts, UI-TARS has been developed as an end-to-end system that surpasses these sophisticated performance approaches. Experimental evaluations across more than ten GUI agent benchmarks assessing perception, grounding, and task execution demonstrate its state-of-the-art performance. For example, on the OSWorld benchmark, UI-TARS achieves scores of 24.6 and 22.7 for 50-step and 15-step evaluations, respectively, outperforming comparable models like Claude, while on AndroidWorld, it scores 46.6 compared to GPT-4's 34.5. The system integrates several key innovations: it employs enhanced perception through a large-scale dataset of GUI screenshots for context-aware understanding; it utilizes unified action modeling to standardize interactions across platforms; it incorporates System-2 reasoning to enable deliberate multi-step decision-making via task decomposition, reflection, and milestone recognition; and it leverages iterative training with reflective online traces to continuously refine its performance with minimal human intervention. Available in 2B, 7B, and 72B parameter configurations, UI-TARS exemplifies a significant step forward in building versatile, reasoning-capable GUI agents that can perform complex interactions across diverse computing environments.

\subsection{Tülu 3 405B model}

The recent study on Tulu 3 \cite{lambert2025tulu3pushingfrontiers} presents a significant advancement in language model post-training, offering a fully transparent and reproducible framework for refining and enhancing LLMs. Unlike proprietary post-training pipelines, which often lack transparency regarding training data and methodologies, Tulu 3 provides a comprehensive open-source alternative, sharing the model weights and the complete post-training recipe, including datasets, training code, infrastructure, and evaluation tools. Built upon Llama 3.1 base models, Tulu 3 surpasses existing instruct-tuned models such as Llama 3.1-Instruct, Qwen 2.5, Mistral, and even proprietary models like GPT-4o-mini and Claude 3.5-Haiku in multiple benchmarks. The training methodology incorporates Supervised Fine-tuning (SFT), Direct Preference Optimization (DPO), and an innovative technique termed Reinforcement Learning with Verifiable Rewards (RLVR), which ensures reliable and measurable improvements in model behavior. To further enhance post-training effectiveness, the authors introduce a multi-task evaluation framework, which includes development and unseen benchmarks, standardized performance metrics, and rigorous decontamination of existing open datasets.

\subsection{Mistral AI Small 3 24B}

Mistral Small 3 \cite{mistral_small_3} is a highly efficient language model, competing with Llama 3.3 70B and Qwen 32B, while being 3× faster on the same hardware. Optimized for 80\% of generative AI tasks, it excels in instruction-following with low latency and achieves 81\% accuracy on MMLU at 150 tokens/s. Fewer layers significantly reduce inference time, making it ideal for local deployment. Unlike models such as DeepSeek R1, Mistral Small 3 has not been trained using Reinforcement Learning (RL) or synthetic data, making it an earlier-stage model in the production pipeline with the potential for additional refinement.

\begin{figure*}
    \centering
    \includegraphics[width=1\textwidth]{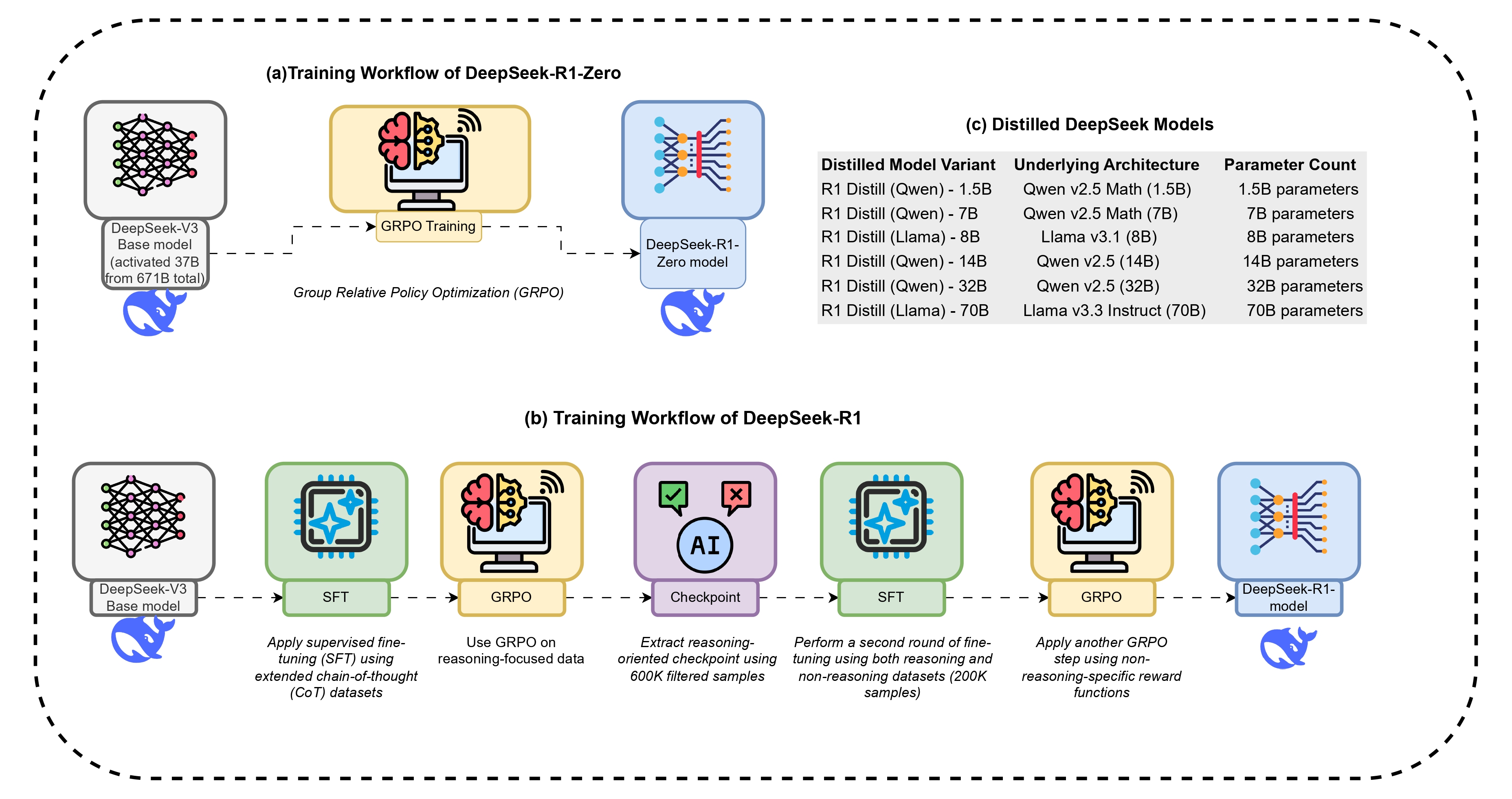}
    \caption{\textcolor{black}{Training Pipelines and Model Variants of DeepSeek-R1 and R1-Zero.}}
    \label{fig:fig_r2}
\end{figure*}

\subsection{DeepSeek-R1}
The DeepSeek-R1 \cite{deepseekai2025} series proposes advanced reasoning-optimized language models, including DeepSeek-R1-Zero, which is trained exclusively via Reinforcement Learning (RL) without Supervised Fine-Tuning (SFT), and DeepSeek-R1, which incorporates multi-stage training with cold-start data for improved reasoning and language coherence. DeepSeek-R1 matches OpenAI’s o1-1217 in reasoning tasks, leveraging key architectural innovations: Mixture of Experts (MoE) for efficient parameter utilization, Multihead Latent Attention (MLA) for reduced memory and compute costs, Multi-Token Prediction (MTP) for 2-3x faster token generation, and FP8 Quantization, which cuts memory usage by 75\% while maintaining stability. DeepSeek has released DeepSeek-R1-Zero, DeepSeek-R1, and six dense models (1.5B–70B) to support open research, providing a scalable, high-performance foundation for AI-driven reasoning and inference.

\textcolor{black}{In January 2025, DeepSeek introduced two influential models, DeepSeek-R1-Zero and DeepSeek-R1, which stood out for their impressive reasoning capabilities achieved at minimal training costs. These models challenged the notion that advanced reasoning requires extensive computational resources, highlighting a new direction in efficient AI development. As presented in Fig. \ref{fig:fig_r2} (a), DeepSeek-R1-Zero is built directly from the DeepSeek-V3-Base model without any supervised fine-tuning. Instead, it uses a rule-based reinforcement learning method called Group Relative Policy Optimization (GRPO), which uses structured reasoning rules to guide training. This design enables a simpler and more scalable pipeline. DeepSeek-R1 builds upon this foundation by addressing R1-Zero’s limitations, such as low output fluency and language inconsistency. It incorporates carefully selected high-quality data and multiple training stages, including reinforcement learning with rejection sampling, to enhance its general-purpose abilities and improve alignment with human preferences as presented in Fig. \ref{fig:fig_r2} (b).}

\textcolor{black}{To ensure accessibility across various platforms, DeepSeek also released a family of distilled models based on DeepSeek-R1, ranging from 1.5 billion to 70 billion parameters as presented in Fig. \ref{fig:fig_r2} (c). These smaller models are fine-tuned on synthetic data generated by the original R1 model, making them lightweight yet capable of strong reasoning performance. This approach allows the models to run efficiently on devices with limited hardware capabilities. Across multiple benchmarks in math, coding, general knowledge, and writing, DeepSeek-R1 models show strong competitiveness while offering significant cost advantages over other commercial options. In some cases, the models are up to 20 to 50 times more cost-effective than counterparts like OpenAI's o1, making them a practical solution for research and real-world applications.}

\subsection{Qwen 2.5 VL}

Qwen2.5-VL \cite{qwen2.5_vl} is a significant upgrade to Qwen2-VL \cite{wang2024}, enhancing vision-language capabilities across three sizes (3B, 7B, 72B). The Qwen2-VL Series proposes Naive Dynamic Resolution (NDR) for adaptive visual tokenization, enhancing efficiency and accuracy in vision-language tasks. It integrates Multimodal Rotary Position Embedding (M-RoPE) for improved text-image-video fusion and follows a unified image-video processing paradigm. By scaling both model size (2B, 8B, 72B) and training data, Qwen2-VL achieves competitive performance, with Qwen2-VL-72B matching GPT-4o and Claude 3.5-Sonnet on multimodal benchmarks. These innovations position Qwen2-VL as a state-of-the-art LVLM, advancing vision-language integration.

\subsection{HuatuoGPT-o1}

Chen et al. \cite{chen2024huatuogpt} addressed the challenge of enhancing reasoning in domain-specific applications by focusing on the medical field. In this domain, reliable reasoning is critical yet underexplored compared to mathematical tasks. The authors propose a novel two-stage training framework that leverages verifiable medical problems paired with a robust verifier to assess the correctness of model outputs. In the first stage, complex reasoning trajectories are generated and fine-tuned using a set of 40K carefully curated verifiable problems, ensuring each problem possesses a unique and verifiable solution. This stage is designed to guide the model through the intricacies of medical reasoning by using techniques such as backtracking, verification, and self-correction. The second stage involves reinforcement learning (using methods like PPO, outperforming alternatives such as RLOO and DPO) with verifier-based rewards, further refining the model's reasoning capabilities. The resulting model, HuatuoGPT-o1, demonstrates substantial improvements over both general and medical-specific baselines, achieving an increase of approximately 80\% on benchmarks like MedQA and PubMedQA. Additionally, experiments indicate that more complex reasoning chains (exceeding 700 tokens) yield better performance than straightforward reasoning approaches. Overall, the proposed model provides compelling evidence that a verifiable, two-stage training process can significantly enhance the complex reasoning abilities of LLMs in specialized domains such as medicine.

\subsection{Search-o1}

Li et al. \cite{li2025search} introduced Search-o1, an innovative framework designed to enhance the extended stepwise reasoning capabilities of large reasoning models (LRMs) by integrating a retrieval-augmented generation (RAG) mechanism with an advanced "Reason-in-Documents" module. LRMs such as OpenAI-o1 exhibit impressive multi-step reasoning but are often hampered by knowledge gaps and uncertainties that can compromise the accuracy of their outputs. To mitigate these limitations, Search-o1 dynamically retrieves external knowledge during uncertain reasoning points using an agentic search workflow marked by specialized tokens to trigger search queries. The typically verbose and noisy retrieved documents are refined via the Reason-in-Documents module. This dedicated LLM call analyzes and distills the essential information before reintegrating it into the ongoing chain of thought. Extensive experiments across diverse domains, including science, mathematics, coding, and open-domain question answering, reveal that Search-o1 not only enhances overall accuracy outperforming base models by 4.7\% and traditional RAG systems by up to 29.6\% on multi-hop questions but also achieves higher performance than human experts in fields like physics and biology. This approach represents a significant step in creating more reliable and versatile intelligent systems by effectively filtering out noise and seamlessly merging retrieved knowledge with internal reasoning processes.

\subsection{InternLM2}

Cai et al. \cite{cai2024internlm2} propose InternLM2, a cutting-edge open-source large language model (LLM) that sets new standards in comprehensive evaluations across multiple dimensions and benchmarks. Developed by the Shanghai AI Laboratory, InternLM2 distinguishes itself through innovative pre-training and optimization techniques, effectively capturing long-term dependencies with an impressive context window of up to 128k tokens. The model's pre-training process is meticulously detailed, encompassing diverse data types such as text, code, and specialized long-context data, and transitions from an initial 4k-token training phase to extended 32k-token stages, culminating in robust performance on challenging tests like the 200k "Needle-in-a-Haystack" benchmark. Furthermore, InternLM2 is refined via Supervised Fine-Tuning (SFT) and a novel Conditional Online Reinforcement Learning from Human Feedback (COOL RLHF) strategy, which adeptly addresses issues like conflicting human preferences and reward hacking. Notably, the 8B parameter version of InternLM2 demonstrates remarkable efficiency, having been trained on only 4 trillion tokens yielding over 75\% in cost savings compared to similar models while delivering superior results on complex reasoning tasks (scoring 83.0\% on MATH-500 and 20.0\% on AIME2024) and matching or surpassing contemporaries like Llama3.1-8B, Qwen2.5-7B, and OpenAI GPT-4o mini. This work marks a significant step forward in the open-source LLM landscape and provides valuable insights into model evolution and the practical realization of advanced reasoning capabilities in a cost-effective framework.

\subsection{MiniCPM-o 2.6}

Yao et al. \cite{yao2024minicpm} introduced MiniCPM-o 2.6, an advanced on-device multimodal large language model (MLLM) that extends the capabilities of its predecessors in the MiniCPM-V series. Uniquely designed to handle a wide array of modalities, including image, video, text, and audio, it delivers high-quality text and speech outputs in an end-to-end manner, all within an 8-bit parameter framework. Notably, MiniCPM-o 2.6 demonstrates leading visual capabilities, achieving an average score of 70.2 on OpenCompass and outperforming proprietary models like GPT-4o-202405, Gemini 1.5 Pro, and Claude 3.5 Sonnet in single and multi-image as well as video understanding tasks. Its state-of-the-art speech capabilities are evident in bilingual real-time conversations, enhanced ASR, speech-to-text translation performance, and flexible voice customization features such as emotion, speed, and style control. Furthermore, the model excels in multimodal live streaming, effectively processing continuous video and audio streams with real-time interaction, and exhibits robust OCR performance on images of varied aspect ratios and resolutions, surpassing models under 25B parameters on OCRBench. Another highlight is its superior efficiency MiniCPM-o 2.6 achieves a remarkable token density, encoding 1.8 million pixels in only 640 tokens, thereby reducing latency and resource consumption.

\subsection{KaLM-Embedding}

Hu et al. \cite{hu2025kalm} introduced KaLM-Embedding, a multilingual embedding model that emphasizes the importance of training data quality in the era of retrieval-augmented generation. Unlike prior work that often overlooks data curation, KaLM-Embedding is trained on a large corpus of cleaner, more diverse, and domain-specific data. The model leverages several key techniques to boost performance: persona-based synthetic data generation, which diversifies examples distilled from large language models; ranking consistency filtering, designed to eliminate noisy and uninformative samples; and semi-homogeneous task batch sampling to enhance training efficiency. Departing from traditional BERT-like architectures, the authors build on the Qwen2-0.5B pre-trained model, effectively adapting an auto-regressive language model framework to general embedding tasks. Extensive evaluations on the MTEB benchmark, spanning multiple languages, demonstrate that KaLM-Embedding outperforms comparable models with fewer than 1B parameters, achieving an average score of 64.53. Additionally, the model supports flexible embedding dimensions through its Matryoshka Representation Learning strategy, setting a new standard for multilingual embeddings in low-parameter regimes.

\subsection{rStar-Math}

Guan et al. \cite{guan2025rstar} introduced rStar-Math, a framework that leverages small language models (SLMs) to achieve state-of-the-art performance in mathematical reasoning, rivaling and even surpassing the capabilities of larger models like OpenAI o1-preview. The approach employs deep thinking via Monte Carlo Tree Search (MCTS), where an SLM-based policy model generates multiple code-augmented chain-of-thought (CoT) reasoning trajectories. Each step in these trajectories comprises a natural language explanation and executable Python code, ensuring that reasoning steps are verified through code execution. A key innovation lies in the novel synthesis of CoT data via extensive MCTS rollouts, which produce step-by-step verified reasoning paths to train the policy model. Complementing this, the authors propose an improved process reward model (PRM) training method that avoids naive step-level scoring by leveraging binary code verification outcomes, and they introduce a self-evolution strategy whereby both the policy SLM and PRM are iteratively refined from an initial dataset of 747k math problems. After four rounds of self-evolution involving millions of synthesized solutions, rStar-Math boosts accuracy on the MATH benchmark to 90.0\% using a 7B parameter model, outperforming o1-preview, and solves 53.3\% of problems on the USA Math Olympiad (AIME), ranking within the top 20\% among high school competitors.

\subsection{phi-4}

Microsoft \cite{abdin2024phi} introduced phi-4, a 14-billion parameter language model that significantly advances performance on STEM-focused QA tasks by leveraging a training strategy centered on data quality. Unlike many models that rely predominantly on organic data sources such as web content or code, phi-4 integrates large-scale, high-quality synthetic data throughout its pre-training process. This synthetic data is generated through multi-agent, self-revision workflows, enabling Phi-4 to distill and surpass the capabilities of its teacher model, GPT-4, particularly in reasoning, mathematics, and code generation. Despite minimal modifications to the phi-3 architecture, phi-4, trained on 9.8 trillion tokens and supporting a 16K token context length, performs comparably to models with 70B parameters. The model was developed with a comprehensive post-training scheme including supervised fine-tuning (SFT) and Direct Preference Optimization (DPO) for safety alignment and required 21 days of training on 1920 H100-80G GPUs. Phi-4 exemplifies how strategic data generation and refined training curricula can dramatically enhance model performance, especially in specialized STEM domains.

\subsection{DeepSeek-V3}

DeepSeek Team proposed \cite{liu2024deepseek} DeepSeek-V3, a state-of-the-art Mixture-of-Experts (MoE) language model comprising 671 billion parameters, with 37 billion activated per token during inference. DeepSeek-V3 leverages advanced architectures such as Multi-head Latent Attention (MLA) and DeepSeekMoE to ensure efficient inference and cost-effective training, previously validated in its predecessor, DeepSeek-V2. Notably, the model pioneers an auxiliary-loss-free strategy for load balancing and adopts a multi-token prediction training objective to boost performance. Pre-training was conducted on an expansive corpus of 14.8 trillion high-quality tokens, followed by rigorous Supervised Fine-Tuning and Reinforcement Learning stages, enabling DeepSeek-V3 to achieve performance that matches or exceeds leading closed-source models like OpenAI's GPT-4o and Anthropic’s Claude-Sonnet-3.5. Despite its scale, the training process was remarkably stable and efficient, requiring only 2.788 million H800 GPU hours, approximately \$5.5 million in cost, while supporting FP8 training and achieving a throughput of 60 tokens per second. Additionally, the model offers significant cost advantages with low per-input and per-output expenses, and it is trained on both English and Chinese data, marking a notable engineering milestone in large-scale language models.

\subsection{DeepSeekMath 7B}

DeepSeek Team proposed \cite{shao2024deepseekmath} DeepSeekMath 7B, a language model specifically designed to tackle the challenges of mathematical reasoning. Building on the DeepSeek-Coder-Base-v1.5 7B architecture, the model is further pre-trained on an extensive corpus of 120 billion math-related tokens extracted from Common Crawl, complemented by natural language and code data. As a result, DeepSeekMath 7B achieves an impressive 51.7\% accuracy on the competition-level MATH benchmark without the use of external toolkits or ensemble voting techniques. This performance approaches that of more advanced models like Gemini-Ultra and GPT-4. Furthermore, when employing self-consistency over 64 samples, the model's score rises to 60.9\%. Two key factors underpin this performance: a meticulously engineered data selection pipeline that capitalizes on publicly available web data and the introduction of Group Relative Policy Optimization (GRPO), a Proximal Policy Optimization (PPO) variant. GRPO not only enhances the mathematical reasoning capabilities of the model but also optimizes memory usage by obviating the need for a separate value function, integrating a KL term directly into the loss function, and working effectively with rule-based and score-based reward models. 

\subsection{Qwen 2.5 model}

Qwen team \cite{yang2024qwen2} introduced Qwen2.5, a comprehensive series of large language models meticulously engineered to address a wide range of applications. Building on previous iterations, Qwen2.5 significantly enhances both pre-training and post-training processes. In the pre-training stage, the dataset has been scaled up from 7 trillion to 18 trillion high-quality tokens filtered and classified using existing LLMs to robustly ground the models in common sense, expert knowledge, and reasoning capabilities. On the post-training side, the framework incorporates an extensive supervised fine-tuning regimen with over one million samples covering long texts, mathematics, coding, multilingual tasks, and a multistage reinforcement learning pipeline. This includes offline RL (via Direct Preference Optimization on 150K training pairs) and online RL (using GRPO with a 72B reward model for truthfulness, helpfulness, and safety) to refine complex reasoning and instruction further following. Additionally, synthetic data generation augmented by combining chain-of-thought with rejection sampling and translating instructions to boost multilingual performance is critical in enhancing performance. Architecturally, Qwen2.5 leverages modern design elements including GQA, SwiGLU, RoPE, QKV bias in attention, and RMSNorm. The models are released in a variety of sizes (from 0.5B to 72B parameters) and formats, with open-weight base and instruction-tuned versions, as well as proprietary MoE variants like Qwen2.5-Turbo and Qwen2.5-Plus available on Alibaba Cloud Model Studio. Comprehensive evaluations across benchmarks for language understanding, reasoning, mathematics, coding, and human preference alignment reveal that the flagship Qwen2.5-72 B-Instruct outperforms many existing models, including some that are substantially larger, underscoring the pivotal role of data quality and advanced training strategies in achieving state-of-the-art performance.

\subsection{ModernBERT}

Warner et al. \cite{warner2024smarter} proposed ModernBERT, a next-generation encoder-only transformer model that significantly advances the performance and efficiency of traditional models like BERT and RoBERTa. Recognizing the enduring value of encoder architectures for retrieval and classification tasks, the authors incorporate a suite of modern optimizations such as an extended native context length of 8,192 tokens, Flash Attention 2, RoPE embeddings, and an alternating attention mechanism, thereby delivering a significant Pareto improvement over older encoders. Trained on 2 trillion tokens, predominantly sourced from English and code data, ModernBERT is released in two sizes (base at 139M parameters and large at 395M parameters) and achieves state-of-the-art results across a broad spectrum of evaluation benchmarks, including diverse classification tasks and both single- and multi-vector retrieval scenarios. In addition to its superior downstream performance, ModernBERT demonstrates remarkable speed and memory efficiency. It operates 2–4 times faster than comparable models with mixed-length inputs and is designed to run effectively on common GPUs.

\subsection{Gemini 2.0 Flash}

Google DeepMind introduced Gemini 2.0 Flash, \cite{googleblog2024gemini}, marking a significant evolution in the Gemini series for developers. Building on the success of Gemini 1.5 Flash, this new model is engineered to be twice as fast while delivering enhanced performance across key benchmarks. Notably, Gemini 2.0 Flash incorporates a Multimodal Live API that supports real-time audio and video streaming, along with native multimodal output capabilities enabling the generation of integrated text, audio, and image responses from a single API call. The update also features native tool use, where the model can seamlessly call external tools such as Google Search and execute code, thereby facilitating the development of agentic coding assistants like the experimental AI-powered code agent Jules. Furthermore, the model supports multilingual native audio output with eight high-quality voices and extends its language support to 109 languages. These advancements improve the efficiency and interactivity of AI applications and enhance their performance on real-world software engineering tasks, as evidenced by a 51.8\% % achievement on SWE-bench Verified.

\subsection{Llama 3.3 70B-Instruct and Llama 405B}

Meta team released Llama 3.3 70 B-Instruct \cite{meta_llama2023llama33}, a 1-to-1 architectural update that significantly boosts performance through improved post-training techniques. Without changing the underlying model architecture, Llama 3.3 achieves an overall ~4\% enhancement across benchmarks, with particularly notable gains in math and coding tasks registering a +9\% improvement on the MATH benchmark and +8\% on HumanEval. These enhancements enable it to match the performance of high-end models such as OpenAI GPT-4o, Anthropic Haiku 3.5, and Gemini Flash. Maintaining the same knowledge cutoff as its predecessors, Llama 3.3 supports eight languages, including English, German, French, Italian, and Spanish, with training extending to additional languages.

Meta's Llama 3.1 \cite{llama31} is a series of large language models in three distinct sizes: 8B, 70B, and an impressive 405B parameter variant. Notably, the 405B model is reported to match or exceed the performance of GPT-4o on various text benchmarks, signaling a significant leap in LLM capabilities. All variants are provided in Instruct and Base versions and support an extensive 128k token context, enhancing their ability to handle long documents and complex tasks. The models are trained on an expansive corpus of over 15 trillion tokens and fine-tuned with 25 million human and synthetic samples. This bolsters their multilingual support across eight languages: English, German, and French. Additionally, the release emphasizes a commercial-friendly license that facilitates using model outputs for improving other LLMs, along with optimized inference through various quantization techniques such as FP8, AWQ, and GPTQ. Enhanced coding and instruction following capabilities, improving up to 12\% as well as support for tool use and function calling, further highlight the comprehensive improvements of Llama 3.1.

\subsection{Amazon Nova}

Amazon proposed Nova \cite{aws2023amazonNova}, a comprehensive suite of large language models designed to deliver industry-leading price performance and versatility across various applications. Amazon Nova is offered in multiple variants tailored to different use cases: the Micro model for text-only tasks, the Lite model for multimodal applications, the Pro model for high-capability scenarios, and the forthcoming Premier version scheduled for release in 2025. In addition, specialized generation models are provided on Canvas for image generation and Reel for video generation, enabling dynamic and creative content production. Notably, the Nova suite supports exceptionally long context lengths of up to 300,000 tokens and can process over 200 languages, achieving benchmark performance comparable to models like Llama 3. Currently available exclusively in AWS Regions in the United States, Amazon Nova also incorporates watermarking features to enhance output security and integrity. Moreover, the models are designed for seamless integration and fine-tuning within Amazon Bedrock, offering a flexible development environment. Pricing is structured to meet diverse needs, with the Micro variant priced at \$0.035 per million input tokens and \$0.14 per million output tokens, Lite at \$0.06/\$0.24, and Pro at \$0.80/\$3.20, underscoring the suite’s commitment to cost efficiency alongside high performance.

\subsection{QwQ-32B-Preview}

QwQ-32B-Preview \cite{qwenlm2023qwq32b} represents a significant milestone as the first open-weight release for an OpenAI-o1-like reasoning model from the Qwen team. Built on the Qwen2.5 architecture, this experimental model encompasses 32.5 billion parameters and supports an extensive context length of 32,768 tokens. It demonstrates competitive performance across a suite of challenging benchmarks achieving 65.2\% on GPQA, 50.0\% on AIME, 90.6\% on MATH-500, and 50.0\% on LiveCodeBench thereby surpassing OpenAI's O1 mini and positioning itself as a strong contender relative to O1 preview models. However, despite its promising results, QwQ-32B-Preview exhibits notable limitations, including issues with language mixing, recursive reasoning loops, and safety concerns, which signal avenues for future refinement. Released under the Apache 2.0 license and available on Hugging Face, QwQ-32B-Preview offers researchers transparent access to advanced reasoning capabilities while highlighting key challenges in developing robust, large-scale reasoning systems.

\subsection{Qwen2.5-Coder-32B}

Qwen2.5-Coder-32B \cite{qwenlm2023qwen25} marks a significant advancement in code-focused language models, demonstrating performance that nearly matches that of Claude Sonnet 3.5 despite its relatively smaller size of 32 billion parameters. Early evaluations and user feedback suggest that Qwen2.5-Coder-32B excels in diverse coding benchmarks, notably outperforming Claude Sonnet 3.5 on HumanEval (92.7 vs. 92.1) and EvalPlus (86.3 vs. 85.9), and delivering superior results in fill-in-the-middle tasks. Moreover, it exhibits high proficiency in code repair, achieving a score of 73.7 on the Aider benchmark. Released in both Base and Instruct versions, the model supports over 40 languages and accommodates an impressive 128K token context length, effectively handling long and complex code contexts. Its training leverages extended system prompts often exceeding 16K tokens with rich examples and documentation to enhance its understanding and code generation.

\begin{figure*}[htbp]
  \centering
  \begin{tikzpicture}
    \begin{groupplot}[
      group style={
        group size=1 by 2,
        vertical sep=2.7cm
      },
      width=15cm,
      height=6cm,
      ybar stacked,
      enlarge x limits=0.05,
      symbolic x coords={
        Gemini-1.5-pro,
        Mistral-NeMo,
        Codegemma,
        Llama3.1-Nemotron,
        GPT-4o-mini,
        GPT-4o,
        O1-mini,
        Deepseek-R1,
        Qwen3-32B,
        Qwen3-235B-A22B,
        Gemini 2.5 Pro,
        ChatGPT-4o
      },
      xtick=data,
      xticklabel style={rotate=45,anchor=east,font=\small},
      nodes near coords,
      label style={font=\footnotesize},
tick label style={font=\footnotesize},
nodes near coords style={font=\footnotesize},
      cycle list={
        {fill=green!50!white,draw=black},
        {fill=red!50!white,draw=black},
        {fill=black!50!white,draw=black}
      },
      legend style={
        at={(0.5,1.2)},anchor=south,legend columns=3,
        /tikz/every even column/.append style={column sep=1cm}
      }
    ]

      % (a) Mathematics
      \nextgroupplot[
        title={(a) Mathematics Questions},
        ymin=0,ymax=260,
        legend entries={Correct,Incorrect,Skipped}
      ]
      \addplot coordinates {
        (Gemini-1.5-pro,16)  (Mistral-NeMo,19)
        (Codegemma,21)       (Llama3.1-Nemotron,27)
        (GPT-4o-mini,28)     (GPT-4o,37)
        (O1-mini,85)         (Deepseek-R1,175)
        (Qwen3-32B,180)      (Qwen3-235B-A22B,190)
        (Gemini 2.5 Pro,194) (ChatGPT-4o,196)
      };
      \addplot coordinates {
        (Gemini-1.5-pro,229) (Mistral-NeMo,226)
        (Codegemma,224)      (Llama3.1-Nemotron,218)
        (GPT-4o-mini,217)    (GPT-4o,205)
        (O1-mini,64)         (Deepseek-R1,50)
        (Qwen3-32B,45)       (Qwen3-235B-A22B,43)
        (Gemini 2.5 Pro,41)  (ChatGPT-4o,40)
      };
      \addplot coordinates {
        (Gemini-1.5-pro,0)   (Mistral-NeMo,0)
        (Codegemma,0)        (Llama3.1-Nemotron,0)
        (GPT-4o-mini,0)      (GPT-4o,3)
        (O1-mini,96)         (Deepseek-R1,19)
        (Qwen3-32B,19)       (Qwen3-235B-A22B,11)
        (Gemini 2.5 Pro,10)  (ChatGPT-4o,9)
      };

      % (b) CS & Security
      \nextgroupplot[
        title={(b) Computer Science and Cybersecurity Questions},
        ymin=0,ymax=520
      ]
      \addplot coordinates {
        (Gemini-1.5-pro,64)   (Mistral-NeMo,65)
        (Codegemma,60)       (Llama3.1-Nemotron,82)
        (GPT-4o-mini,95)     (GPT-4o,145)
        (O1-mini,184)        (Deepseek-R1,175)
        (Qwen3-32B,180)      (Qwen3-235B-A22B,200)
        (Gemini 2.5 Pro,221) (ChatGPT-4o,222)
      };
      \addplot coordinates {
        (Gemini-1.5-pro,441) (Mistral-NeMo,425)
        (Codegemma,445)      (Llama3.1-Nemotron,402)
        (GPT-4o-mini,381)    (GPT-4o,293)
        (O1-mini,137)        (Deepseek-R1,220)
        (Qwen3-32B,222)      (Qwen3-235B-A22B,212)
        (Gemini 2.5 Pro,200) (ChatGPT-4o,201)
      };
      \addplot coordinates {
        (Gemini-1.5-pro,0)   (Mistral-NeMo,15)
        (Codegemma,0)        (Llama3.1-Nemotron,21)
        (GPT-4o-mini,29)     (GPT-4o,67)
        (O1-mini,184)        (Deepseek-R1,110)
        (Qwen3-32B,103)      (Qwen3-235B-A22B,93)
        (Gemini 2.5 Pro,82)  (ChatGPT-4o,82)
      };

    \end{groupplot}
  \end{tikzpicture}
  \caption{\textcolor{black}{Performance of 12 examined models in different categories}}
  \label{fig:performance}
\end{figure*}

\subsection{Hunyuan-Large}

Tencent team introduced Hunyuan-Large \cite{sun2024hunyuan}, the most prominent open-source Transformer-based mixture-of-experts (MoE) model to date, featuring 389 billion parameters in total with 52 billion parameters activated during generation and supporting sequence lengths of up to 256K tokens. Evaluated across a diverse array of tasks, including language understanding, logical reasoning, mathematical problem-solving, coding, long-context processing, and aggregated benchmarks, Hunyuan-Large outperforms Llama3.1-70B and is competitive with the substantially larger Llama3.1-405B model. Key innovations underpinning its performance include utilizing an unprecedented scale of synthetic data (1.5 trillion tokens out of 7 trillion total tokens), a mixed expert routing strategy, key-value cache compression, and an expert-specific learning rate schedule. The model comprises 16 experts and is released in multiple versions (Pretrain, Instruct, and FP8), with efficient deployment possible on a single H100 node (8×) in FP8 precision. Additionally, the authors investigate scaling laws and learning rate schedules specific to MoE architectures, providing valuable insights for future model development and optimization. 

\subsection{Granite 3.0}

IBM team introduced Granite 3.0 \cite{ibm2023granite}, the next generation of lightweight, open foundation models designed for enterprise applications, including on-premise and on-device scenarios. Developed by IBM, Granite 3.0 comprises four variants: 8B and 2B Dense models alongside 3B and 1B Mixture-of-Experts (MoE) models that collectively offer native support for 12 natural languages and 116 programming languages. Trained on a massive 12 trillion tokens, these models are engineered with advanced features such as an additional Speculator Head that accelerates token processing by up to 220\%, and they demonstrate strong safety performance along with capabilities in multilingual processing, coding, and function calling. Comprehensive evaluations reveal that Granite 3.0 consistently achieves state-of-the-art performance for its size, outperforming Llama 3.1 8B on RAGBench across 11 datasets and scoring 37.6 on the OpenLLM Leaderboard compared to Llama 3.1 8B’s 37.3. The report details both pre-training and post-training methodologies, offering transparent insights into the datasets and techniques used, and all models are released under the permissive Apache 2.0 license on Hugging Face. Upcoming enhancements include support for 128K token context windows, improved multilingual capabilities, and multimodal functionality, further broadening the applicability of Granite 3.0 in diverse real-world settings.

\subsection{Pyramid Flow SD3, Molmo and Moshi models}

Jin et al. \cite{jin2024pyramidal} proposed an approach to video generation that addresses the immense computational challenges of modeling extensive spatiotemporal data. Traditional methods often rely on a cascaded architecture to reduce the computational burden by handling lower resolutions in preliminary stages, but these methods suffer from isolated sub-stage optimizations that limit knowledge transfer and flexibility. In contrast, the authors propose a unified pyramidal flow matching algorithm that reconceptualizes the denoising trajectory as a series of interconnected pyramid stages, with only the final stage operating at full resolution. This design preserves continuity across scales and enables an efficient, end-to-end optimization using a single Diffusion Transformer (DiT) model. Additionally, the framework incorporates an autoregressive temporal pyramid to compress historical full-resolution information, further reducing computational demands. Experimental results demonstrate the method's efficacy by generating high-quality videos up to 10 seconds long at 768p resolution and 24 FPS within 20.7k A100 GPU training hours. The release of Pyramid Flow SD3 under the MIT license, featuring a 2B parameter unified DiT, marks a significant milestone as the first robust open-source text-to-video and image-to-video generation model, broadening accessibility and practical applications in video synthesis research.

Ai2 team introduced Molmo \cite{deitke2024molmo}, a groundbreaking family of open-source vision-language models (VLMs) that set a new state-of-the-art in the open-weight domain. Unlike many advanced VLMs that rely on synthetic data distilled from proprietary models, Molmo is built entirely from scratch using novel, high-quality datasets collectively known as PixMo. PixMo comprises meticulously curated resources, including highly detailed image captions for pre-training, a free-form image Q\&A dataset for fine-tuning, and an innovative 2D pointing dataset, all collected without external proprietary VLMs. Molmo offers four variants, ranging from a 7B MoE (A1B) to dual 7B and a 72B version, with the 72B model based on Qwen2-72B and utilizing OpenAI CLIP as its vision backbone. Extensive evaluations across 11 academic benchmarks and 325,231 human pairwise comparisons demonstrate that Molmo 72B not only outperforms other open models like Llama 3.2 90B, Pixtral 12B, and Qwen 2 72B VL, but also matches the performance of larger proprietary models such as OpenAI GPT-4o, Google Gemini 1.5 Pro, and Anthropic Claude Sonnet 3.5.

Moshi \cite{defossez2024moshi} is a cutting-edge speech-text foundation model and full-duplex spoken dialogue framework that reimagines how AI handles spoken conversations by treating dialogue as a unified speech-to-speech generation process. Unlike traditional systems that rely on separate components for voice activity detection, speech recognition, textual dialogue, and text-to-speech which often introduce significant latency and lose crucial non-linguistic cues Moshi employs a 7B parameter Temporal Transformer to generate speech tokens directly from the residual quantizer of a neural audio codec, named Mimi, while concurrently modeling its speech and that of the user in parallel streams. This innovative "Inner Monologue" approach, which predicts time-aligned text tokens as a prefix to audio tokens, not only enhances the linguistic quality of the generated speech but also enables real-time streaming, with a theoretical latency of just 160ms (200ms in practice on an L4 GPU) and an efficient 12.5 Hz representation at only 1.1 kbps bandwidth. Available under a permissive cc-by-4.0 license on Hugging Face and implemented in both Python (via PyTorch and MLX) and Rust, Moshi supports fine-tuned male (Moshiko) and female (Moshika) voice models, offers various quantization options, and represents a significant advancement in real-time, on-device spoken dialogue systems by effectively handling overlapping speech, interruptions, and other complex conversational dynamics.

\subsection{\textcolor{black}{Performance on DIA Benchmark Dataset}}

\begingroup
\color{black}

Figure~\ref{fig:performance} summarizes the performance of the twelve evaluated models on the DIA Benchmark Dataset \cite{tihanyi2024dynamic}, broken down into (a) mathematics questions and (b) computer science \& cybersecurity questions. We report the number of questions answered correctly, incorrectly, and skipped for each model. The DIA Benchmark Dataset combines visual and textual components ranging from PDF snippets and encoded files to CAPTCHAs and steganographic images with cryptography, reverse engineering, web security, mathematics, and logic challenges. Crucially, many tasks cannot be solved through reasoning alone and require external tools such as a Python interpreter or Linux command-line utilities. Since standard LLM API endpoints do not support tool invocation, an effective model must recognize when a problem exceeds pure text-based inference (e.g., computing the next prime of a large integer) and choose to skip rather than guess, as correct answers often involve long strings or very large numbers that are infeasible to derive manually.

In the mathematics subset (Figure~\ref{fig:performance}a), ChatGPT-4o achieved 196 correct answers, 40 incorrect answers, and skipped 9 items. Gemini 2.5 Pro followed closely with 194 correct, 41 incorrect, and 10 skips, while Qwen3-235B-A22B recorded 190 correct, 43 incorrect, and 11 skips. Mid-sized models such as Deepseek-R1 and Qwen3-32B answered 175 and 180 correctly, respectively, but had higher skip counts (19 each) and 50–45 incorrect answers. GPT-4o answered 37 correctly, made 205 incorrect attempts, and skipped 3, demonstrating occasional self-deference. In contrast, O1-mini solved 85 items correctly, with 64 incorrect answers and a high skip count of 96, indicating a conservative strategy. The smallest models such as Codegemma (21 correct, 224 incorrect, 0 skipped) and Mistral-NeMo (19 correct, 226 incorrect, 0 skipped) attempted nearly every question but achieved minimal success.

For computer science and cybersecurity (Figure~\ref{fig:performance}b), ChatGPT-4o again led with 222 correct, 201 incorrect, and 82 skips. Gemini 2.5 Pro achieved 221 correct, 200 incorrect, and 82 skips, and Qwen3-235B-A22B answered 200 correctly, with 212 incorrect and 93 skips. Deepseek-R1 and Qwen3-32B both solved around 175–180 items, but Deepseek-R1 made 220 incorrect attempts with 110 skips, while Qwen3-32B had 222 incorrect and 103 skips. O1-mini delivered 184 correct, 137 incorrect, and 184 skips, reflecting a highly cautious approach. GPT-4o and GPT-4o-mini answered 145 and 95 correctly, respectively, with 293 and 381 incorrect, and 67 and 29 skips, showing improved willingness to attempt conceptual problems but still deferring on harder cases. Codegemma and Mistral-NeMo again exhibited low performance (60–65 correct) with minimal skipping.

\endgroup

%=========================%
%      Table 1: Groups 1-4
%=========================%
\begin{table*}[htbp]
\centering
\scriptsize
\caption{Training Methodologies (Groups 1--6): General Training Approaches, MoE \& Architectural Innovations, Retrieval-Augmented Generation (RAG), Chain-of-Thought \& Self-Improvement, Test-Time Compute Scaling \& Distillation, and Others.}
\label{tab:training_methodologies_1}
\begin{adjustbox}{max width=\textwidth}
\begin{tabularx}{\textwidth}{@{}%
  >{\raggedright\arraybackslash}p{3.2cm}  % Work Name
  >{\centering\arraybackslash}p{1.1cm}    % Year
  >{\raggedright\arraybackslash}p{2.8cm}  % Category
  X                                     % Key Techniques / Ideas
  X                                     % Main Results / Impact
@{}}
\toprule
\textbf{Work Name} & \textbf{Year} & \textbf{Category} & \textbf{Key Techniques / Ideas} & \textbf{Main Results / Impact} \\
\midrule
\multicolumn{5}{l}{\textbf{Group 1: General Training Approaches}} \\
\midrule
Agent Q \cite{putta2024agent} & 2024 & General Training & Combines guided Monte Carlo Tree Search, self-critique, and off-policy DPO. & Boosts zero-shot success from 18.6\% to 81.7\% (95.4\% with online search). \\
LoRA \cite{biderman2024lora} & 2024 & Parameter-Efficient FT & Fine-tunes low-rank perturbations as a regularizer. & Maintains diverse generations, though underperforms full finetuning on target tasks. \\
CoALA \cite{sumers2023cognitive} & 2023 & Cognitive Architecture & Integrates modular memory systems with defined internal/external action spaces. & Enhances grounding and reasoning in language agents. \\
Learn-by-interact \cite{su2025learn} & 2025 & Data-Centric Interaction & Synthesizes agent-environment trajectories via backward construction. & Improves in-context and fine-tuning performance (up to 19.5\% gains). \\
SFT \cite{chu2025sft} & 2025 & Supervised FT & Compares SFT-only, RL-only, and hybrid SFT$\rightarrow$RL pipelines. & RL improves generalization while SFT stabilizes output formatting. \\
\midrule
\multicolumn{5}{l}{\textbf{Group 2: MoE \& Architectural Innovations}} \\
\midrule
MoE \cite{jiang2024mixtral} & 2024 & Mixture-of-Experts & Routes tokens to multiple expert networks via a learned router. & Enables vast parameter scaling with efficient inference. \\
HMoE \cite{wang2024hmoe} & 2024 & Heterogeneous MoE & Uses experts of varying sizes with a penalty favoring smaller ones. & Achieves lower loss with fewer activated parameters. \\
ESFT \cite{tang2024understanding} & 2024 & Expert-Specialized FT & Selectively fine-tunes task-relevant experts in MoE models. & Reduces memory usage by up to 90\% and training time by ~30\%. \\
MoA \cite{wang2024mixture} & 2024 & Mixture-of-Agents & Layers multiple LLM agents for iterative response refinement. & Achieves state-of-the-art scores on benchmarks like AlpacaEval 2.0. \\
Uni-MoE \cite{li2025uni} & 2025 & Multimodal MoE & Uses modality-specific encoders with expert-level parallelism. & Matches or outperforms existing unified multimodal models. \\
MLA \cite{meng2025transmla} & 2025 & Attention Mechanism & Compresses latent states in key-value layers via low-rank matrices. & Reduces KV cache size and speeds up inference. \\
M-RoPE \cite{wang2024} & 2024 & Positional Encoding & Decomposes rotary embeddings into temporal, height, and width components. & Enhances multi-modal positional encoding and extrapolation. \\
MTP \cite{deepseekai2025} & 2025 & Multi-Token Prediction & Predicts multiple tokens in parallel with adaptive granularity. & Achieves 1.5× faster generation and 18\% improved long-form perplexity. \\
\midrule
\multicolumn{5}{l}{\textbf{Group 3: Retrieval-Augmented Generation (RAG)}} \\
\midrule
Long-context RAG \cite{databricks2023longcontext} & 2023 & RAG (Long Context) & Evaluates RAG with context lengths up to 2M tokens from vector databases. & Identifies optimal performance in the 32k–64k token range. \\
Searching RAG \cite{wang2024searching} & 2024 & RAG (Search) & Dissects multi-step retrieval workflows including advanced chunking and reranking. & Provides recommendations to balance performance and efficiency. \\
CoRAG \cite{wang2025chain} & 2025 & CoRAG & Iteratively reformulate queries to build a chain of retrievals. & Improves exact match scores on multi-hop QA by $>10$ points. \\
Reason-in-Docs \cite{li2025search} & 2025 & RAG (Reasoning) & Integrates a dedicated module to refine retrieved content. & Enhances the reliability of multi-step reasoning. \\
Contextualized Embeddings \cite{morris2024contextual} & 2024 & RAG (Embeddings) & Incorporates neighboring document context into dense embeddings. & Achieves state-of-the-art retrieval on benchmarks (MTEB). \\
RAG with Self-Reasoning \cite{xia2024improving} & 2024 & RAG (Self-Reasoning) & Uses internal reasoning trajectories to refine retrieved evidence. & Improves factual accuracy and traceability in generated answers. \\
\midrule
\multicolumn{5}{l}{\textbf{Group 4: Chain-of-Thought \& Self-Improvement}} \\
\midrule
Coconut \cite{hao2024training} & 2024 & Continuous Thought & Replaces textual CoT with latent-space iterative processing via special tokens. & Outperforms standard CoT on complex planning tasks (e.g., ProntoQA). \\
STaR \cite{zelikman2022star} & 2022 & Self-Taught Reasoner & Iteratively refines self-generated CoT with corrective feedback loops. & Substantially improves GSM8K and CommonsenseQA performance. \\
V-STaR \cite{hosseini2024v} & 2024 & Self-Taught Reasoner & Utilizes both correct and incorrect outputs to form preference pairs. & Yields 4–17\% gains on math and code tasks. \\
Quiet-STaR \cite{zelikman2024quiet} & 2024 & Self-Taught Reasoner & Generates parallel token-level "thoughts" to guide predictions via REINFORCE. & Boosts GSM8K from 5.9\% to 10.9\% and improves CommonsenseQA similarly. \\
Self-Taught Evaluators \cite{wang2024selftaughtevaluators} & 2024 & Self-Taught Evaluators & Trains evaluators with synthetic preference pairs from self-generated outputs. & Raises evaluator accuracy from ~75\% to ~88\%. \\
START \cite{li2025start} & 2025 & Tool-Augmented Reasoning & Enhances long CoT reasoning with external tool use via hint-based self-learning. & Achieves state-of-the-art performance on AMC23 (95.0\%), AIME24 (66.7\%), and LiveCodeBench (47.3\%). \\
\midrule
\multicolumn{5}{l}{\textbf{Group 5: Test-Time Compute Scaling \& Distillation}} \\
\midrule
Test-Time Compute Scaling \cite{brown2024large, huggingface2023scaling} & 2024 & Test-Time Scaling & Uses repeated sampling (e.g., Best-of-N, majority voting) to boost coverage. & Enables smaller models to rival larger ones with sufficient computing. \\
SKD \cite{xu2024speculative} & 2024 & Knowledge Distillation & Interleaves student-generated tokens with teacher predictions for adaptive replacement. & Outperforms traditional KD methods in various tasks. \\
\midrule
\multicolumn{5}{l}{\textbf{Group 6: Others}} \\
\midrule
Mind Evolution \cite{lee2025evolving} & 2025 & Evolutionary Search & Uses iterative, evolutionary strategies to generate and refine candidate responses. & Achieves $>$98\% success rate on planning benchmarks (e.g., TravelPlanner). \\
SemHash \cite{minishlab2025semhash} & 2025 & Data Deduplication & Leverages Model2Vec embeddings with ANN for semantic deduplication. & Processes 1.8M WikiText records in 83 seconds on CPU. \\
Agentless \cite{xia2024agentless} & 2024 & Non-Agent Framework & Simplifies software repair into localization and targeted patch generation. & Achieves 27.33\% solve rate on SWE-bench Lite at a lower cost. \\
\midrule
\bottomrule
\end{tabularx}
\end{adjustbox}
\end{table*}

%=========================%
%      Table 2: Groups 5-7
%=========================%
\begin{table*}[htbp]
\centering
\scriptsize
\caption{Training Methodologies (Group 7): Reinforcement Learning Methods}
\label{tab:training_methodologies_2}
\begin{adjustbox}{max width=\textwidth}
\begin{tabularx}{\textwidth}{@{}%
>{\raggedright\arraybackslash}p{3.2cm}  % Work Name
>{\centering\arraybackslash}p{1.1cm}    % Year
>{\raggedright\arraybackslash}p{2.8cm}  % Category
	X % Key Techniques / Ideas
	X % Main Results / Impact
@{}}
\toprule
\textbf{Work Name} & \textbf{Year} & \textbf{Category} & \textbf{Key Techniques / Ideas} & \textbf{Main Results / Impact} \\
\midrule
\multicolumn{5}{l}{\textbf{Group 7: Reinforcement Learning Methods}} \\
\midrule
APO \cite{d2024anchored} & 2024 & RL (APO) & Leverages contrastive learning from AI revisions (via CLAIR) to create highly contrastive preference pairs; defines APO-zero and APO-down objectives for fine-grained control. & Improves Llama-3-8B-Instruct by 7.65\%, closing 45\% of the gap with GPT-4-turbo. \\ \hline
RLVR \cite{lambert2025tulu3pushingfrontiers} & 2025 & RL (Verifiable Rewards) & Replaces traditional reward models with a binary verification function to assign rewards only when responses meet correctness criteria. & Boosts performance on GSM8K and related benchmarks. \\ \hline
RLHF (REINFORCE++) \cite{hu2025reinforce++} & 2025 & RLHF & Combines PPO-style clipping with token-level KL penalties and eliminates the need for a critic network. & Reduces training time by ~30\% while matching or surpassing more complex methods. \\ \hline
RLEF \cite{gehring2024rlef} & 2024 & RL with Execution Feedback & Integrates execution feedback (e.g., code tests) in an iterative loop to refine responses through error correction. & Achieves state-of-the-art results on competitive programming tasks. \\ \hline
SCoRe \cite{kumar2024training} & 2024 & RL Self-Correction & Applies multi-turn RL with a KL constraint to generate and refine intermediate reasoning steps. & Improves MATH benchmark by 15.6\% and HumanEval by 9.1\%. \\ \hline
CGPO \cite{xu2024perfect} & 2024 & Constrained RL & Utilizes a Mixture of Judges (MoJ) and a DPO warm-up phase to enforce multiple constraints during policy optimization. & Yields 7.4–12.5\% improvements across diverse tasks. \\ \hline
Process Reward Models \cite{zhang2025lessons} & 2025 & RL (Process Reward Models) & Supervises intermediate reasoning by combining MC estimation with LLM-as-a-judge evaluations and consensus filtering. & Enhances step-level error identification and overall reasoning reliability. \\ \hline
CLoud \cite{ankner2024critique} & 2024 & RL (Critique Reward Models) & Generates natural language critiques before reward assignment to explicitly capture strengths and weaknesses in responses. & Improves pairwise preference classification accuracy by 4.65–5.84\%. \\ \hline
GenRM \cite{zhang2024generative} & 2024 & RL (Generative Reward Models) & Jointly trains on next-token prediction for both verification and solution generation using a maj@K strategy. & Boosts problem-solving accuracy by 16–64\% on algorithmic and grade-school math tasks. \\ \hline
Meta-Rewarding LM \cite{wu2024meta} & 2024 & RL (Meta-Rewarding) & Allows a single LLM to assume actor, judge, and meta-judge roles, refining its own evaluation via DPO on generated preference pairs. & Raises win rates on AlpacaEval2 (from 22.9\% to 39.4\%) and Arena-Hard (from 20.6\% to 29.1\%). \\ \hline
BOND \cite{sessa2024bond} & 2024 & RLHF & Uses Jeffreys divergence within a distribution-matching framework and a moving anchor model to mimic Best-of-N sampling without high inference cost. & Outperforms traditional RLHF on tasks like abstractive summarization and Gemma model fine-tuning. \\ \hline
SimPO \cite{meng2024simpo} & 2024 & RLHF & Refines DPO by employing an implicit, length-normalized reward (average log probability) with a target reward margin to distinguish winning and losing outputs. & Reduces processing time by 20\%, cuts GPU memory usage by 10\%, and improves win rates on multiple benchmarks. \\ \hline
TPO \cite{wu2024thinking} & 2024 & RL (TPO) & Separates internal chain-of-thought (CoT) from final answers by sampling multiple CoTs and using a judge model to construct preference pairs for iterative DPO training. & Achieves approximately 20\% performance gains on benchmarks like AlpacaEval and Arena-Hard. \\ \hline
GRPO \cite{shao2024deepseekmath} & 2024 & RL (GRPO) & Optimizes policies via groupwise candidate comparisons and unbiased KL divergence estimation without an explicit KL penalty term. & Yields roughly a 5\% improvement on benchmarks such as GSM8K and MATH. \\ \hline
DAPO \cite{yu2025dapoopensourcellmreinforcement} & 2025 & RL (DAPO) & Introduces four key techniques: asymmetric clipping (Clip-Higher) to prevent entropy collapse; dynamic sampling to filter out prompts with 0\% or 100\% accuracy; token-level policy gradient loss to control response length while preserving reasoning quality; and a length-aware penalty to reduce reward noise. & Achieves 50 points on AIME 2024 with 50\% fewer training steps compared to GRPO, thereby setting a new benchmark and enhancing reproducibility in large-scale LLM RL. \\ \hline
DPO \cite{rafailov2024direct} & 2024 & Preference Optimization & Reframes RLHF as a classification problem by directly aligning the policy with human preferences through a Bradley--Terry objective, bypassing explicit RL. & Matches or exceeds the performance of traditional RLHF in controlling response attributes. \\ \hline
iLR-DPO \cite{liu2024iterative} & 2024 & Preference Optimization & Extends DPO by incorporating a length penalty term (balancing preference margin with verbosity) to mitigate excessive response length. & Enables 7B models to achieve performance comparable to GPT-4 with a 50.5\% length-controlled win rate on AlpacaEval 2.0. \\ \hline
LIFT-DPO \cite{yuan2024following} & 2024 & Preference Optimization & Augments instruction-following datasets with explicit length constraints and constructs new preference pairs that penalize verbosity. & Reduces length violations and improves win rates in length-controlled evaluations without degrading standard task performance. \\
\midrule
\bottomrule
\end{tabularx}
\end{adjustbox}
\end{table*}

\begin{table*}[htbp]
\centering
\caption{\textcolor{black}{Summary of Works on Multilingual LLMs and Training Strategies}}
\label{tab:multilingual_llms}
\renewcommand{\arraystretch}{1.2}
\begin{tabular}{p{2cm}|p{1cm}|p{2cm}|p{1.5cm}|p{2.5cm}|p{3cm}|p{3cm}}
\hline
\textcolor{black}{\textbf{Authors}} & \textcolor{black}{\textbf{Year}} & \textcolor{black}{\textbf{Model/Technique}} & \textcolor{black}{\textbf{Params}} & \textcolor{black}{\textbf{Focus Area}} & \textcolor{black}{\textbf{Key Contributions}} & \textcolor{black}{\textbf{Benchmarks/Results}} \\
\hline

\textcolor{black}{Zhu et al.~\cite{zhu2024question}} & \textcolor{black}{2024} & \textcolor{black}{Question Alignment} & \textcolor{black}{LLaMA2-13B} & \textcolor{black}{Multilingual Reasoning} & \textcolor{black}{Aligns X-English question pairs with instruction tuning} & \textcolor{black}{+11.3\% MGSM, +16.1\% MSVAMP} \\
\hline

\textcolor{black}{She et al.~\cite{she2024mapo}} & \textcolor{black}{2024} & \textcolor{black}{MAPO} & \textcolor{black}{N/A} & \textcolor{black}{Reasoning Consistency} & \textcolor{black}{Preference optimization using translated answers} & \textcolor{black}{+16.2\% MSVAMP, +13.3\% MNumGLUESub} \\
\hline

\textcolor{black}{Gao et al.~\cite{gao2024towards}} & \textcolor{black}{2024} & \textcolor{black}{XConST} & \textcolor{black}{N/A} & \textcolor{black}{Cross-lingual Translation} & \textcolor{black}{Prompt-based regularization for zero-shot translation} & \textcolor{black}{Boost on ALMA, Tower, LLaMA-2} \\
\hline

\textcolor{black}{Chen et al.~\cite{chen2024orion}} & \textcolor{black}{2024} & \textcolor{black}{Orion-14B} & \textcolor{black}{14B} & \textcolor{black}{Scalable Multilingual Family} & \textcolor{black}{Scheduled data + fine-tuned variants for dialogue} & \textcolor{black}{SOTA on multilingual benchmarks} \\
\hline

\textcolor{black}{Blevins et al.~\cite{blevins2024breaking}} & \textcolor{black}{2024} & \textcolor{black}{X-ELM} & \textcolor{black}{N/A} & \textcolor{black}{Modular Architecture} & \textcolor{black}{Language-specialized experts in an ensemble} & \textcolor{black}{Outperforms multilingual baseline} \\
\hline

\textcolor{black}{Shliazhko et al.~\cite{shliazhko2024mgpt}} & \textcolor{black}{2024} & \textcolor{black}{mGPT} & \textcolor{black}{GPT-3 scale} & \textcolor{black}{Multilingual GPT Extension} & \textcolor{black}{Pretrained on 61 languages, evaluated on 33} & \textcolor{black}{Strong in-context + probing results} \\
\hline

\textcolor{black}{Chen et al.~\cite{chen2023tigerbot}} & \textcolor{black}{2023} & \textcolor{black}{TigerBot} & \textcolor{black}{7–180B} & \textcolor{black}{Chinese + English LLM} & \textcolor{black}{Open bilingual models w/ top benchmark scores} & \textcolor{black}{+6\% EN, +20\% ZH} \\
\hline

\textcolor{black}{Chen et al.~\cite{chen2023monolingual}} & \textcolor{black}{2023} & \textcolor{black}{Multilingual Alpaca} & \textcolor{black}{N/A} & \textcolor{black}{Instruction Tuning} & \textcolor{black}{Low-rank and full tuning using translated Alpaca data} & \textcolor{black}{Comparable or better than mono models} \\
\hline

\textcolor{black}{Luo et al.~\cite{luo2023yayi}} & \textcolor{black}{2023} & \textcolor{black}{YAYI 2} & \textcolor{black}{30B} & \textcolor{black}{Chinese-Centric Model} & \textcolor{black}{Pretrained from scratch on 2.65T multilingual tokens} & \textcolor{black}{Outperforms open-source models on MMLU} \\
\hline

\textcolor{black}{Yang et al.~\cite{yang2023mono}} & \textcolor{black}{2023} & \textcolor{black}{Hungarian GPT} & \textcolor{black}{6.7B} & \textcolor{black}{Regional Multilingual} & \textcolor{black}{First Hungarian instruction-following model} & \textcolor{black}{Built from Alpaca prompts} \\
\hline

\textcolor{black}{Holmstrom et al.~\cite{holmstrom2023bridging}} & \textcolor{black}{2023} & \textcolor{black}{Cross-lingual Transfer} & \textcolor{black}{N/A} & \textcolor{black}{Low-Resource Adaptation} & \textcolor{black}{Transfer to Swedish tasks from English LLMs} & \textcolor{black}{Cultural + reasoning tasks} \\
\hline

\textcolor{black}{Chung et al.~\cite{chung2023unimax}} & \textcolor{black}{2023} & \textcolor{black}{UniMax} & \textcolor{black}{N/A} & \textcolor{black}{Data Sampling Strategy} & \textcolor{black}{Capped token repetitions across 107 languages} & \textcolor{black}{29T token mC4, better than temp. sampling} \\
\hline

\textcolor{black}{Le et al.~\cite{le2023bloom}} & \textcolor{black}{2023} & \textcolor{black}{BLOOM} & \textcolor{black}{176B} & \textcolor{black}{Open Access LLM} & \textcolor{black}{59-language decoder-only transformer} & \textcolor{black}{Strong multilingual benchmarks} \\
\hline

\textcolor{black}{Li et al.~\cite{li2024eliciting}} & \textcolor{black}{2024} & \textcolor{black}{Instruction-tuned XGLM} & \textcolor{black}{7.5B} & \textcolor{black}{Translation Emergence} & \textcolor{black}{Instruction tuning without parallel corpora} & \textcolor{black}{Generalizes to unseen pairs} \\
\hline

\textcolor{black}{Lin et al.~\cite{lin2022few}} & \textcolor{black}{2022} & \textcolor{black}{Few-shot LLM} & \textcolor{black}{7.5B} & \textcolor{black}{Multilingual Prompting} & \textcolor{black}{Cross-lingual few-shot learning strategies} & \textcolor{black}{+7.4\% zero-shot, +9.4\% 4-shot} \\
\hline

\textcolor{black}{Kale et al.~\cite{kale2021nmt5}} & \textcolor{black}{2021} & \textcolor{black}{n-mT5} & \textcolor{black}{T5-scale} & \textcolor{black}{Parallel Data in Pretraining} & \textcolor{black}{Adds MT task in multilingual pretraining} & \textcolor{black}{Helpful for low-resource LMs} \\
\hline

\textcolor{black}{Xue et al.~\cite{xue2020mt5}} & \textcolor{black}{2020} & \textcolor{black}{mT5} & \textcolor{black}{T5-scale} & \textcolor{black}{Pretraining 101 Languages} & \textcolor{black}{Adapted text-to-text framework on CC100} & \textcolor{black}{Strong zero-shot generalization} \\
\hline

\textcolor{black}{Uthus et al.~\cite{uthus2023mlongt5}} & \textcolor{black}{2023} & \textcolor{black}{mLongT5} & \textcolor{black}{LongT5-scale} & \textcolor{black}{Long Input Processing} & \textcolor{black}{Extends mT5 with UL2 for summarization and QA} & \textcolor{black}{Beats mBART, mBERT} \\
\hline

\end{tabular}
\end{table*}

\section{\textcolor{black}{Multilingual LLM models}}\label{sec:r1} 

\textcolor{black}{The emergence of multilingual large language models (MLLMs) has played a key role in extending the capabilities of traditional LLMs beyond English. While early LLMs focused primarily on English, limiting their use in many regions, MLLMs are designed to understand and generate text in multiple languages, helping bridge linguistic gaps and improve global accessibility. Current research in this area generally falls into two main approaches: one focuses on training models with multilingual data to enhance cross-lingual understanding, while the other explores the use of advanced prompting techniques during inference without modifying the underlying model parameters \cite{qin2025survey}. Table. ~\ref{tab:multilingual_llms} presents a comprehensive overview of recent advancements in multilingual large language models (MLLMs), categorized by their core contributions such as reasoning alignment, instruction tuning, translation capabilities, and architectural innovations. Each entry highlights the model or technique proposed, its parameter scale, targeted multilingual challenge, key methodological insights, and performance on relevant benchmarks.}

\subsection{\textcolor{black}{Multilingual Reasoning and Alignment Strategies}}

\subsubsection{\textcolor{black}{Question Alignment for Enhanced Reasoning}}

\textcolor{black}{Zhu et al. \cite{zhu2024question} propose a novel technique called question alignment to enhance the multilingual reasoning capabilities of large language models (LLMs), which typically underperform in non-English languages due to English-dominant training data. Instead of relying on costly and often inaccurate translation-training methods, they fine-tune models using X-English parallel question pairs to align multilingual inputs with English instruction data. This targeted in-domain alignment enables better utilization of English reasoning capabilities. Their experiments on LLaMA2-13B demonstrate that question alignment significantly outperforms translate-training, yielding average accuracy improvements of 11.3\% on MGSM and 16.1\% on MSVAMP across ten languages.}

\subsubsection{\textcolor{black}{MAPO: Multilingual Preference Optimization}}

\textcolor{black}{She et al. \cite{she2024mapo} propose MAPO (Multilingual-Alignment-as-Preference Optimization), a framework designed to improve the reasoning capabilities of large language models (LLMs) in non-dominant languages by aligning them with the reasoning patterns exhibited in dominant languages like English. Recognizing that multilingual training data imbalances lead to inconsistent reasoning performance, MAPO leverages off-the-shelf translation models to evaluate the consistency between answers in different languages and uses this as a preference signal for optimization through techniques such as Direct Preference Optimization (DPO) or Proximal Policy Optimization (PPO). Experimental results demonstrate that MAPO significantly enhances multilingual reasoning performance across various models and benchmarks, achieving gains of +16.2\% on MSVAMP, +6.1\% on MGSM, and +13.3\% on MNumGLUESub, while also improving cross-lingual consistency in reasoning.}

\subsection{\textcolor{black}{Cross-Lingual Translation and Transfer Techniques}}

\subsubsection{\textcolor{black}{Cross-Lingual Consistency with XConST}}

\textcolor{black}{Gao et al. \cite{gao2024towards} propose enhancing many-to-many multilingual translation in large language models (LLMs), particularly in zero-shot settings, by integrating prompt strategies and a cross-lingual consistency regularization technique called XConST. While XConST is an adaptation of the existing CrossConST method, it is tailored here for translation instruction finetuning. The approach aims to close the representation gap between languages and improve the effectiveness of zero-shot translation. Experimental evaluations on ALMA, Tower, and LLaMA-2 models confirm that the proposed method consistently boosts translation performance across multilingual scenarios.}

\subsubsection{\textcolor{black}{Instruction-Based Emergence of Translation Capabilities}}

\textcolor{black}{Li et al. \cite{li2024eliciting} propose an in-depth investigation into how large language models (LLMs) acquire multilingual translation capabilities without direct training on parallel corpora. By finetuning the XGLM-7.5B model with translation instructions, they reveal that multilingual LLMs are more capable at translation than previously thought. Their analysis shows that a language’s translation performance is influenced by its closeness to English and representation in the pretraining data. Moreover, the ability to follow translation instructions depends on instruction comprehension and cross-lingual alignment. Notably, the study finds that multilingual instruction finetuning enables LLMs to generalize well, even to unseen language pairs.}

\subsubsection{\textcolor{black}{Cross-Lingual Task Transfer for Low-Resource Languages}}

\textcolor{black}{Holmstrom et al. \cite{holmstrom2023bridging} propose applying the “one model–many models” framework, traditionally used for task transfer, to the domain of language transfer, aiming to explore alternatives to language-specific pre-training for smaller languages. Focusing on Swedish as a case study, they evaluate the capabilities of non-Swedish monolingual and multilingual large language models (LLMs) on Swedish NLP tasks. Their results show that LLMs with limited exposure to Swedish can still perform remarkably well by leveraging cross-lingual transfer from English, exhibiting advanced abilities such as mathematical reasoning and culturally informed responses. These findings highlight that effective and resource-efficient strategies exist for supporting smaller languages, reducing the need for full-scale language-specific pre-training.
}

\subsection{\textcolor{black}{Monolingual and Regional Multilingual Models}}

\subsubsection{\textcolor{black}{Hungarian Multilingual Instruction-Tuned Models}}

\textcolor{black}{Yang et al. \cite{yang2023mono} propose the development of large-scale Hungarian-centric language models as a response to the global trend of training increasingly massive Transformer-based models. Rather than competing directly with major tech companies, they focus on leveraging the benefits of knowledge transfer by training a Hungarian monolingual and a Hungarian-English-Chinese trilingual GPT model, each with 6.7 billion parameters, using over 1TB of textual data. To enhance the model’s practical utility, they fine-tuned it using prompt-based instruction tuning from the Stanford Alpaca dataset. The outcome of this process is the creation of an instruction-tuned GPT model, which, to the authors' knowledge, is the first multilingual instruction-following LLM developed in this linguistic region.}

\subsubsection{\textcolor{black}{Chinese-Centric Multilingual Models: YAYI 2}}
\textcolor{black}{Luo et al. \cite{luo2023yayi} propose YAYI 2, a multilingual large language model featuring both base and chat versions, each with 30 billion parameters, specifically designed to address the underperformance of existing open-source LLMs in Chinese-language contexts. Unlike many models primarily optimized for English, YAYI 2 is pretrained from scratch on a multilingual corpus comprising 2.65 trillion tokens, curated through a robust data processing pipeline. The base model is further aligned with human preferences through supervised fine-tuning on millions of instructions and reinforcement learning from human feedback. Experimental results across various benchmarks, including MMLU and CMMLU, demonstrate that YAYI 2 consistently outperforms other open-source models of similar size, highlighting its effectiveness in multilingual and culturally specific tasks.}

\subsection{\textcolor{black}{Instruction Tuning for Multilingual Generalization}}

\subsubsection{\textcolor{black}{Cost-Efficient Multilingual Instruction-Tuning}}

\textcolor{black}{Chen et al. \cite{chen2023monolingual} propose cost-effective strategies for instruction-tuning large language models (LLMs) in multilingual settings, aiming to extend their applicability beyond monolingual use cases such as chat assistants and open-domain question answering. Leveraging the Alpaca dataset and its machine-translated versions, they create multilingual training data to fine-tune LLMs using low-rank adaptation and full-parameter tuning approaches. Through controlled experiments with fixed computational budgets, they demonstrate that multilingual instruction tuning performs as well as or even better than training separate models for each language. Additionally, their results reveal that using downsampled multilingual data can yield models that are not only comparably effective but also more robust. These insights offer practical guidance for scaling LLM capabilities across languages.}

\subsubsection{\textcolor{black}{BLOOM: Democratized Multilingual Instruction Model}}

\textcolor{black}{Le et al. \cite{le2023bloom} propose BLOOM, a 176-billion-parameter open-access language model developed through a large-scale collaboration to democratize access to powerful language technologies typically restricted to resource-rich organizations. BLOOM is a decoder-only Transformer trained on the ROOTS corpus, including data from 46 natural and 13 programming languages, totaling 59 languages. The model demonstrates competitive performance across a broad range of benchmarks, with notable improvements following multitask prompted fine-tuning.}

\subsection{\textcolor{black}{Architectural Innovations and Training Techniques}}

\textcolor{black}{Blevins et al. \cite{blevins2024breaking} introduce Cross-lingual Expert Language Models (X-ELM) to address the performance gap between multilingual and monolingual models, which arises from competition among languages for shared model parameters. X-ELM mitigates this issue by training separate language experts on subsets of a multilingual corpus, allowing each expert to specialize in a specific language while functioning cohesively as part of a multilingual ensemble. Experimental results reveal that, under equal computational constraints, X-ELM consistently outperforms jointly trained multilingual models across all evaluated languages and demonstrates improved transferability to downstream tasks. Beyond performance, X-ELM enables the modular addition of new language experts without catastrophic forgetting and supports asynchronous training, thus lowering hardware requirements and making multilingual model development more accessible.}

\subsubsection{\textcolor{black}{X-ELM: Expert Modular Models}}

\textcolor{black}{Chung et al. \cite{chung2023unimax} present UniMax, a sampling strategy for pretraining multilingual large language models that improves upon traditional temperature-based approaches by ensuring more uniform coverage of high-resource languages while preventing overfitting on low-resource ones. UniMax achieves this by explicitly capping the number of repeats per language corpus. Through comprehensive ablation studies across various model scales and multilingual benchmarks, the authors demonstrate that UniMax consistently outperforms temperature-based sampling, with its advantages becoming more pronounced at larger scales. As part of their contributions, they also release a refreshed mC4 corpus containing 29 trillion characters across 107 languages and a collection of umT5 model checkpoints trained using the UniMax method.}

\subsubsection{\textcolor{black}{Orion-14B: Scalable Multilingual Family}}

\textcolor{black}{Chen et al. \cite{chen2024orion} propose Orion-14B, a suite of multilingual large language models comprising 14 billion parameters, designed to support a wide range of languages and applications. They adopt a data scheduling strategy to train the foundational model on a massive and diverse corpus of 2.5 trillion tokens drawn from English, Chinese, Japanese, Korean, and additional languages. Beyond the base model, they fine-tune specialized variants optimized for conversational and task-specific scenarios. Evaluation results show that Orion-14B delivers state-of-the-art performance across various benchmarks, highlighting its effectiveness in multilingual and domain-adapted settings.}

\subsubsection{\textcolor{black}{TigerBot: Open LLMs with Strong Bilingual Results}}

\textcolor{black}{Chen et al. \cite{chen2023tigerbot} introduce the TigerBot family of large language models, which includes both base and chat variants ranging in size from 7 to 180 billion parameters. Building on architectures like Llama-2 and BLOOM, TigerBot advances the state of the art by improving training data, algorithms, infrastructure, and application tooling. The models demonstrate notable performance gains over leading open-source alternatives, achieving a 6\% improvement in English and a 20\% improvement in Chinese tasks. TigerBot models also secure top rankings across major academic and industry benchmarks. Committed to open development, the authors publicly release their models and share the methodologies behind them, with a focus on democratizing access to cutting-edge LLMs and enabling practical, real-world adoption.}

\subsection{\textcolor{black}{Multilingual LLMs from Pretraining to Transfer}}

\subsubsection{\textcolor{black}{mGPT: Expanding GPT to 61 Languages}}

\textcolor{black}{Shliazhko et al. \cite{shliazhko2024mgpt} propose mGPT, a multilingual extension of GPT-3, pretrained on data from 61 languages spanning 25 linguistically diverse language families using sources such as Wikipedia and the C4 Corpus. They outline the model's architecture and pretraining process, followed by a comprehensive evaluation that includes language modeling across all target languages, performance on cross-lingual natural language understanding (NLU) benchmarks in 33 languages, and knowledge probing tasks in 23 languages. The results show that mGPT achieves in-context learning capabilities comparable to other contemporary LLMs, while significantly expanding language coverage to include underrepresented and low-resource languages, particularly those spoken in the Commonwealth of Independent States and among indigenous communities in Russia.}

\subsubsection{\textcolor{black}{Multilingual Few-Shot LLM}}

\textcolor{black}{Lin et al. \cite{lin2022few} propose training large-scale multilingual generative language models on a diverse corpus encompassing a wide range of languages to improve cross-lingual generalization, addressing the English-centric bias present in models like GPT-3. Focusing on few-shot and zero-shot learning capabilities, they develop a 7.5 billion parameter model that achieves state-of-the-art performance in over 20 languages. Compared to GPT-3 of similar size, their model shows substantial improvements in multilingual commonsense reasoning (+7.4\% in zero-shot and +9.4\% in four-shot settings) and natural language inference (+5.4\% in both settings). On the FLORES-101 benchmark, the model outperforms GPT-3 in 171 out of 182 translation directions with just 32 training examples and exceeds the supervised baseline in 45 directions. The study also provides a detailed analysis of multilingual prompting strategies, demonstrating that effective few-shot performance across languages can be achieved through cross-lingual transfer using both templates and examples.}

\subsubsection{\textcolor{black}{mT5: Scaling T5 to 101 Languages}}

\textcolor{black}{Xue et al. \cite{xue2020mt5} propose mT5, a multilingual extension of the T5 model, which adapts the original text-to-text framework to a broader linguistic context by pretraining on a Common Crawl-based dataset spanning 101 languages. They detail the architectural and training modifications required to scale T5 to the multilingual setting and report state-of-the-art results across multiple multilingual NLP benchmarks. Additionally, they introduce a simple yet effective method to mitigate "accidental translation" in zero-shot scenarios, where the model may inadvertently generate responses in an unintended language.}

\subsubsection{\textcolor{black}{n-mT5: Parallel Data in Pretraining}}

\textcolor{black}{Kale et al. \cite{kale2021nmt5} propose enhancing the pre-training of mT5, which is a massively multilingual extension of T5 by incorporating parallel data to improve performance on multilingual and cross-lingual tasks. Introducing multi-task objectives such as machine translation during the pre-training phase demonstrates that this approach effectively boosts downstream task performance. Their analysis reveals that while the benefits of parallel data diminish as model size increases, it still offers measurable advantages, particularly in scenarios with limited labeled data. This suggests that parallel data remains a valuable component in the pre-training process, especially for low-resource applications and smaller-scale models.}

\subsubsection{\textcolor{black}{mLongT5: Handling Long Sequences}}

\textcolor{black}{Uthus et al. \cite{uthus2023mlongt5} present mLongT5, a multilingual and efficient text-to-text Transformer model designed to handle long input sequences. Building on the LongT5 architecture, mLongT5 integrates multilingual pretraining datasets from mT5 and incorporates the UL2 pretraining objectives to enhance performance across diverse tasks. The model is evaluated on multilingual summarization and question-answering benchmarks, consistently outperforming existing multilingual models such as mBART and M-BERT. These results underscore mLongT5’s effectiveness in managing complex multilingual tasks involving extended textual inputs.}

\begin{table*}[htbp]
\centering
\caption{\textcolor{black}{Overview of Recent Works Based on State Space Models (SSMs)}}
\label{tab:ssm-models}
\renewcommand{\arraystretch}{1.5}
\begin{tabular}{p{1.8cm}|p{0.8cm}|p{2cm}|p{3.2cm}|p{3.2cm}|p{3.2cm}}
\hline
\textcolor{black}{\textbf{Model}} &
\textcolor{black}{\textbf{Year}} &
\textcolor{black}{\textbf{Authors}} &
\textcolor{black}{\textbf{Key Innovation}} &
\textcolor{black}{\textbf{Use Case / Domain}} &
\textcolor{black}{\textbf{Performance / Advantage}} \\
\hline

\textcolor{black}{Cobra} & 
\textcolor{black}{2025} & 
\textcolor{black}{Zhao et al. \cite{zhao2025cobra}} & 
\textcolor{black}{Multi-modal SSM replacing Transformer backbones} & 
\textcolor{black}{Multimodal alignment (text + vision)} & 
\textcolor{black}{3--4$\times$ faster than LLaVA; 48\% parameter fine-tuning} \\
\hline
\textcolor{black}{Llamba} &
\textcolor{black}{2025} &
\textcolor{black}{Bick et al. \cite{bick2025llamba}} &
\textcolor{black}{Distills LLaMA-3.x into Mamba using MOHAWK for efficient recurrent modeling} &
\textcolor{black}{Language modeling on edge/mobile devices} &
\textcolor{black}{High throughput, 0.1\% data use, optimized for smartphones and constrained environments} \\
\hline

\textcolor{black}{MoE-Mamba} & 
\textcolor{black}{2024} & 
\textcolor{black}{Pioro et al. \cite{pioro2024moe}} & 
\textcolor{black}{Combines MoE with Mamba for scalable SSMs} & 
\textcolor{black}{Efficient large-scale language modeling} & 
\textcolor{black}{2.35$\times$ fewer steps than Mamba; fast inference retained} \\
\hline

\textcolor{black}{Jamba} & 
\textcolor{black}{2024} & 
\textcolor{black}{Lieber et al. \cite{lieber2024jamba}} & 
\textcolor{black}{Hybrid Transformer–Mamba MoE architecture} & 
\textcolor{black}{Long-context LLM on limited hardware} & 
\textcolor{black}{Fits on 80GB GPU; 256K token context handling} \\
\hline

\textcolor{black}{SiMBA} & 
\textcolor{black}{2024} & 
\textcolor{black}{Patro et al. \cite{patro2024simba}} & 
\textcolor{black}{Einstein FFT + Mamba for stability in vision} & 
\textcolor{black}{Image and time-series tasks} & 
\textcolor{black}{SOTA on ImageNet and 7 time-series benchmarks} \\
\hline

\textcolor{black}{MambaFormer} & 
\textcolor{black}{2024} & 
\textcolor{black}{Park et al. \cite{park2024can}} & 
\textcolor{black}{Hybrid SSM + Attention for ICL} & 
\textcolor{black}{In-context learning (ICL)} & 
\textcolor{black}{Better ICL on sparse parity; hybrid gains on benchmarks} \\
\hline

\textcolor{black}{Mamba4Rec} & 
\textcolor{black}{2024} & 
\textcolor{black}{Liu et al. \cite{liu2024mamba4rec}} & 
\textcolor{black}{Mamba SSM tailored for sequential recommendation} & 
\textcolor{black}{Recommender systems} & 
\textcolor{black}{Outperforms RNNs and Transformers in recommendation} \\
\hline

\textcolor{black}{ReMamba} & 
\textcolor{black}{2024} & 
\textcolor{black}{Yuan et al. \cite{yuan2024remamba}} & 
\textcolor{black}{Two-stage re-forward to boost long-context Mamba} & 
\textcolor{black}{Long-sequence NLP} & 
\textcolor{black}{+3.2/+1.6 points on LongBench and L-Eval} \\
\hline

\textcolor{black}{Samba} & 
\textcolor{black}{2024} & 
\textcolor{black}{Ren et al. \cite{ren2024samba}} & 
\textcolor{black}{Mamba + SWA hybrid for ultra-long context} & 
\textcolor{black}{Token memory tasks; streaming} & 
\textcolor{black}{3.7$\times$ faster, 1M token context; 256K extrapolation} \\
\hline

\textcolor{black}{Mamba} & 
\textcolor{black}{2023} & 
\textcolor{black}{Gu and Dao \cite{gu2023mamba}} & 
\textcolor{black}{Input-conditioned SSM with parallel recurrent core} & 
\textcolor{black}{Language, genomics, audio} & 
\textcolor{black}{5$\times$ faster than Transformers; matches 2$\times$ size LMs} \\
\hline

\end{tabular}
\end{table*}

\section{\textcolor{black}{State Spaces Models}}\label{sec:r2} 

\textcolor{black}{State Space Models (SSMs) have emerged as a compelling alternative to Transformer-based architectures in large language models, particularly for handling long-sequence data more efficiently. Unlike Transformers, which rely heavily on attention mechanisms and often face scalability issues, SSMs, especially in their modern forms like S4 and Mamba, offer streamlined computation through structured and discretized formulations. Originally designed to represent continuous-time physical systems, SSMs were later adapted to process sequential data across diverse modalities such as text, audio, and images. Over time, their evolution from basic models with limited applicability to sophisticated architectures like Mamba has enabled them to achieve competitive performance in practical domains, including healthcare and media, while maintaining advantages in speed and memory efficiency\cite{wang2024state,lv2025technologies}.}

\textcolor{black}{Table. \ref{tab:ssm-models} presents a comprehensive overview of recent advancements in State Space Model (SSM)-based architectures, highlighting their growing significance as efficient alternatives to Transformer models. Covers a diverse set of models, including Mamba, Samba, ReMamba, and hybrid variants like Jamba and Cobra, detailing their core innovations, application domains, and performance benefits. These models demonstrate the versatility of SSMs in handling long-sequence data across language, vision, recommendation, and multi-modal tasks, while maintaining competitive accuracy and offering substantial improvements in inference speed and memory efficiency.}

\subsection{\textcolor{black}{Mamba model}}
\textcolor{black}{Gu and Dao \cite{gu2023mamba} propose Mamba, a novel sequence modeling architecture that forgoes attention and MLP layers, offering a highly efficient alternative to Transformers for long-context tasks. Recognizing that prior subquadratic-time models such as linear attention, gated convolutions, and structured state space models (SSMs) struggle with content-based reasoning, especially in language tasks, the authors enhance SSMs by making their parameters input-dependent. This enables selective information propagation and forgetting across tokens, improving adaptability to discrete modalities. Despite losing the convolutional efficiency, they introduce a hardware-aware parallel recurrent algorithm to maintain high performance. Integrated into a streamlined architecture, Mamba delivers 5× faster inference than Transformers and scales linearly with sequence length. It achieves state-of-the-art results across multiple domains, including language, audio, and genomics. In language modeling, the Mamba-3B model outperforms Transformers of the same size and matches those twice as large in pretraining and downstream tasks.}

\subsection{\textcolor{black}{Samba model}}
\textcolor{black}{Ren et al. \cite{ren2024samba} propose Samba, a novel hybrid architecture designed to efficiently model sequences with extremely long context lengths by combining the strengths of Mamba a selective State Space Model (SSM) and Sliding Window Attention (SWA). By compressing input sequences into recurrent hidden states while preserving fine-grained recent memory through attention, Samba addresses the limitations of previous models that either suffer from quadratic complexity or weak length generalization. Scaled to 3.8 billion parameters and trained on 3.2 trillion tokens, Samba achieves state-of-the-art performance across multiple benchmarks. It shows strong zero-shot perplexity up to 1 million token contexts and, when fine-tuned on 4K-length sequences, reliably extrapolates to 256K with perfect memory recall on the Passkey Retrieval task. Additionally, Samba outperforms full-attention models in retrieval on the Phonebook task and demonstrates significant efficiency, offering a 3.73× throughput gain for 128K-length prompts and a 3.64× speedup for 64K-token streaming generation, establishing it as a scalable and high-performing linear-time sequence model.}

\subsection{\textcolor{black}{ReMamba model}}
\textcolor{black}{Yuan et al. \cite{yuan2024remamba} propose ReMamba, an enhanced variant of the Mamba architecture aimed at improving long-context comprehension while preserving its inherent efficiency advantages. Although Mamba is known for its strong performance and inference speed in short-context NLP tasks, it struggles with modeling extended sequences compared to transformer-based models. To address this limitation, ReMamba introduces selective compression and adaptation mechanisms through a lightweight two-stage re-forward process, adding minimal inference overhead. Experimental evaluations on the LongBench and L-Eval benchmarks reveal that ReMamba outperforms the original baselines by 3.2 and 1.6 points, respectively, and achieves performance nearly equivalent to transformer models of the same scale, making it a promising solution for efficient long-context reasoning.}

\subsection{\textcolor{black}{Mamba4Rec model}}
\textcolor{black}{Liu et al. \cite{liu2024mamba4rec} propose Mamba4Rec, the first sequential recommendation model to leverage selective State Space Models (SSMs) for efficient and effective user behavior modeling. Aiming to overcome the inference inefficiency of Transformer-based approaches caused by the quadratic complexity of attention mechanisms Mamba4Rec builds upon the Mamba block, a selective SSM equipped with a hardware-aware parallel algorithm. The authors introduce tailored sequential modeling techniques to enhance recommendation accuracy while preserving computational efficiency. Experimental results on public datasets show that Mamba4Rec successfully balances the trade-off between effectiveness and efficiency, outperforming both RNN- and attention-based baselines in recommendation tasks.}

\subsection{\textcolor{black}{MambaFormer model}}
\textcolor{black}{Park et al. \cite{park2024can} propose MambaFormer, a hybrid architecture that combines the efficiency of State Space Models (SSMs), specifically Mamba, with the strengths of attention mechanisms to improve in-context learning (ICL) capabilities in language models. While SSMs like Mamba are designed to reduce the quadratic computational cost of attention by incorporating gating, convolutions, and input-dependent token selection, their ability to perform ICL a key emergent property of modern LLMs has been less studied. Through a comprehensive evaluation, the authors find that Mamba performs on par with Transformers in standard regression-based ICL tasks and even outperforms them in sparse parity learning. However, it underperforms in tasks requiring complex retrieval. To overcome this, MambaFormer integrates attention blocks into the Mamba framework, improving performance across a broader range of ICL benchmarks. The study highlights the potential of hybrid models in bridging the performance gaps and advancing ICL in non-transformer-based architectures.}

\subsection{\textcolor{black}{SiMBA model}}
\textcolor{black}{Patro et al. \cite{patro2024simba} propose SiMBA, a novel architecture that advances State Space Models (SSMs) by addressing the limitations of existing approaches, particularly the instability of Mamba when scaled to large networks in vision tasks. SiMBA integrates the Mamba block for efficient sequence modeling with a new component called Einstein FFT (EinFFT), which enhances channel modeling through eigenvalue-based transformations. This combination allows SiMBA to model long sequences while maintaining stability and computational efficiency. Extensive evaluations across image and time-series benchmarks demonstrate that SiMBA consistently outperforms previous SSMs and narrows the gap with state-of-the-art Transformer models. Notably, SiMBA sets new performance records on ImageNet, transfer learning datasets like Stanford Cars and Flowers, and seven time-series benchmarks, establishing itself as the new leading SSM architecture for both vision and sequential learning tasks.}

\subsection{\textcolor{black}{Jamba model}}
\textcolor{black}{Lieber et al. \cite{lieber2024jamba} propose Jamba, a novel large language model that introduces a hybrid architecture combining Transformer and Mamba layers within a Mixture-of-Experts (MoE) framework. By interleaving Transformer and Mamba blocks, Jamba harnesses the strengths of both architectures Transformer’s generalization and Mamba’s efficiency while selectively incorporating MoE layers to expand model capacity without significantly increasing active parameter usage. This design allows for flexible adaptation to various computational and task-specific constraints. In the implemented configuration, Jamba achieves strong performance while fitting within a single 80GB GPU, offering high throughput and a reduced memory footprint. Experimental results demonstrate state-of-the-art performance on standard benchmarks and robust handling of long-context tasks, with impressive results up to 256K tokens.}

\subsection{\textcolor{black}{MoE-Mamba model}}
\textcolor{black}{Pioro et al. \cite{pioro2024moe} propose MoE-Mamba, a novel architecture that combines State Space Models (SSMs) with the Mixture of Experts (MoE) paradigm to enhance the scalability and performance of SSM-based models. Building upon Mamba a recent SSM known for its strong sequential modeling capabilities and efficient inference MoE-Mamba integrates expert gating mechanisms to increase model capacity while maintaining computational efficiency. The results show that MoE-Mamba not only surpasses the performance of both the original Mamba and Transformer-based MoE models but also achieves equivalent performance to Mamba in 2.35× fewer training steps. Importantly, it retains the inference speed advantages of SSMs, demonstrating the effectiveness of merging MoE with SSM architectures for efficient and scalable language modeling.}

\subsection{\textcolor{black}{Cobra model}}
\textcolor{black}{Zhao et al. \cite{zhao2025cobra} propose Cobra, a multi-modal large language model (MLLM) that replaces the traditional Transformer backbone with a state-space model architecture, specifically leveraging pre-trained Mamba models to improve efficiency and scalability. Aimed at overcoming the quadratic computational complexity of Transformer-based MLLMs, Cobra enables linear-time sequence modeling while maintaining strong multi-modal alignment capabilities. The model integrates various Mamba variants with visual encoders and explores strategies for effectively aligning textual and visual modalities. Experimental results across multiple multi-modal benchmarks reveal that Cobra achieves 3× to 4× faster inference than leading efficient baselines such as LLaVA-Phi and MobileVLM v2. Moreover, by fine-tuning only about 48\% of its parameters, Cobra significantly improves performance over LLaVA, demonstrating the effectiveness of combining state-space modeling with efficient fine-tuning for multi-modal tasks.}

\subsection{\textcolor{black}{Llamba model}}

\textcolor{black}{Bick et al. \cite{bick2025llamba} propose Llamba, a family of efficient recurrent language models that distill the capabilities of Llama-3.x into the Mamba architecture, aiming to deliver high-performance language modeling with improved efficiency. The Llamba series comprising Llamba-1B, Llamba-3B, and Llamba-8B offers significantly higher inference throughput and supports larger batch sizes than Transformer-based models, all while maintaining competitive benchmark performance. Leveraging MOHAWK, a cross-architecture distillation method, the models achieve these results using less than 0.1\% of the typical training data required for models of similar scale. To extend their utility, the authors provide an optimized implementation of Llamba for deployment on resource-constrained environments such as smartphones and edge devices. Llamba represents a compelling balance between speed, memory efficiency, and accuracy, offering a practical alternative to Transformers for accessible and efficient language modeling.}

\section{Training methodologies}\label{sec:4}

This section comprehensively overviews state‑of‑the‑art approaches to improve LLMs and advance their reasoning capabilities. Specifically, the section covers various techniques from general training methods and architectural innovations (including various Mixture-of-Experts and attention modifications) to reinforcement learning strategies and retrieval‑augmented generation methods. It further details advanced chain‑of‑thought approaches, test‑time compute scaling strategies, and other innovative techniques that address challenges such as self‑improvement, efficient inference, and data deduplication. These techniques are instrumental in advancing LLM reasoning by enabling models to process better, synthesize, and apply knowledge in complex, multi-step tasks. 

Table. \ref{tab:training_methodologies_1} presents a comprehensive overview of training methodologies for LLMs, categorizing them into groups that span general training approaches, mixture-of-experts (MoE) and architectural innovations, retrieval-augmented generation (RAG), chain-of-thought and self-improvement techniques, as well as test-time compute scaling and distillation methods. Notably, many of these approaches have been developed and reported between 2024 and 2025, reflecting a period of rapid innovation. By detailing the key techniques and their main impacts, this table provides valuable insights into how contemporary training paradigms are evolving to improve model performance, efficiency, and adaptability across diverse tasks.

Table. \ref{tab:training_methodologies_2} focuses specifically on reinforcement learning (RL) methods applied to LLMs, showcasing a range of techniques from direct preference optimization to complex self-correction and constrained RL strategies. Emphasizing contributions from 2024 to 2025, the table highlights how RL-based approaches have been leveraged to enhance reasoning accuracy, reduce training time, and optimize overall model behavior. Collectively, the methodologies outlined in this table underscore the critical role of reinforcement learning in advancing LLM capabilities and bridging the performance gap with human-like reasoning.

\subsection{Agent Q}

Putta et al. \cite{putta2024agent} presents a novel framework for empowering large language models to perform complex, multi-step reasoning in dynamic, interactive environments. Recognizing the limitations of traditional supervised pre-training and expert demonstration fine-tuning which often lead to compounding errors and limited exploration the proposed method combines guided Monte Carlo Tree Search (MCTS) with a self-critique mechanism and an off-policy Direct Preference Optimization (DPO) algorithm. This integrated approach, dubbed Agent Q, enables LLM agents to learn effectively from successful and unsuccessful trajectories, significantly improving task performance. Validated in a simulated e-commerce environment (WebShop), Agent Q, when applied to a Llama-3 70B model, boosts zero-shot success rates from 18.6\%  to 81.7\% , further reaching 95.4\%  with online search and even surpassing average human performance. These results underscore the potential of this framework to revolutionize autonomous agent decision-making, with applicability extending to various agentic workflows such as code execution and API interactions.

The Agent Q produces a sequence of composite
actions $\{a_t\}$ by considering its history $h_t$ at each time step. It builds upon a ReAct-style agent \cite{yao2023react} but incorporates a dedicated planning step (PlanReact) \cite{liu2023bolaa} along with additional elements. The agent's behavior is decomposed into four main components:

\begin{enumerate}
    \item Planning Step. The agent first generates a \emph{plan action}
    $a_t^{(\mathrm{plan})}$ conditioned on the current history $h_t$ and environment observations. This plan informs subsequent actions.

    \item Capabilities. The agent can execute different 
    environment interactions (e.g., clicking, scrolling, typing). Each action
    $a_t^{(\mathrm{env})}$ is sampled based on the planned intention and the
    environmental context.

    \item Next Action Generation. After forming the plan, the agent produces
    the next \emph{environment action} $a_t^{(\mathrm{env})}$. This action can include
    user-interface interactions or queries intended to progress the
    agent's objective.

    \item Explanation Action. An additional \emph{explanation action}
    $a_t^{(\mathrm{expl})}$ is introduced to provide a rationale for the chosen plan
    and environment actions. This step helps clarify how the agent interprets the
    current situation and why it makes certain decisions.
\end{enumerate}

The probability of each composite action (plan, environment, and explanation) is formally modeled using a log-likelihood expression. For instance, the total log-probability of actions at a given time step $t$ can be decomposed into terms that reflect the contributions of the planning action, the environment action, and the explanation action:

\begin{align*}
\label{eq:composite_action_probability}
\log \pi(a_t \mid h_t) 
&= \log \pi\bigl(a_t^{(\mathrm{plan})} \mid h_t\bigr) \\
&\quad + \log \pi\bigl(a_t^{(\mathrm{env})} \mid h_t, a_t^{(\mathrm{plan})}\bigr) \\
&\quad + \log \pi\bigl(a_t^{(\mathrm{expl})} \mid h_t, a_t^{(\mathrm{plan})}, a_t^{(\mathrm{env})}\bigr)
\end{align*}

This framework thereby extends standard ReAct agents by introducing a clear planning phase and an explanatory step, aiming to improve interpretability and performance in environments requiring sequential decision-making.

\subsection{Low-Rank Adaptation (LoRA)}

Biderman et al. \cite{biderman2024lora} comprehensively compare Low-Rank Adaptation (LoRA) and full finetuning for large language models in programming and mathematics. By conducting experiments on Llama-2 7B and 13B models utilizing both instruction tuning (with roughly 100K prompt-response pairs) and continued pretraining (using approximately 10B unstructured tokens) the study demonstrates that while LoRA, which fine-tunes only low-rank perturbations to selected weight matrices, acts as a strong regularizer by maintaining diverse generations and preserving performance on tasks outside the target domain, it substantially underperforms full finetuning on the target domain tasks such as those measured by HumanEval and GSM8K. The results further reveal that full finetuning learns perturbations with a rank 10 to 100 times greater than typical LoRA configurations, which may explain its superior accuracy and sample efficiency in specialized tasks.

LoRA (Low-Rank Adaptation) is a technique designed to reduce the memory and compute
requirements for fine-tuning large pretrained models. Instead of modifying the full weight matrix $W_{\text{pretrained}} \in \mathbb{R}^{d \times k}$, LoRA learns a
low-rank update $\Delta$ such that:
\begin{equation}
    W_{\text{finetuned}} = W_{\text{pretrained}} + \Delta, \quad
    \Delta = \gamma \, A \, B,
\end{equation}
where $A \in \mathbb{R}^{d \times r}$ and $B \in \mathbb{R}^{r \times k}$ for a small
rank $r \ll d,k$. The scalar $\gamma$ typically takes the form $\alpha / r$, where
$\alpha$ is a hyperparameter. Common practice initializes $A$ and $B$ to zero and
applies LoRA only to selected target modules. By restricting updates to low-rank factors, the method greatly decreases both the parameter count and the computational
overhead compared to full fine-tuning. For example, applying LoRA to a 7 B-parameter model (with $d=k=4096$) can reduce the trained parameter portion to approximately 1\%
of the original size.

\subsection{Cognitive Architectures for Language Agents (CoALA)}

Sumers et al. \cite{sumers2023cognitive} introduced Cognitive Architectures for Language Agents (CoALA), a structured framework that integrates insights from cognitive science and symbolic AI with contemporary large language models to enhance the capabilities of language agents. CoALA posits that by endowing language agents with modular memory systems including working, long-term, and procedural memory and a clearly defined action space for both internal (e.g., reasoning and memory updates) and external (e.g., tool or API interactions) operations, these agents can more effectively manage grounding and reasoning tasks. The framework further incorporates a generalized decision-making process that involves proposing, evaluating, and selecting actions, ensuring that the agent's behavior aligns with its objectives. This structured approach organizes a broad array of recent developments in language agents and lays out actionable directions for future research, ultimately contextualizing modern language agents within the rich tradition of cognitive architectures and marking a promising path toward more versatile, general intelligence in language-based AI systems.

\subsection{Learn-by-interact}

Learn-by-interact \cite{su2025learn} is a novel data-centric framework designed to enhance the adaptability of large language model (LLM) agents across diverse digital environments without the need for human annotations. The framework synthesizes realistic agent-environment interaction trajectories by leveraging documentation such as manuals, API references, and tutorials. A key innovation in this process is "backward construction," which involves summarizing or abstracting interaction histories to generate clear, aligned instructions from observed behaviors. The authors validate the effectiveness of this synthetic data in both fine-tuning and in-context learning settings, achieving notable improvements: up to 12.2\% enhancement for in-context learning with Claude-3.5 and 19.5\% for fine-tuning with Codestral-22B. Extensive experiments conducted on benchmarks like SWE-bench, WebArena, OSWorld, and Spider2-V across coding, web, and desktop domains underscore the robustness of the approach. Ablation studies further confirm the critical contribution of backward construction and the superiority of their retrieval pipeline over traditional methods such as retrieval-augmented generation. Overall, Learn-by-interact offers a scalable solution for synthesizing high-quality agent data, paving the way for more capable and adaptive UI agents in real-world applications.

\subsection{Supervised Fine-tuning (SFT)}

Chu et al. \cite{chu2025sft} presented a systematic study comparing supervised fine-tuning (SFT) and reinforcement learning (RL) in terms of their impact on model generalization and memorization, particularly within both textual rule-based and visual domains. Using the arithmetic reasoning card game GeneralPoints and the real-world navigation environment V-IRL as testbeds, the study evaluates the performance of a Llama-3.2-Vision-11B model under different training regimes: SFT-only, RL-only, and a hybrid SFT→RL pipeline. The results reveal that while SFT tends to memorize the training data, leading to strong in-distribution performance, but poor out-of-distribution generalization, RL, mainly when guided by outcome-based rewards, effectively generalizes across both rule-based textual and visual tasks. Notably, RL enhances generalization and improves the model's visual recognition capabilities. However, the findings also emphasize that SFT remains a critical precursor to RL, as it stabilizes the model’s output format and ensures that the backbone model adheres to instructions, enabling RL to realize its performance gains fully. Additional experiments with multiple verification iterations (or "Reject Sampling") further demonstrate that this hybrid approach can boost generalization by up to approximately 6\%.

\subsection{Mixture of Experts (MoE)}
MoE \cite{jiang2024mixtral} comprises multiple "expert" neural networks alongside a router that probabilistically directs incoming tokens to the most appropriate expert. Unlike traditional models, where every parameter is active for each inference, MoE leverages its router to selectively activate a subset of experts, enabling the model to encompass many parameters while keeping inference computationally efficient. This selective activation is governed by carefully managed load balancing, ensuring that all experts contribute effectively during training. The approach is versatile, extending beyond language models to vision applications, as exemplified by Mixtral 8x7B, which incorporates a staggering 46.7 billion parameters but only engages 12.8 billion during inference. This design paradigm underscores the potential of MoE to scale model capacity significantly without proportionally increasing the computational burden during deployment \cite{deepseekai2025, liu2024deepseek,puigcerver2023sparse,riquelme2021scaling}.

A Sparse Mixture of Experts (MoE) layer is a mechanism that routes different input tokens to specialized \emph{experts} based on the output of a \emph{gating network}. Suppose there are $n$ experts $\{E_0, E_1, \dots, E_{n-1}\}$. For an input $x$, the MoE layer output is computed by combining each expert's contribution according to the gating values:

\begin{equation}
\sum_{i=0}^{n-1} g_i(x) \cdot E_i(x),
\end{equation}

where $g_i(x)$ denotes the gating value for expert $E_i$ on input $x$, and $E_i(x)$ is the expert's output. When the gating network is \emph{sparse}, only the top-$k$ experts receive nonzero gating weights, significantly reducing computational cost.

A popular approach is to select the top-$k$ gating values per token, commonly referred to as \emph{TopK} gating. This method directs each token to only the $k$ most relevant experts, thereby lowering the overall computation and enabling larger-scale models within the same hardware constraints. The capacity of the MoE layer can be scaled by increasing the total number of experts, while only a subset of them processes each token.

MoE layers are typically distributed across multiple GPUs or devices to handle large numbers of experts. Techniques such as \emph{Expert Parallelism} (EP) split experts among different devices. In a Transformer model, each MoE layer is placed within a feed-forward block (or a sub-block, depending on the architecture), and the gating network determines how tokens are routed to the experts. Frameworks like GShard implement \emph{Top2} gating and use \emph{Switch} or \emph{SwiGLU} operations to process each token's expert assignments in a distributed fashion. This significantly allows for efficient training and inference when scaling to large models \cite{lepikhin2020gshard,shazeer2017outrageously}.

\subsubsection{Heterogeneous Mixture of Experts (HMoE)}

Wang et al. \cite{wang2024hmoe} presents a novel approach to the Mixture of Experts (MoE) paradigm by introducing a Heterogeneous Mixture of Experts (HMoE) framework, which departs from traditional homogeneous designs by employing experts of varying sizes. The key insight is that differing token complexities in natural language processing tasks benefit from experts with specialized capacities rather than a one-size-fits-all model. To this end, the authors propose a unique training objective leveraging a penalty mechanism that preferentially activates smaller experts to improve computational efficiency and parameter utilization. Experimental evaluations reveal that HMoE not only achieves a lower loss but also requires fewer activated parameters than conventional MoE models, demonstrating superior performance across a range of pre-training benchmarks. This innovative method underscores the potential for more adaptive expert allocation strategies in large-scale language modeling.

\subsubsection{Expert-Specialized Fine-Tuning (ESFT)}

Tang et al. \cite{tang2024understanding} investigate parameter-efficient fine-tuning (PEFT) methods for LLMs with sparse architectures, explicitly focusing on Mixture-of-Experts (MoE) models. While most PEFT research has centered on dense-architecture models, this work explores the unique challenges and opportunities in fine-tuning MoE models. The authors first analyze the dispersion of activated experts across tasks, revealing that the routing distribution is highly concentrated for individual tasks but varies significantly across different tasks. Building on this observation, they introduce Expert-Specialized Fine-Tuning (ESFT), a method that selectively tunes the most relevant experts for a given downstream task while freezing the others. Experimental results demonstrate that ESFT not only improves tuning efficiency reducing memory usage by up to 90\% (from 28.6GB to as low as 2.57GB for Token and 3.20GB for Gate ESFT) and training time by up to 30\% (from 28.5 minutes to 19.8 minutes) but also achieves performance comparable to or even surpassing full-parameter fine-tuning and outperforms LoRA by up to 10\%. Moreover, the study highlights that MoE models with finer-grained experts are better at selecting task-relevant combinations, whereas the approach is less practical for coarse-grained MoEs like Mixtral. These findings underscore the benefits of task-specific expert specialization in PEFT for MoE-based LLMs and provide valuable insights into optimizing resource usage and training efficiency.

\subsubsection{Mixture-of-Agents}

Wang et al. \cite{wang2024mixture} propose a novel Mixture-of-Agents (MoA) approach to harness the collective strengths of multiple LLMs for enhanced natural language generation and understanding. In the MoA framework, a layered architecture is employed where each layer consists of several LLM agents that take the aggregated outputs from the previous layer as input, thereby iteratively refining the final response. This method leverages different models' diversity and complementary capabilities, enabling them to generate more accurate and contextually rich outputs. Empirical evaluations on benchmarks such as AlpacaEval 2.0, MT-Bench, and FLASK demonstrate that MoA models achieve state-of-the-art performance for instance, a score of 65.1\% on AlpacaEval 2.0 compared to 57.5\% by GPT-4 Omni. Furthermore, a variant called MoA-Lite outperforms GPT-4 Turbo by 4\% while being twice as cost-efficient, although the layered aggregation process can lead to increased latency in generating the first token.

\subsubsection{Multimodal Mixture of Experts}

Li et al. \cite{li2025uni} introduce Uni-MoE, a unified multimodal large language model that leverages a sparse Mixture-of-Experts (MoE) architecture to efficiently process a wide array of modalities, including audio, speech, image, text, and video, within a single framework. Unlike traditional MoE approaches that employ homogeneous experts and are limited in modality coverage, Uni-MoE utilizes modality-specific encoders with connectors to build a unified multimodal representation, enabling both modality-level data parallelism and expert-level model parallelism. The proposed model is trained using a progressive, three-phase strategy: first, cross-modality alignment is achieved via various connectors; second, modality-specific experts are trained with cross-modality instruction data to activate expert preferences; and finally, the system is fine-tuned using Low-Rank Adaptation (LoRA) on mixed multimodal instruction data. Extensive evaluations across ten vision and audio tasks demonstrate that Uni-MoE matches or outperforms existing unified multimodal models, significantly reducing performance bias in handling diverse and mixed datasets.

\subsection{Multihead Latent Attention (MLA)}

Meng et al. \cite{meng2025transmla} address the communication bottlenecks in modern LLMs by introducing Multi-head Latent Attention (MLA). This innovative mechanism uses low-rank matrices in the key-value layers to compress latent states, thereby reducing the KV cache size and speeding up inference. MLA employs an up-projection matrix to preserve expressiveness, effectively trading additional computation for reduced communication overhead. The authors also demonstrate that while existing models typically use Group Query Attention (GQA), any GQA can be represented with MLA, though not vice versa, highlighting MLA's superior flexibility. To facilitate broader adoption, the paper introduces TransMLA, a post-training conversion method that transforms widely used GQA-based pre-trained models (e.g., LLaMA, Qwen, and Mixtral) into MLA-based models, allowing further training to boost expressiveness without incurring extra cache costs. 

Multi-Head Latent Attention (MLA) modifies standard Multi-Head Attention (MHA) and
Generalized Query Attention (GQA) by allowing each head to use its own latent vectors
while removing RoPE (rotary positional embeddings). The queries are not compressed,
but the keys and values can be reduced to a smaller latent dimension \(r\). Specifically,
let:
\[
Q = X\,W^Q,\quad
K = X\,W^K,\quad
V = X\,W^V,
\]
where \(X \in \mathbb{R}^{T \times D}\), \(Q,K,V \in \mathbb{R}^{T \times (n_h \cdot d_h)}\),
\(n_h\) is the number of attention heads, and \(d_h\) is the dimension per head.

MLA then splits \(Q, K, V\) into \(n_h\) heads. Each head \(i\) computes attention as:
\[
O_i = \mathrm{softmax}\!\Bigl(\frac{Q_i K_i^{\top}}{\sqrt{d_h}}\Bigr)\,V_i,
\]
and the outputs are combined:
\[
O = \sum_{i=1}^{n_h} O_i \in \mathbb{R}^{T \times D}.
\]

To reduce memory usage, MLA stores only the intermediate latent representation for the keys and values, denoted \(K^r\) and \(V^r\), where \(r \ll d_h\). During inference, an \emph{absorb} operator merges certain weight matrices (e.g., \(W^K\) into \(W^Q\)) to avoid increasing the size of the key-value latent dimension. As a result, MLA can be interpreted as a multi-query style attention (MQA) where each head operates over a compressed dimension \(r\).

\subsection{Multimodal Rotary Position Embedding (M-RoPE)}

Wang et al. \cite{wang2024} propose the Multimodal Rotary Position Embedding (M-RoPE), which is an innovative extension of the traditional 1D rotary position embedding designed to encode positional information across diverse modalities effectively. By decomposing the original rotary embedding into three distinct components temporal, height, and width M-RoPE adapts to the unique spatial and temporal structures inherent in text, images, and videos. Textual data functions identically to conventional 1D-RoPE by using uniform position IDs, whereas images assign separate height and width identifiers while maintaining a constant temporal ID. In the case of videos, the temporal component increments with each frame, while the height and width follow the same schema as images. Additionally, when handling multiple modalities within a single input, M-RoPE sequentially increments the maximum position ID of the preceding modality to initialize the next, ensuring coherent integration across channels. This design enhances the model's ability to capture and generalize positional information and reduces the numerical range of position IDs, thereby enabling efficient extrapolation to longer sequences during inference.

\subsection{Multi-Token Prediction (MTP)}

DeepSeek-R1 is a next-generation language model architecture that significantly accelerates inference by employing Multi-Token Prediction (MTP) \cite{deepseekai2025} to generate multiple tokens in parallel. The design incorporates cross-depth residual connections, enabling deeper MTP layers to leverage features from earlier ones. It also features adaptive prediction granularity that dynamically adjusts the number of tokens predicted based on the complexity of the input. Furthermore, DeepSeek-R1 employs depth-aware loss weighting using a sigmoid-based function to prioritize updates at mid-range depths, and it uses memory-efficient parameter sharing through depth-conditioned routing to reduce redundancy across transformer layers. Optimized speculative decoding, which relies on probabilistic agreement checking rather than exact matches, further speeds up inference. These architectural innovations result in substantial empirical improvements, with a 22\% faster training convergence, a 1.5× increase in generation speed, and an 18\% enhancement in long-form perplexity over its predecessor, DeepSeek-V3, demonstrating its superiority in training and inference efficiency.

\subsection{Reinforcement Learning}

Reinforcement Learning (RL) \cite{cheng2023impact} for LLMs is a training approach where the model learns to improve its outputs based on feedback rather than solely relying on static labeled data. In this setting, the LLM acts as an agent that generates responses in reaction to prompts and then receives a reward signal derived either from human feedback or automated metrics that reflects the quality and relevance of its output \cite{lim2023reinforcement}. 

\subsubsection{Anchored Preference Optimization (APO)}

D'Oosterlinck et al. \cite{d2024anchored} explore advanced alignment techniques for large language models by leveraging contrastive learning and preference optimization. The authors identify that more contrastive preference pairs provide a stronger learning signal and that alignment objectives with greater control yield better performance. To capitalize on these insights, they introduce Contrastive Learning from AI Revisions (CLAIR)---a synthetic data generation method that produces minimally revised, highly contrastive preference pairs using a stronger model (e.g., GPT-4-turbo)---and Anchored Preference Optimization (APO), an RLHF approach that fine-tunes model output probabilities with nuanced control over winning and losing responses. Evaluated on MixEval-Hard, their best model, trained on 32K CLAIR preference pairs with APO, improves Llama-3-8B-Instruct's performance by 7.65\%, closing 45\% of the gap with GPT-4-turbo, and demonstrates superior results compared to traditional discriminative reward models. Their approach also highlights the benefits of on-policy data and controlled optimization (using RMSProp with a learning rate of $2\times10^{-7}$ over 18 epochs), emphasizing that the contrastiveness of preference pairs is a major driver of these improvements.

Anchored Preference Optimization (APO) is a family of alignment objectives that offers
fine-grained control over the likelihood of winning and losing outputs during training.
Let \(r_\theta(x,y)\) be a learned reward function and \(\sigma(\cdot)\) a chosen
transformation (for instance, a logistic sigmoid). Two key variants are defined: APO-zero and APO-down.

APO-zero increases the likelihood of winning outputs while limiting the likelihood of losing outputs:
\[
\mathcal{L}_{\mathrm{zero}}^{\mathrm{APO}}(x, y_w, y_l; \theta)
\;=\;
-\,\sigma\bigl(r_\theta(x, y_w)\bigr)
\;+\;
\sigma\bigl(r_\theta(x, y_l)\bigr).
\]

When \(r_\theta(x, y_w) > r_\theta(x, y_l)\), the winning output is favored, while the losing output is constrained from becoming too likely.

APO-down places additional emphasis on reducing the likelihood of losing outputs:

\[
\mathcal{L}_{\mathrm{down}}^{\mathrm{APO}}(x, y_w, y_l; \theta)
\;=\;
-\,\sigma\bigl(r_\theta(x, y_w)\bigr)
\;-\;
\sigma\bigl(r_\theta(x, y_l)\bigr).
\]

Here, winning and losing terms contribute negative values, pushing the losing output's likelihood even lower.

These objectives can be viewed as imposing constraints on how the model's likelihood of winning and losing outputs shifts relative to a reference model. A link to Kullback--Leibler (KL) divergence is established, offering fine-grained control over the final policy. In practice, these objectives allow for careful reward shaping to ensure that improvements in model behavior align with specific alignment criteria.

\subsubsection{Reinforcement Learning with Verifiable Rewards (RLVR)}

Lambert et al. \cite{lambert2025tulu3pushingfrontiers} propose Reinforcement Learning with Verifiable Rewards (RLVR) in the Tülu 3 model, which is a novel approach that redefines alignment for language models on tasks with verifiable outcomes, such as mathematical problem-solving and instruction following. RLVR builds on traditional RLHF by replacing the typical reward model with a verification function that provides a binary signal rewarding the model only when its generated response is objectively correct. This simple yet effective strategy leverages answer matching or constraint verification to offer clear, targeted feedback, thereby streamlining the training process. RLVR improves performance on specialized benchmarks like GSM8K and maintains robustness across diverse tasks when integrated into a generalist training pipeline. By extending previous self-improvement and RL with execution feedback methods, RLVR demonstrates that reinforcing only verifiably correct outputs can lead to significant gains in model reliability and precision.

Reinforcement Learning with Verifiable Rewards (RLVR) introduces a reward mechanism that only grants positive feedback when specific responses are correct. Let \(r(x,y)\) be a verifiability function that returns one if the response \(y\) satisfies a given correctness criterion for input \(x\), and 0 otherwise. The policy \(\pi_\theta\) is then trained to maximize the expected verifiable reward:

\begin{align*}
\max_{\pi_\theta}
\mathbb{E}_{(x,y)\sim c}\Bigl[R_{\mathrm{RLVR}}(x,y)\Bigr]
&=
\max_{\pi_\theta}
\mathbb{E}_{(x,y)\sim c}\Bigl[
r(x,y) \\
- \beta\,\mathrm{KL}\bigl(\pi_\theta(a\mid x)\,\|\,
\pi_{\mathrm{ref}}(a\mid x)\bigr)
\Bigr],
\end{align*}

where \(c\) is a distribution over input-output pairs, and \(\mathrm{KL}\) denotes the Kullback--Leibler divergence from a reference policy \(\pi_{\mathrm{ref}}\). The term \(\beta\) controls the trade-off between maximizing the verifiable reward and staying close to the reference policy.

The training process uses standard reinforcement learning algorithms (e.g., PPO) to adjust \(\pi_\theta\). Whenever the policy generates an output \(y\) that passes the verification check, a reward of 1 is assigned, and otherwise, 0. This setup encourages the policy to produce correct completions, as only verifiable responses contribute to the return.

\subsubsection{Reinforcement Learning from Human Feedback (RLHF)}

Reinforcement Learning from Human Feedback (RLHF) is an approach that leverages human judgments to shape the reward function used in training reinforcement learning agents. In the work by Christiano et al. \cite{christiano2017deep}, the reward model \(r_\theta(x,y)\) is trained using human comparisons. For each prompt \(x\), two responses are provided: one preferred \(y_w\) and one less preferred \(y_l\). The training objective ensures the model assigns a higher score to \(y_w\) than to \(y_l\). This is achieved by minimizing the loss function

\[
L(\theta) = -\mathbb{E}_{(x, y_w, y_l) \sim D} \left[ \ln \sigma\Big( r_\theta(x,y_w) - r_\theta(x,y_l) \Big) \right],
\]

where \(\sigma(\cdot)\) is the sigmoid function. This formulation encourages the reward model to reflect human preferences accurately, providing a more reliable learning signal for policy optimization.

In a subsequent extension of this methodology, Ouyang et al. \cite{ouyang2022training} applied RLHF to fine-tune large language models. Here, the language model’s policy \(\pi_\phi(y \mid x)\) is adapted using a reward model while constraining the updated policy to remain close to a baseline obtained through supervised fine-tuning, denoted as \(\pi_{\text{SFT}}(y \mid x)\). The overall objective for updating the policy is given by

\[
J(\phi) = \mathbb{E}_{x \sim D,\, y \sim \pi_\phi(\cdot \mid x)} \left[ r_\theta(x,y) - \beta \ln \frac{\pi_\phi(y \mid x)}{\pi_{\text{SFT}}(y \mid x)} \right],
\]

Where \(\beta\) is a hyperparameter that moderates the penalty imposed by the Kullback–Leibler divergence between the updated and the baseline policy. This strategy refines the language model’s responses to align more closely with human expectations while preserving the core competencies learned during supervised training.

Hu et al. \cite{hu2025reinforce++}  introduce REINFORCE++, an enhanced variant of the classical REINFORCE algorithm designed to streamline Reinforcement Learning from Human Feedback (RLHF) for large language models. By integrating optimization techniques inspired by Proximal Policy Optimization (PPO) while eliminating the need for a critical network, REINFORCE++ achieves a more straightforward training pipeline with improved stability and reduced computational overhead. Notably, the algorithm incorporates token-level KL penalties to mitigate divergence at a granular level. It employs a PPO-style clipping mechanism to ensure stable policy updates without the complexity of maintaining a separate value function. Empirical evaluations demonstrate that REINFORCE++ reduces training time by approximately 30\%  compared to PPO and holds, or even surpasses, the performance of more complex methods such as GRPO, particularly in tasks like mathematical reasoning. These results underscore the potential of REINFORCE++ to provide more efficient and stable RLHF training, paving the way for more cost-effective and robust alignment of language models with human preferences.

\subsubsection{Reinforcement learning with execution feedback (RLEF)}

Meta team \cite{gehring2024rlef} presented a novel reinforcement learning approach that integrates execution feedback for iterative code synthesis, addressing a critical limitation in current LLM-based agents which typically rely on independent sampling. The proposed method termed Reinforcement Learning with Execution Feedback (RLEF), trains models to optimize not only for first-attempt success but also for error correction through a cyclical process: the model generates code in response to an instruction, the code is executed and evaluated against public test cases, and the resulting feedback is used to guide subsequent corrections until either the public tests are passed or a predefined limit is reached. A final reward signal, derived from private test cases, is then used in a PPO framework to further refine performance. Experimental benchmarks on competitive programming tasks demonstrate that this method yields new state-of-the-art results for both small (8B parameter) and significant (70B parameter) models, significantly reducing the number of samples required by an order of magnitude. Notably, applying RLEF to Llama 3.1 70B improved its performance on CodeContests from 27.5\% to 40.1\% with up to three iterative corrections. It even outperformed state-of-the-art models like AlphaCodium that leverage GPT-4.

Let's consider a two-stage Reinforcement Learning from Human Feedback (RLHF) framework.
In the first stage, a reward model is learned from pairwise preference data; in the
second stage, a policy is optimized to maximize this learned reward while remaining
close to a reference policy.

\paragraph{Learning the Reward Model.}
Suppose each context-action pair \((x,y)\) is assigned a scalar reward \(r(x,y)\).
Let \(y\succ y'\) denote a preference for \(y\) over \(y'\) given the same context
\(x\). A Bradley--Terry style approach models the probability that \(y\) is preferred
as:
\begin{align*}
p(y \succ y') 
&= 
\sigma\bigl(r(x,y) - r(x,y')\bigr),
\end{align*}
where \(\sigma(\cdot)\) is the sigmoid function. Let
\(\{(x_i, y_i, y_i')\}_{i=1}^N\)
be a dataset of pairwise preferences, where \(y_i\) is the preferred action over
\(y_i'\). The reward function is trained by minimizing the negative log-likelihood
of these comparisons:
\begin{align*}
\mathcal{L}(r)
&=
-\,\mathbb{E}_{(x,\,y,\,y') \,\sim\,D}
\Bigl[
\log\Bigl(\sigma\bigl(r(x,y) - r(x,y')\bigr)\Bigr)
\Bigr].
\end{align*}
Actions that are consistently preferred are assigned higher reward values, while
less preferred actions receive lower values.

\paragraph{Policy Optimization with the Learned Reward.}
Once \(r(x,y)\) is learned, the next step is to optimize a policy \(\pi\) that
maximizes the expected reward while remaining close to a reference policy
\(\pi_{\mathrm{ref}}\). One can introduce a KL-regularized objective:
\begin{align*}
J(\pi)
&=
\mathbb{E}_{x \sim \rho}
\Bigl[r\bigl(x,y\bigr)\Bigr]
\;-\;
\tau\,D_{\mathrm{KL}}\!\Bigl(\pi(\cdot\mid x)\,\bigl\|\,
\pi_{\mathrm{ref}}(\cdot\mid x)\Bigr),
\end{align*}
where \(x\) is drawn from a context distribution \(\rho\), \(y\) is sampled from
\(\pi(\cdot\mid x)\), and \(\tau\) balances the trade-off between maximizing
the reward and remaining close to \(\pi_{\mathrm{ref}}\). In practice, algorithms
such as PPO approximate this optimization by iteratively updating \(\pi\) to
increase reward while constraining KL divergence from \(\pi_{\mathrm{ref}}\).

\subsubsection{Self-Correction via Reinforcement Learning}

SCoRe (Self-Correction via Reinforcement Learning) \cite{kumar2024training} is a novel multi-turn online RL approach that dramatically enhances the self-correction abilities of large language models using entirely self-generated data, addressing the shortcomings of traditional supervised fine-tuning methods that often suffer from distribution mismatches and ineffective correction behaviors. By employing a two-stage process first using REINFORCE with a KL-divergence constraint to generate high-reward revisions while preserving the original response, and then lifting this restriction to allow full correction guided by a shaped reward that prioritizes changes from incorrect to correct answers SCoRe enables models like Gemini 1.0 Pro and 1.5 Flash to significantly improve their performance, achieving a 15.6\% boost on the MATH benchmark and a 9.1\% improvement on HumanEval. Key insights from this approach include the critical role of on-policy sampling in multi-turn self-correction, the superior performance of REINFORCE over alternative methods like STaR, and further gains (an additional 10.5\% improvement) when combined with inference-time scaling techniques, ultimately demonstrating that iterative, reinforcement-based self-correction allows models to learn from their mistakes and enhance complex reasoning capabilities.

\subsubsection{Constrained Generative Policy Optimization (CGPO) }

Meta team \cite{xu2024perfect} propose Constrained Generative Policy Optimization (CGPO), a novel post-training paradigm aimed at addressing key limitations of Reinforcement Learning from Human Feedback (RLHF) in multi-task learning scenarios. Traditional RLHF methods such as PPO and DPO often encounter challenges with reward hacking and balancing multiple, sometimes conflicting, objectives, typically requiring extensive hyper-parameter tuning guided by human intuition. In contrast, CGPO leverages a Mixture of Judges (MoJ) approach that combines rule-based and LLM-based evaluators to enforce constraints in a principled manner, ensuring that the resulting policy reaches a pareto-optimal trade-off across a broad array of objectives. The methodology incorporates a DPO warm-up phase and then applies constrained policy optimization via stratification, effectively mitigating reward hacking while enhancing performance. Empirical results demonstrate that CGPO significantly outperforms standard RLHF algorithms, achieving improvements of 7.4\% on AlpacaEval-2 for general chat tasks, 12.5\% on Arena-Hard for STEM and reasoning tasks, and consistent gains in domains such as math and coding. Although the approach demands higher computational resources due to the involvement of multiple reward models and judges, it offers strong theoretical guarantees and a plug-and-play solution for aligning general-purpose LLMs across diverse applications.

Constrained Generative Policy Optimization (CGPO) aims to mitigate reward hacking by
enforcing multiple constraints alongside the reward objective. In a single-task,
single-objective setting, CGPO considers a large language model (LLM) that must
satisfy constraints of the form \(\{C_1, C_2, \ldots, C_M\}\), and a reward function
\(r\). The state-action space is restricted to the feasible region that meets all
constraints \(\{C_i\}\). Formally, the problem is:

\[
\max_\pi
\quad
\mathbb{E}_{\pi}\bigl[r(x,y)\bigr]
\quad
\text{subject to}
\quad
\mathrm{Prob}_{\pi}\bigl(C_i(x,y) \le \varepsilon_i\bigr),
\]

where \(\pi\) is the policy to be optimized, and each constraint \(C_i\) imposes a
requirement such as a safety or quality threshold. Additional terms, such as a KL
limit \(\mathrm{KL}(\pi,\pi_{\mathrm{init}}) \le \kappa\), may be included to keep
the policy close to an initialization \(\pi_{\mathrm{init}}\). By integrating these
constraints, CGPO encourages the model to avoid undesired behaviors and better manage
the trade-off between maximizing the reward and adhering to the specified constraints.

\subsubsection{Process Reward Models}

Zhang et al.  \cite{zhang2025lessons} investigated Process Reward Models (PRMs) as a method for supervising the intermediate reasoning steps in mathematical problem-solving with LLMs. The authors highlight the inherent challenges in developing effective PRMs, particularly regarding data annotation and evaluation. Their experiments reveal that traditional Monte Carlo (MC) estimation-based data synthesis, which uses a completion model to assess the correctness of each reasoning step, often underperforms compared to approaches that incorporate LLM-as-a-judge and human annotations. Specifically, the study identifies that MC estimation tends to produce inaccurate step verification and that conventional Best-of-N (BoN) evaluation strategies introduce biases by over-rewarding outcomes even when the underlying reasoning process is flawed. The authors propose a consensus filtering mechanism that integrates MC estimation and LLM-as-a-judge evaluations to mitigate these issues. This ensures that only steps with consistent labels are retained for training. This combined approach enhances the reliability and accuracy of step-level error identification and significantly improves overall model performance and data efficiency. The paper culminates in releasing state-of-the-art PRMs in 7B and 72B parameter configurations, demonstrating superior performance over existing open-source alternatives and offering practical guidelines for future advancements in process supervision models for mathematical reasoning.

Let's consider a mathematical problem \(p\) along with a proposed solution \(s\). Two
reward modeling strategies to evaluate the correctness and quality of \(s\) are: Observational Reward Model (ORM) and Process Reward Model (PRM) \cite{wang2023math}.

ORM assigns a single label \(y_s \in \{0,1\}\) to indicate whether \(s\) is correct.
Let \(r_s\) denote the model's predicted probability that \(s\) is correct. The cross-entropy
loss for ORM is given by:
\[
\mathcal{L}_{\mathrm{ORM}}
= - \Bigl[y_s \log\bigl(r_s\bigr) + \bigl(1 - y_s\bigr)\log\bigl(1 - r_s\bigr)\Bigr].
\]
This approach treats the entire solution \(s\) as one binary classification task.

Let's divide the solution \(s\) into \(K\) reasoning steps \(\{s_1, s_2, \ldots, s_K\}\).
Each step \(s_i\) is assigned a label \(y_{s_i} \in \{0,1\}\) and the predicted probability
of correctness is denoted by \(r_{s_i}\). The corresponding cross-entropy loss is:
\[
\mathcal{L}_{\mathrm{PRM}}
= - \sum_{i=1}^{K}
\Bigl[y_{s_i}\,\log\bigl(r_{s_i}\bigr)
\;+\;
\bigl(1 - y_{s_i}\bigr)\,\log\bigl(1 - r_{s_i}\bigr)\Bigr].
\]

This step-wise formulation provides more granular feedback, allowing the model to
capture partial correctness in the reasoning process.

Cui et al. \cite{cui2024process,yuan2024implicitprm} introduced PRIME, an open-source online reinforcement learning (RL) framework that leverages implicit Process Reward Modelling (PRM) to enhance the reasoning abilities of LLMs significantly. Unlike traditional approaches, PRIME directly learns a Q-function that assigns token-level rewards, enabling dynamic and online updates based solely on outcome improvements. The algorithm initializes both the policy model and the PRM with a supervised fine-tuning (SFT) baseline, and then generates rollouts from multiple prompts. These rollouts are scored using an implicit PRM in conjunction with an outcome verifier, with only those prompts yielding moderate success rates retained to stabilize training. PRIME computes binary outcome rewards and token-level process rewards, applies advantage estimation via RLOO, and ultimately updates the policy using a PPO loss function. Empirical results are impressive: for instance, the PRIME 7B model achieved a pass@1 rate of 26.7\% on AIME 2024 substantially higher than the 3.3\% from SFT and 9.3\% from GPT-4o along with an average improvement of 16.7\% across key mathematical reasoning benchmarks. Additionally, the online updating of the PRM mitigates reward hacking and enhances reward accuracy while filtering out too-easy or too-hard prompts, further stabilizing the training process.   

Setlur et al. \cite{setlur2024rewarding} present a novel approach to enhancing reasoning in large language models by leveraging process reward models (PRMs) that provide step-by-step feedback throughout a multi-step reasoning process. Unlike traditional outcome reward models (ORMs) that only offer input at the final step, this method defines an effective process reward as the improvement or "progress" in the likelihood of eventually producing a correct response, measured before and after each reasoning step. Crucially, this progress is assessed under a distinct "prover" policy, which can be a weaker LLM that helps guide exploration and error correction. The authors theoretically characterize the qualities of good prover models and empirically validate their approach by training process advantage verifiers (PAVs) to predict these step-level advantages. Their experimental results demonstrate that employing PAVs in test-time search increases accuracy by more than 8\%  and achieves 1.5 to 5 times better compute efficiency than standard ORMs. Moreover, reinforcement learning with dense rewards from PAVs yields a 5- to 6-fold gain in sample efficiency and over a 6\% improvement in accuracy.

Lightman et al. \cite{lightman2023let} investigate how to process supervision, which provides feedback on each intermediate reasoning step, can significantly improve the reliability of large language models on complex tasks compared to outcome supervision, which only evaluates the final result. By training models on the challenging MATH dataset, the study demonstrates that process-supervised models augmented with active learning to optimize human feedback achieve a 78\% solve rate on a representative subset, outperforming outcome reward models (72.4\%) and effectively teaching models to verify each step in their chain-of-thought. The methodology involves fine-tuning a base GPT-4 model to output delimited reasoning steps, collecting 800,000 human-labeled step-level feedback entries (PRM800K) across 75,000 solutions to 12,000 problems, and then using reinforcement learning with these fine-grained rewards to guide the model toward more accurate and reliable reasoning.

\subsubsection{Critique-out-Loud Reward Models}

Critique-out-Loud (CLoud) reward models \cite{ankner2024critique} introduce a generative approach to reward modeling in RLHF by explicitly generating natural language critiques for each response before assigning a scalar reward. Unlike traditional reward models that rely on a single forward pass to evaluate response quality implicitly, CLoud leverages an LM head to produce detailed critiques and a reward head to predict rewards based on the prompt, response, and generated critique. This method, implemented through supervised fine-tuning on oracle critiques from Llama-3.1-405B-Instruct followed by dataset augmentation with self-generated critiques, yields significant improvements boosting pairwise preference classification accuracy on RewardBench by 4.65\% for 8B models and 5.84\% for 70B models, and achieving a Pareto improvement in win rate on ArenaHard when used in Best-of-N scoring. Using self-generated (on-policy) critiques and prompts from UltraFeedback and UltraInteract, CLoud demonstrates a promising enhancement in reward modeling that allows for a more explicit and practical evaluation of model responses.

Let's consider a dataset 
\(
D = \bigl\{\,(x,\,y^+,\,y^-,\,c^+,\,c^-)\,\bigr\}_{i=1}^{N}
\)
of \(N\) examples. Each example contains an input \(x\), a \emph{chosen} response \(y^+\) 
with an associated oracle critique \(c^+\), and a \emph{rejected} response \(y^-\) with 
its oracle critique \(c^-\). These critiques provide reasoning feedback on how well each 
response addresses the input. A reward model \(r_\theta(x,y,c)\) produces a scalar 
indicating the quality of a response \(y\) for input \(x\), conditioned on critique \(c\).

The model is trained using a pairwise objective that encourages higher scores for chosen 
responses than for rejected ones. Let \(\sigma(\cdot)\) be the sigmoid function. The loss 
is expressed as:

\begin{align*}
\mathcal{L}_{\mathrm{RM}}(\theta; D) 
 = -\,\mathbb{E}_{(x,\,y^+,\,y^-,\,c^+,\,c^-)\,\sim\,D} \\ \Biggl[
\log\Bigl(
\sigma\Bigl(r_\theta(x,y^+,c^+)-r_\theta(x,y^-,c^-)\Bigr)
\Bigr)
\Biggr]. 
\end{align*}

This formulation aligns the reward model to rank the chosen response \(y^+\) above the 
rejected response \(y^-\), guided by the corresponding critiques \(c^+\) and \(c^-\). 
By conditioning on critiques, the model gains insights into which aspects of the response 
are correct or need improvement.

Training typically proceeds with standard gradient-based optimization. The resulting 
reward model can be integrated into a more extensive reinforcement learning pipeline or used 
directly to guide response generation. In either case, the chosen response is favored 
when aligned with the critique's reasoning, and the rejected response is disfavored 
if its critique points to deficiencies. This approach helps the system learn to generate 
higher-quality responses based on explicit feedback.

\subsubsection{Generative Reward Models}

Zhang et al. \cite{zhang2024generative} introduce GenRM, a generative verifier for large language models that leverages the ubiquitous next-token prediction objective to jointly train on verification and solution generation, as opposed to traditional discriminative classifiers used in Best-of-N methods. By doing so, GenRM seamlessly integrates with instruction tuning and chain-of-thought reasoning and can utilize additional inference compute through majority voting, resulting in a 16–64\% improvement in problem-solving accuracy on algorithmic and grade-school math tasks compared to standard approaches. The method trains an LLM on synthetic chain-of-thought data to label each section of a response as "correct" or "incorrect" ultimately outputting an overall "Yes" or "No" and uses a maj@K strategy to average token probabilities for final scoring. Notably, fine-tuned small GenRMs have been shown to outperform larger judge models and performance scales favorably with additional training data (an extra ~160,000 examples) and model capacity, offering a promising alternative to conventional reward models akin to OpenAI’s CriticGPT approach.

\subsubsection{Meta-Rewarding Language Models}

Wu et al. \cite{wu2024meta} propose a self-improvement approach for large language models that emphasizes enhancing both response generation and the model’s internal judgment capabilities through an unsupervised meta-rewarding mechanism. Traditional methods typically rely on human data or self-rewarding schemes focused solely on improving responses, often leading to early saturation in the model’s ability to evaluate its outputs. In contrast, this approach allows a single LLM to assume three distinct roles actor, judge, and meta-judge where the actor generates responses, the judge scores them via multiple evaluations, and the meta-judge assesses the quality of these judgments to create refined preference pairs. By using Direct Preference Optimization (DPO) on these actor and judge preference pairs, the model significantly improves its ability to both generate and evaluate responses, as evidenced by substantial performance gains on challenging benchmarks: the win rate on AlpacaEval2 increased from 22.9\% to 39.4\% and on Arena-Hard from 20.6\% to 29.1\%. This meta-rewarding strategy, combined with careful length-control to prevent response length explosion, highlights a promising path for self-supervised, iterative model improvement without costly human feedback.

\subsubsection{Best-of-N Distillation (BOND)}

Sessa et al. \cite{sessa2024bond} propose a novel reinforcement learning from human feedback (RLHF) algorithm, Best-of-N Distillation (BOND), which aims to replicate the benefits of Best-of-N sampling while eliminating its heavy inference-time computational cost. BOND achieves this by aligning the output distribution of the policy model with that of an ideal Best-of-N distribution using a distribution matching framework. Central to this approach is the use of Jeffreys divergence a balanced combination of forward and backward KL divergences to navigate the trade-off between exploring a diverse set of outputs (mode-covering) and honing in on high-quality outputs (mode-seeking). The algorithm employs an iterative process that leverages a moving anchor model as a reference point, updated via exponential moving averages, to guide the policy's improvement. Empirical evaluations on tasks like abstractive summarization and with models from the Gemma family reveal that BOND, particularly its J-BOND variant as used in fine-tuning Google DeepMind’s Gemma 1.1 2B and 7B models, outperforms traditional RLHF methods such as REINFORCE in terms of the reward/KL trade-off.

\subsubsection{Simple Preference Optimization (SimPO)}

Meng et al. \cite{meng2024simpo} introduce SimPO (Simple Preference Optimization), a novel RLHF method that refines the Direct Preference Optimization (DPO) framework by streamlining the reward formulation. Instead of relying on a separate reference model, SimPO uses the average log probability of a generated sequence as an implicit, length-normalized reward, and it incorporates a target reward margin into the Bradley-Terry objective to enforce a more significant distinction between winning and losing responses. This design reduces computational and memory overhead cutting processing time by about 20\%  and GPU memory usage by roughly 10\%  compared to traditional DPO and significantly improves performance. Evaluations across multiple state-of-the-art training setups (including models like Mistral, Llama 3, and Gemma 2) and benchmarks such as AlpacaEval 2, MT-Bench, and Arena-Hard reveal that SimPO outperforms conventional approaches by up to 6.4 points on AlpacaEval 2 and 7.5 points on Arena-Hard. Notably, the top-performing model built on Gemma-2-9B-it achieved a 72.4\% length-controlled win rate on AlpacaEval 2, a 59.1\% win rate on Arena-Hard, and ranked first on Chatbot Arena among sub-10B models with real user votes.

\begin{figure}[t]
    \centering
    \includegraphics[width=\linewidth]{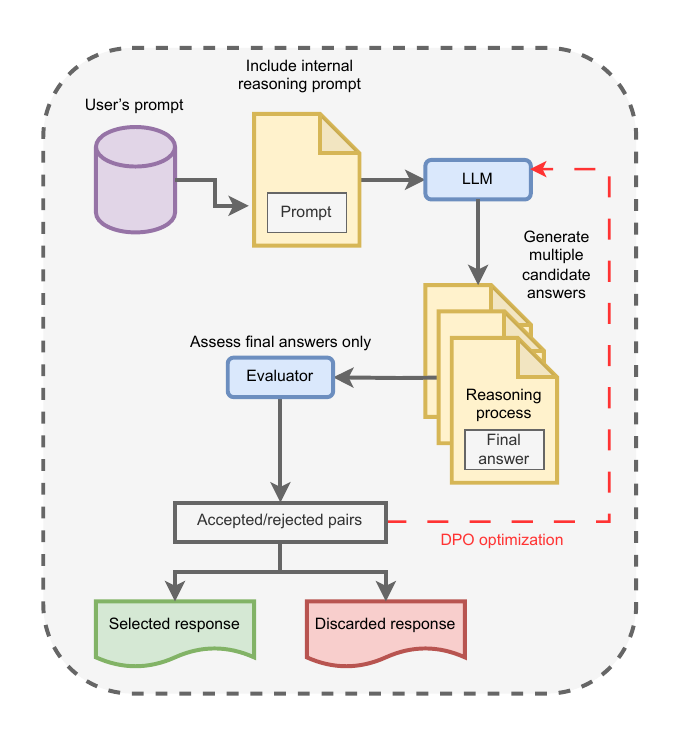}
    \caption{Thought Preference Optimization \cite{wu2024thinking} – The process begins by instructing the language model to produce an internal reasoning sequence before generating its final answer. After multiple potential responses are sampled, these answers are passed to an evaluator model that identifies the most and least preferred options. The complete outputs corresponding to these top and bottom choices serve as accepted and rejected examples for DPO optimization \cite{rafailov2023direct}. This cycle is repeated over multiple training iterations.}
    \label{fig:tpo}
\end{figure}

\subsubsection{Thought Preference Optimization (TPO)}

Wu et al. \cite{wu2024thinking} proposed a novel training methodology designed to endow LLMs with explicit "thinking" capabilities before generating their final responses, as presented in Fig. \ref{fig:tpo}. Traditionally, LLMs are optimized to answer user queries directly, but they often lack a structured, intermediate reasoning process crucial for complex tasks. The proposed approach termed Thought Preference Optimization (TPO) leverages an iterative search and optimization procedure that generates multiple chain-of-thought (CoT) sequences, explicitly separating the model’s internal reasoning ("thoughts") from its final answers. For each instruction, a judge model (e.g., ArmoRM) scores the final responses, and preference pairs are constructed by comparing the highest and lowest-scoring outputs. These pairs are then used in a length-controlled Direct Preference Optimization (DPO) training loop, repeated for several iterations to progressively refine the model's ability to "think" before answering. Empirical results demonstrate that this procedure yields substantial improvements, with approximately 20\% performance gains on benchmarks such as AlpacaEval and Arena-Hard across reasoning-intensive tasks and domains traditionally relying on more straightforward responses, like marketing, health, and general knowledge. While initial performance may temporarily lag behind direct response generation, the iterative process leads to significant overall gains, and TPO outperforms standard supervised fine-tuning on preferred outputs by around 3–4\%. Notably, the reward model evaluates only the final responses, underscoring the importance of a separate, optimized thought process. This design choice may reduce performance on datasets like GSM8K if the training data or judge model is not perfectly aligned.

\subsubsection{Proximal Policy Optimization (PPO)}

Proximal Policy Optimization (PPO) \cite{ouyang2022training} is a reinforcement learning algorithm that achieves robust policy improvement using a clipped surrogate objective to constrain updates and maintain training stability. Unlike Trust Region Policy Optimization (TRPO), which relies on complex second-order methods, PPO employs a first-order approach that limits the ratio of the new policy to the old policy within a predefined range, thereby preventing drastic changes that could destabilize learning.

Let's consider a pipeline that consists of three stages: supervised fine-tuning (SFT), preference model training, and reinforcement learning via Proximal Policy Optimization (PPO) \cite{schulman2017proximal,stiennon2020learning}.

\paragraph{Supervised Fine-Tuning (SFT)}
A base language model is first fine-tuned on labeled data \(\{(x_i, y_i)\}\), using
standard supervised objectives such as next-token prediction. This step aligns the
model with human-provided demonstrations, ensuring it produces coherent responses.

\paragraph{Preference Model Training}
A reward model \(r_\theta(x, y)\) is trained from pairs of responses \((y^+, y^-)\),
where \(y^+\) is preferred over \(y^-\) for a given input \(x\). A cross-entropy
objective on these pairwise comparisons guides the model to assign a higher reward
scores to more desirable responses:
\begin{align*}
\mathcal{L}_{\mathrm{RM}}
&=
-\,\mathbb{E}_{(x,\,y^+,\,y^-)\,\sim\,D}
\Bigl[
\log\Bigl(
\sigma\bigl(r_\theta(x,y^+) - r_\theta(x,y^-)\bigr)
\Bigr)
\Bigr].
\end{align*}
This process yields a learned reward function that can distinguish better responses
from worse ones.

\paragraph{Proximal Policy Optimization (PPO)}
Once the reward model is trained, the policy \(\pi_\phi\) (initially obtained from
SFT) is further optimized via PPO. Let \(\phi_{\mathrm{old}}\) be the parameters
of the policy before the current update. The probability ratio is
\(\displaystyle r_t(\phi) = \frac{\pi_\phi(a_t \mid x_t)}{\pi_{\phi_{\mathrm{old}}}(a_t \mid x_t)}\).
A typical clipped PPO objective is
\begin{align*}
\mathcal{L}_{\mathrm{PPO}}(\phi)
&=
\mathbb{E}
\Bigl[
\min\Bigl(r_t(\phi)\,A_t,\;
\mathrm{clip}\bigl(r_t(\phi),\,1-\epsilon,\,1+\epsilon\bigr)\,A_t\Bigr)
\Bigr],
\end{align*}
where \(A_t\) is an advantage estimate derived from the reward model \(r_\theta\),
and \(\epsilon\) is a hyperparameter (e.g., \(0.1\) or \(0.2\)) that controls
the clipping range. By using \(r_\theta\) to compute returns or advantages, the
policy learns to generate responses that achieve higher scores under the preference
model, while the clipping mechanism in PPO constrains large policy updates.

\subsubsection{Group Relative Policy Optimization (GRPO)}

DeepSeek team propose DeepSeekMath 7B \cite{shao2024deepseekmath}, a model specifically designed to tackle the challenges of mathematical reasoning by building on DeepSeek-Coder-Base-v1.5 7B and leveraging an enormous corpus of 120 billion math-related tokens alongside natural language and code data. Notably, DeepSeekMath 7B achieves an impressive 51.7\% accuracy on the MATH benchmark without the need for external toolkits or voting techniques, and its performance further rises to 60.9\% with self-consistency over 64 samples, approaching the levels of high-end models like Gemini-Ultra and GPT-4. A key innovation in this work is the introduction of Group Relative Policy Optimization (GRPO), an RLHF method that refines mathematical reasoning by optimizing the memory usage of the conventional PPO algorithm. GRPO distinguishes itself by eliminating the need for a separate value function and directly integrating the KL-divergence term into the loss function, leading to roughly a 5\%  improvement on benchmarks such as GSM8K and MATH. Utilizing an iterative approach with 144,000 chain-of-thought prompts and training its reward model via the “Math-Shepherd” process enables more efficient and specialized reasoning, setting a new standard for open-source language models in mathematical problem-solving.

SWE-RL \cite{wei2025swerladvancingllmreasoning} tackles the significant challenge of scaling reinforcement learning-based reasoning to the real-world domain of software engineering, where tasks like debugging and issue resolution require not only technical expertise but also nuanced reasoning over complex, evolving codebases. Unlike previous RL approaches focusing on competitive coding or math, SWE-RL leverages extensive open-source software evolution data including code snapshots, pull requests, and issue contexts to train LLMs to recover developers' reasoning processes autonomously. By employing a lightweight, rule-based reward function that compares generated solutions to ground-truth patches, and optimizing using Group Relative Policy Optimization (GRPO), SWE-RL achieves state-of-the-art performance among medium-sized models, with a 41.0\% solve rate on the human-verified SWE-Bench.

Group Relative Policy Optimization (GRPO) as a reinforcement learning algorithm
that optimizes a policy \(\pi_\theta\) through groupwise comparisons among candidate responses,
rather than relying solely on an explicit KL penalty. GRPO is designed to steer \(\pi_\theta\)
toward generating higher-reward outputs while maintaining closeness to a reference policy
\(\pi_{\mathrm{ref}}\). In this framework, the key components are:
\begin{itemize}
    \item An initial policy \(\pi_\theta\) (e.g., obtained from pre-training or fine-tuning).
    \item A reference policy \(\pi_{\mathrm{ref}}\) that serves as a baseline.
    \item A reward model \(r_\varphi(x,y)\) that assigns a scalar reward to each context-response
    pair \((x,y)\).
    \item A dataset \(\mathcal{D}\) of task prompts.
\end{itemize}

The GRPO procedure involves the following steps:
\begin{itemize}
    \item Sampling: Draw a batch of prompts from \(\mathcal{D}\) and, for each prompt,
    generate multiple candidate responses using \(\pi_\theta\).
    \item Reward Evaluation: Compute the rewards \(r_\varphi(x, y)\) for each generated
    response.
    \item Relative Advantage Estimation: Within each group of candidate responses, estimate the
    relative advantages through pairwise or groupwise comparisons.
    \item Policy Update: Update \(\theta\) by optimizing an objective that leverages these
    relative comparisons to bias the policy towards higher-reward responses.
\end{itemize}

GRPO employs an unbiased estimation of
the KL divergence to control the divergence from the reference policy. Specifically, the KL divergence between \(\pi_\theta\) and
\(\pi_{\mathrm{ref}}\) is approximated as:
\begin{align*}
D_{\mathrm{KL}}\!\Bigl(\pi_\theta\,\|\,\pi_{\mathrm{ref}}\Bigr)
&\approx
\mathbb{E}_{(a,x) \sim \pi_{\mathrm{ref}}}
\!\Biggl[
\frac{\pi_\theta(a\mid x)}{\pi_{\mathrm{ref}}(a\mid x)} - 1
\Biggr].
\end{align*}

This estimation helps to ensure that the updated policy does not deviate excessively from
\(\pi_{\mathrm{ref}}\) without explicitly incorporating a KL penalty term in the loss function. Through iterative updates that combine groupwise relative advantage estimation and KL control, GRPO guides \(\pi_\theta\) toward producing responses with higher rewards while preserving
stability and alignment with the reference policy.

\begin{figure}[t]
    \centering
    \includegraphics[width=\linewidth]{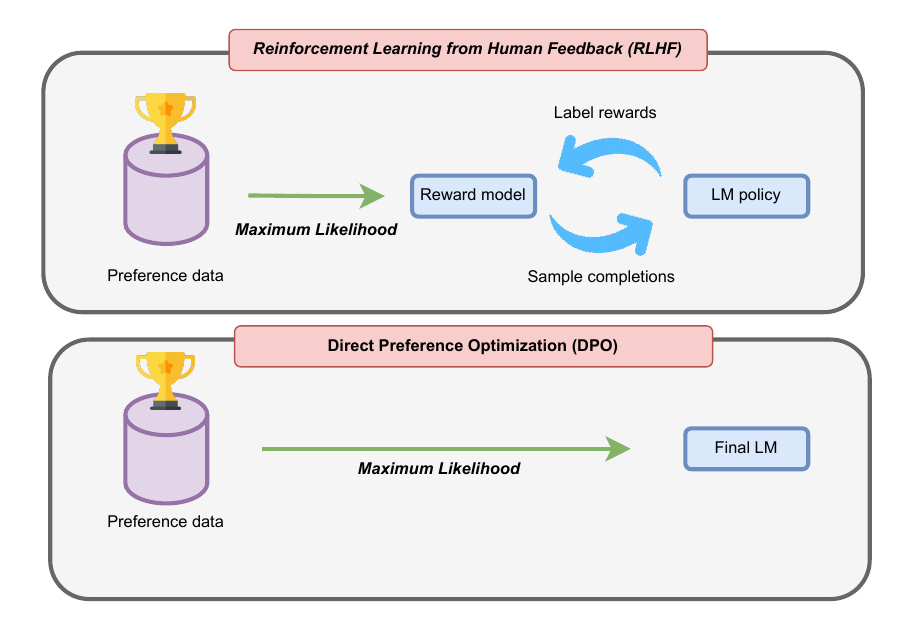}
    \caption{ Direct Preference Optimization (DPO) \cite{rafailov2023direct} – DPO aligns models with human preferences without employing a reinforcement learning stage. Traditional approaches for fine-tuning language models with human feedback \cite{christiano2017deep} typically involve training a reward model on a dataset of prompts and preference annotations, then using RL to discover a policy that maximizes the learned reward. By contrast, DPO directly adapts the policy to fulfill these preferences best using a straightforward classification-based objective, avoiding both explicit reward modeling and RL.}
    \label{fig:dpo}
\end{figure}

\subsubsection{Direct Preference Optimization (DPO)}

Rafailov et al. \cite{rafailov2024direct} propose Direct Preference Optimization (DPO), a streamlined method for aligning large language models with human preferences that reframe the alignment challenge as a classification problem on preference data. By leveraging a mapping between reward functions and optimal policies, DPO bypasses the need for a separate reward model and complex PPO-based reinforcement learning, thus simplifying implementation and reducing computational and hyperparameter tuning requirements as presented in Fig. \ref{fig:dpo}. The approach enhances stability by avoiding the instabilities typically associated with traditional RLHF and has been empirically shown to achieve comparable or superior performance in controlling response attributes such as sentiment in the generated text and improving quality in tasks like summarization and single-turn dialogue.

Let's consider Direct Preference Optimization (DPO) as a preference optimization method
that removes the need for an explicit reward model. Instead, DPO relies on a closed-form
expression connecting the policy model \(\pi_\theta\) and a reference model
\(\pi_{\mathrm{ref}}\), often a supervised fine-tuned model. By incorporating a
Bradley--Terry-style ranking function, DPO defines a pairwise comparison between
a \emph{winning} response \(y_w\) and a \emph{losing} response \(y_l\) for a given
input \(x\). The objective is:

\begin{align*}
\mathcal{L}_{\mathrm{DPO}}(\pi_\theta;\,\pi_{\mathrm{ref}})
=
-\,\mathbb{E}_{(x,\,y_w,\,y_l)\,\sim\,D}
\biggl[
\log \frac{\pi_\theta(y_w\mid x)}{\pi_{\mathrm{ref}}(y_w\mid x)}
\;-\; \\
\log \frac{\pi_\theta(y_l\mid x)}{\pi_{\mathrm{ref}}(y_l\mid x)}
\biggr],
\end{align*}

where \((x,y_w,y_l)\) represents a preference triplet from the dataset \(D\). This
formulation encourages \(\pi_\theta\) to assign higher likelihood to the winning
response compared to the losing one, relative to \(\pi_{\mathrm{ref}}\). The
\(\log Z(x)\) term from the reward definition is implicitly handled by the
pairwise comparison, allowing DPO to optimize policy parameters \(\theta\) in a
direct and efficient manner.

\subsubsection{Iterative length-regularized DPO (iLR-DPO)}

Liu et al. \cite{liu2024iterative} examine Direct Preference Optimization (DPO) and explore its iterative extension using online preferences derived from a trained reward model. The authors identify a critical limitation in the conventional iterative DPO approach: while response quality improves, it inadvertently increases verbosity. To mitigate this issue, they propose an enhanced method iterative length-regularized DPO (iLR-DPO) which incorporates a penalty on response length. Empirical evaluations reveal that iLR-DPO enables a 7B model to achieve performance comparable to GPT-4, without the trade-off of excessive verbosity. Notably, the model attains a 50.5\% length-controlled win rate against GPT-4 Preview on AlpacaEval 2.0 and demonstrates strong results across various benchmarks, including MT-Bench, Arena-Hard, and the OpenLLM Leaderboard. These findings underscore the efficacy of integrating iterative training with length regularization to improve the alignment of language models with human feedback.

iLR-DPO extends Direct Preference Optimization (DPO) by introducing a length penalty
to reduce response verbosity while preserving preference performance. Let
\(\mathrm{pm}\) be the \emph{preference margin}, which measures how much the winning
response \(y_w\) outperforms the losing response \(y_l\), and let \(\mathrm{lm}\)
be the \emph{length margin}, typically defined as \(\lvert y_w \rvert - \lvert y_l \rvert\).
Two hyperparameters \(\beta\) and \(\alpha\) balance the trade-off between maximizing
preference and limiting verbosity. The margin-based cross-entropy objective is:

\begin{align*}
\nabla_{\theta_{i+1}}
\,\mathcal{L}_{\mathrm{LR\text{-}DPO}}
\bigl(\theta_{i+1};\,\theta_i,\,D_i\bigr)
= 
\nabla_{\theta_{i+1}}
\mathbb{E}_{(x,\,y_w,\,y_l)\,\sim\,D_i} \\
\Bigl[
\log\!\Bigl(
\sigma\bigl(\beta\cdot \mathrm{pm} \;+\; \alpha\cdot \mathrm{lm}\bigr)
\Bigr)
\Bigr].
\end{align*}

This formulation encourages the model to maintain a high preference margin while
avoiding unnecessarily lengthy responses. Training typically starts from the final
language model checkpoint, and the length penalty helps control verbosity without
compromising overall performance.

\subsubsection{Length-Instruction DPO (LIFT-DPO)}

Yuan et al. \cite{yuan2024following} tackle the persistent issue of length bias in aligned instruction-following models. While such models generally outperform unaligned ones, they tend to generate unnecessarily lengthy responses a consequence of both evaluation biases and training practices. The authors propose that incorporating explicit length constraints into prompts can resolve this ambiguity, thereby improving both the quality and efficiency of generated outputs. To achieve this, they introduce Length-Instruction Fine-Tuning (LIFT), a method that augments conventional instruction-following datasets with explicit length limits and constructs new preference pairs that penalize verbosity. They also establish new benchmarks AlpacaEval-LI and MT-Bench-LI to rigorously assess how well models adhere to these length instructions. Their experiments demonstrate that state-of-the-art models like GPT-4 Turbo and Claude 3 often violate length constraints, whereas models fine-tuned with LIFT exhibit substantially lower violation rates and improved win rates in length-controlled evaluations, all without degrading performance on standard tasks. This work not only offers a promising strategy to mitigate length bias but also provides a more controllable approach to generating concise responses in practical applications.

\subsubsection{Identity Preference Optimisation (IPO)}
Azar et al. \cite{azar2024general} critically examine the two key approximations underlying current RLHF methods for aligning large language models with human preferences. Traditionally, these methods assume that pointwise rewards can replace pairwise preference data and that a reward model trained on such rewards can generalize to unseen data. While Direct Preference Optimization (DPO) has been introduced to bypass the second approximation by learning a policy directly from collected data without an explicit reward model, it still depends on the first assumption. To address this, the authors derive a new general objective, IPO, formulated entirely in pairwise preferences, thereby eliminating both approximations. They further explore a special IPO case by setting I to the identity function developing an efficient optimization procedure with proven performance guarantees. Empirical results on illustrative examples indicate that this approach not only provides deeper theoretical insights into the behavior of RLHF and DPO but also achieves superior performance compared to DPO, marking a promising direction for more reliable and theoretically grounded preference-based learning.

\subsubsection{Kahneman-Tversky Optimisation (KTO)}

Ethayarajh et al. \cite{ethayarajh2024kto} explore how alignment objectives for LLMs, such as those used in Direct Preference Optimization (DPO), inherently capture human biases similar to those described in Kahneman and Tversky's prospect theory for example, loss aversion. The authors introduce the concept of human-aware losses (HALOs), arguing that these loss functions outperform traditional cross-entropy minimization by implicitly modeling human perception. However, they note that the utility functions assumed by current methods differ from those in prospect theory. To bridge this gap, the authors propose KTO, which directly maximizes the utility of generated outputs based on a Kahneman-Tversky utility model rather than simply optimizing the log-likelihood of preferences. Notably, KTO matches or exceeds the performance of preference-based methods on models ranging from 1B to 30B parameters, even though it learns solely from a binary signal of output desirability.

\subsubsection{Decoupled Clip and Dynamic sAmpling Policy Optimization (DAPO)}

Yu et al. \cite{yu2025dapoopensourcellmreinforcement} introduces DAPO a novel reinforcement learning algorithm designed to enhance the reasoning capabilities of large language models through inference scaling. Unlike previous methods where critical training details were often omitted, DAPO is fully open-sourced, including its code built on the verl framework and a carefully curated dataset (DAPO-Math-17k), which significantly bolsters reproducibility. The approach distinguishes itself through four key innovations: an asymmetric clipping mechanism (Clip-Higher) that prevents entropy collapse; dynamic sampling that filters out prompts yielding trivial outcomes; a token-level policy gradient loss that controls response length while maintaining reasoning integrity; and a length-aware penalty to mitigate reward noise from potentially valid but truncated long responses. These methodological advances enable DAPO, when applied to the Qwen2.5-32B base model, to achieve 50 points on the AIME 2024 benchmark surpassing competitors like GRPO with 50\% fewer training steps thereby marking a notable step forward in the scalable training of reasoning-focused LLMs.

DAPO is a reinforcement learning algorithm that optimizes a policy \(\pi_\theta\)
to produce higher-reward outputs under a learned reward model \(R\). For each prompt
\(q_i\) with partial answer \(a_i\), DAPO samples a set of candidate completions
\(\{o_i^j\}_{j=1}^G\). The policy is then updated to maximize the following objective:

\begin{align*}
J_{\mathrm{DAPO}}(\theta)
=\;
\mathbb{E}_{(q_i,a_i)\,\sim\,D}
\biggl[
\frac{1}{G}
\sum_{j=1}^{G}
\min\!\Bigl(
\,r_i\bigl(o_i^j\bigr)\,A_i^j,\;
\bigl[r_i\bigl(o_i^j\bigr)- \\ \epsilon_{\mathrm{low}},\;1+\epsilon_{\mathrm{high}}\bigr]\,
A_i^j
\Bigr)
\biggr],
\end{align*}

subject to constraints on the candidate set size and filtering conditions 
(\(0 < |o_i| < G\)), where:
\begin{itemize}
    \item \(r_i(o_i^j)\) is the reward model's score for completion \(o_i^j\).
    \item \(\epsilon_{\mathrm{low}},\epsilon_{\mathrm{high}}\) are clipping thresholds.
    \item \(A_i^j\) denotes an advantage-like term (e.g., derived from relative
    performance among the sampled completions).
\end{itemize}

Instead of relying on a fixed KL penalty, DAPO estimates the probability ratio
between the new policy \(\pi_\theta\) and the old policy \(\pi_{\theta_{\mathrm{old}}}\)
as follows:
\begin{align*}
\label{eq:dapo_ratio}
r_i(\theta)
&=\;
\frac{\pi_\theta\bigl(o_i \mid q_i,\,o_{i,<t}\bigr)}%
{\pi_{\theta_{\mathrm{old}}}\bigl(o_i \mid q_i,\,o_{i,<t}\bigr)},
\quad
A_i^j
=\;
\frac{R_i - \mathrm{mean}\bigl(\{R_i\}\bigr)}{\mathrm{std}\bigl(\{R_i\}\bigr)},
\tag{9}
\end{align*}

where \(R_i\) represents raw reward values for the sampled completions, and
\(\mathrm{mean}(\cdot)\), \(\mathrm{std}(\cdot)\) are computed over the batch
to normalize the reward scores into advantage estimates.

\subsection{Retrieval-Augmented Generation (RAG)}

Databricks Mosaic Research \cite{databricks2023longcontext} presents a comprehensive evaluation of long-context Retrieval Augmented Generation (RAG) capabilities across several state-of-the-art language models. The authors compared models including OpenAI o1, Anthropic Claude, Google DeepMind Gemini, and Meta Llama on three diverse datasets Databricks DocsQA, FinanceBench, and Natural Questions by varying context lengths from 2,000 up to 2,000,000 tokens retrieved from a vector database. Using GPT-4o as the judge to assess answer correctness, the study found that Google DeepMind's Gemini 1.5 maintains strong RAG performance even with highly extended contexts (up to 2 million tokens). In contrast, open models like Llama-3.1-405B show significant performance degradation beyond 32k tokens. OpenAI's o1-preview model achieved the highest average performance score of 0.763 across the tested context lengths, although standard OpenAI o1 models underperformed at shorter contexts on specific datasets. Most models demonstrated peak performance in the 32k-64k token range before declining, highlighting critical insights into the optimal context lengths and trade-offs in deploying long-context RAG systems.

Wang \cite{wang2024searching} offers a comprehensive investigation into Retrieval-Augmented Generation (RAG) techniques, which are widely recognized for their ability to integrate current information, reduce hallucinations, and enhance output quality in domain-specific applications. By dissecting the multi-step RAG workflow ranging from query classification, advanced chunking, and hybrid retrieval using methods like HyDE, to effective reranking with models such as MonoT5 the study systematically evaluates various components to establish best practices that strike a balance between performance and efficiency. Additionally, the work explores multimodal retrieval strategies that improve the handling of visual inputs and accelerate the generation of multimodal content via a "retrieval as generation" approach. Although the paper is more of a synthesis of existing methodologies rather than a breakthrough in novel techniques, its extensive experimentation and clear recommendations, including the use of Milvus for vector storage, reverse repacking strategies, and mixed-context fine-tuning for robust generation, provide valuable insights for optimizing RAG systems. 

\subsubsection{Chain-of-Retrieval Augmented Generation (CoRAG)}

Wang et al. \cite{wang2025chain} propose CoRAG (Chain-of-Retrieval Augmented Generation), a novel framework designed to enhance traditional RAG models by incorporating iterative retrieval and reasoning steps before answer generation. Conventional RAG approaches typically rely on a single retrieval phase, which can limit performance when handling complex, multi-hop queries due to incomplete or imperfect retrievals. In contrast, CoRAG dynamically reformulates the query as the model’s internal state evolves, enabling the construction of intermediate retrieval chains that better capture the necessary context. The training process leverages rejection sampling to automatically generate these chains, effectively augmenting datasets that otherwise only provide the final answer. The method employs flexible decoding strategies at inference time to manage computational demands by controlling the length and number of retrieval chains sampled. Empirical results across various benchmarks, including a notable improvement in exact match scores for multi-hop question answering and state-of-the-art performance on the KILT benchmark, validate the efficacy of CoRAG

\subsubsection{RAG mechanism with "Reason-in-Documents" module}

Li et al. \cite{li2025search} propose Search-o1, an innovative framework designed to enhance large reasoning models (LRMs) such as OpenAI-o1, which, despite their robust stepwise reasoning abilities, often encounter limitations due to insufficient knowledge during extended reasoning processes. To overcome these challenges, Search-o1 incorporates an agentic retrieval-augmented generation (RAG) mechanism that dynamically searches for and retrieves external knowledge when uncertainties arise. Furthermore, the framework includes a dedicated Reason-in-Documents module that refines and analyzes the verbose retrieved content, ensuring that only coherent and relevant information is integrated into the reasoning chain. Experimental evaluations across diverse complex reasoning tasks in science, mathematics, coding, and open-domain question answering benchmarks demonstrate that Search-o1 significantly improves performance, thereby enhancing the reliability and versatility of LRMs in handling intricate reasoning challenges.

\subsubsection{Contextualized Document Embeddings}

Morris et al. \cite{morris2024contextual} advanced the field of neural retrieval by proposing contextualized document embeddings that incorporate information from neighboring documents, addressing a key limitation of traditional dense embeddings that are generated in isolation. The authors introduce two complementary approaches: one modifies the contrastive learning objective to explicitly integrate neighbor information into the intra-batch loss, and the other develops a new architectural design that encodes context from neighboring documents directly into the representation. This dual strategy leads to significant improvements over standard biencoder methods, particularly in out-of-domain retrieval scenarios. Notably, the method achieves state-of-the-art performance on the MTEB benchmark without relying on standard techniques such as hard negative mining, score distillation, or extremely large batch sizes. By clustering similar documents to identify neighbors and seamlessly incorporating their context during both training and encoding, the approach yields more robust and context-aware embeddings. This work not only enhances retrieval performance for applications such as Retrieval-Augmented Generation (RAG) but also offers a versatile framework that can be applied to any contrastive learning dataset or biencoder model, underscoring its broad applicability and potential impact on improving retrieval systems.

\subsubsection{RAG with Self-Reasoning}

Xia et al. \cite{xia2024improving} presents an innovative self-reasoning framework designed to enhance the reliability and traceability of Retrieval-Augmented Language Models (RALMs), addressing persistent challenges such as irrelevant document retrieval and insufficient citation in generated responses. The proposed approach leverages the internal reasoning trajectories of the LLM through a structured three-stage process. First, the Relevance-Aware Process (RAP) instructs the model to assess the relevance of retrieved documents to the query. Next, the Evidence-Aware Selective Process (EAP) extracts key sentences from these documents, along with justifications for their selection. Finally, the Trajectory Analysis Process (TAP) synthesizes these reasoning paths to produce a final answer. Evaluated on four diverse public datasets including short-form and long-form QA as well as fact verification tasks the framework demonstrates a performance improvement of over 10\% compared to standard RAG approaches when applied to a Llama 2 7B model, all while requiring only 2,000 training samples generated using GPT-4. This method not only improves the factual accuracy and trustworthiness of RALMs but also offers a scalable solution for integrating structured, self-generated reasoning into retrieval-based systems.

\subsection{Evolutionary search strategy}

Lee et al. \cite{lee2025evolving} introduced Mind Evolution, an innovative evolutionary search strategy designed to enhance inference time computation in LLMs for natural language planning tasks. Rather than requiring an explicit formalization of the inference problem, this approach leverages a language model to iteratively generate, evaluate, and refine candidate solutions, using a programmatic evaluator to guide the process. Mind Evolution significantly outperforms traditional inference strategies such as Best-of-N and Sequential Revision by systematically recombining and optimizing candidate responses over multiple generations. Experimental results on benchmarks like TravelPlanner and Natural Plan demonstrate a success rate exceeding 98\% when using Gemini 1.5 Pro, all while maintaining a more cost-effective inference process compared to sequential revision methods. Additionally, the framework supports a two-stage approach initially employing a faster model for simpler tasks, then transitioning to a more robust model for complex cases thereby showcasing its adaptability and efficiency in tackling planning problems that are challenging to formalize but straightforward to evaluate.

\subsection{Data leakage and deduplication}

SemHash \cite{minishlab2025semhash} is a high-performance semantic deduplication library designed to address critical challenges in data leakage and deduplication during the training of LLMs. SemHash leverages Model2Vec embeddings in conjunction with an approximate nearest neighbor (ANN) similarity search via Vicinity, enabling rapid and efficient deduplication of large-scale text datasets. Notably, the system can process 1.8 million WikiText records in just 83 seconds on a CPU, highlighting its remarkable speed and scalability. By focusing on semantic similarity rather than strict exact matching, SemHash can identify near-duplicate records, thereby enhancing the quality of both training and evaluation datasets whether for cleaning single datasets or for preventing cross-dataset leakage between training and testing splits. The library offers a straightforward Python API with minimal dependencies. It supports the integration of custom encoders, such as those provided by sentence-transformers, along with built-in tools for inspecting duplicates and adjusting similarity thresholds. Benchmark results across 17 popular datasets underscore its effectiveness, making SemHash a valuable tool for researchers and practitioners working on preparing high-quality datasets for LLM training.

\subsection{A non-agent framework}

Xia et al. \cite{xia2024agentless} challenges the prevailing reliance on complex autonomous software agents for automating software development tasks by introducing Agentless, a streamlined, non-agent framework that simplifies the repair process into distinct phases of localization and repair. Instead of employing elaborate agent libraries and decision-making for tool usage, Agentless leverages a tree-structured repository view alongside embedding-based retrieval techniques to isolate suspicious files and extract function and class signatures the “skeletons” of code under investigation. The framework then generates targeted search/replace diffs and corresponding reproduction tests, ultimately validating the best patch via majority voting and regression testing. Evaluated on the SWE-bench Lite benchmark, Agentless not only achieves the highest performance at a 27.33\% solve rate but also incurs significantly lower costs (\$0.34 per issue) compared to existing agent-based approaches. These findings underscore that a focused, interpretable pipeline can effectively address real-world software engineering problems while avoiding the complexity and pitfalls of more verbose autonomous agent systems, thereby resetting the baseline and inspiring future work in autonomous software development.

\subsection{Chain-of-Thought (CoT)}

Chain-of-thought prompting \cite{wei2022chain} guides a language model to articulate intermediate reasoning steps before reaching a final answer. This method mirrors human problem-solving by decomposing complex tasks into simpler, sequential parts. It not only boosts the model's accuracy on challenging tasks such as multi-step arithmetic and symbolic reasoning but also makes its internal decision-making more transparent, thereby aiding in error analysis and debugging.

Cui et al. \cite{cui2024theoretical} investigated the effectiveness of extending few-shot chain-of-thought (CoT) prompting by incorporating both correct and incorrect reasoning paths along with detailed explanations for the errors. Traditional CoT prompting methods typically provide a sequence of in-context learning steps (Stepwise ICL) that guide the transformer’s reasoning process; however, the authors theoretically demonstrate that integrating coherent reasoning from earlier steps enhances error correction and improves prediction accuracy. Their analysis reveals that transformers are particularly sensitive to mistakes made in intermediate reasoning steps rather than solely focusing on the final outcome. Building on this insight, the proposed approach augments standard few-shot CoT prompts with deliberately generated incorrect reasoning examples accompanied by explanations that clarify the nature of the mistakes. The implementation involves selecting a base large language model (e.g., GPT-3.5, GPT-4, Gemini Pro, or DeepSeek 67B), generating correct step-by-step reasoning samples, creating and annotating incorrect reasoning paths, and combining these elements in the demonstration prompts. Experimental results indicate that using model-generated incorrect examples with corresponding explanations significantly enhances reasoning performance across diverse LLMs, while omitting the explanatory component or relying on handcrafted errors can be detrimental.

\subsubsection{Chain of Continuous Thought}

Meta team introduces Coconut (Chain of Continuous Thought)  \cite{hao2024training}, a novel reasoning paradigm for LLMs that shifts from traditional chain-of-thought (CoT) reasoning in natural language to a latent-space-based approach. Instead of decoding hidden states into textual tokens, Coconut directly feeds them back into the model as input, enabling a more flexible and efficient reasoning process. By utilizing two modes Language Mode for standard text generation and Latent Mode for iterative hidden-state processing Coconut facilitates a chain of continuous thoughts, allowing the model to explore multiple alternative reasoning paths simultaneously. This method, marked by special $<bot>$ and $<eot>$ tokens, enables breadth-first search (BFS)-like reasoning, which outperforms CoT in complex planning tasks such as ProntoQA and ProsQA, particularly those requiring substantial backtracking. Additionally, Coconut significantly reduces token generation during inference, making it more efficient while maintaining accuracy. Inspired by iterative CoT (iCoT), the model is trained using a multi-stage curriculum learning strategy that gradually replaces language-based reasoning with latent-space reasoning, ensuring a smooth transition. The findings demonstrate that leveraging latent representations enhances reasoning efficiency, making Coconut a promising advancement in LLM-based problem-solving and strategic planning.

\subsubsection{Chain-of-Retrieval Augmented Generation (CoRAG)}

Wang et al. \cite{wang2025chain} introduced Chain-of-Retrieval Augmented Generation (CoRAG), a novel approach designed to address a key limitation in traditional RAG systems namely, the reliance on a single retrieval step that often falls short when handling complex, multi-hop queries. Conventional RAG methods retrieve supporting information once before the generation phase, which can lead to incomplete or suboptimal evidence for intricate queries. In contrast, CoRAG enables dynamic query reformulation by allowing the model to iteratively retrieve and integrate information, effectively building a chain of reasoning that better supports the final answer. To overcome the challenge of limited annotated data for intermediate steps, the authors employ rejection sampling to automatically generate plausible retrieval chains from existing RAG datasets that typically only include the final correct answer. During inference, they propose flexible decoding strategies that adjust the length and number of retrieval chains, providing a trade-off between computational cost and accuracy. Experimental results demonstrate that CoRAG significantly outperforms strong baselines improving exact match scores by over 10 points on multi-hop question answering tasks and sets a new state-of-the-art on the KILT benchmark across various knowledge-intensive tasks. Comprehensive analyses of the model’s scaling behavior further underscore its potential to guide future research toward developing more factual and grounded foundation models.

Let's consider a question--answering (QA) dataset that provides a query \(Q\) and a
final answer \(A\), without any intermediate retrieval steps. CoRAG introduces an
automated procedure to generate a sequence of sub-queries and sub-answers
\(\{(Q_i, A_i)\}_{i=1}^L\), where \(L\) is a predetermined chain length. Each
sub-query \(Q_i\) is produced by conditioning on \(Q\) and previously obtained
sub-answers. The sub-answers \(A_i\) are then generated by a large language model
(LLM) using the top-\(k\) relevant documents \(\{D_{1:k}\}\). Rejection sampling
selects the best sub-query and sub-answer pairs based on log-likelihood or other
metrics, forming a retrieval chain that connects the original query to the final
answer.

\paragraph{Multi-Task Training.}
The model is fine-tuned to predict three types of outputs, each trained with a
cross-entropy loss:
\begin{align*}
L_{\mathrm{sub\_query}}
&=
-\log P\bigl(Q_i \,\bigm|\,
Q,\,Q_{<i},\,A_{<i}\bigr), \\[6pt]
L_{\mathrm{sub\_answer}}
&=
-\log P\bigl(A_i \,\bigm|\,
Q,\,Q_{1:i},\,A_{<i},\,D_{1:k}\bigr), \\[6pt]
L_{\mathrm{final\_answer}}
&=
-\log P\bigl(A \,\bigm|\,
Q,\,Q_{1:L},\,A_{1:L},\,D_{1:k}\bigr).
\end{align*}
Here, \(L_{\mathrm{sub\_query}}\) trains the model to generate each next sub-query,
\(L_{\mathrm{sub\_answer}}\) trains it to produce the correct sub-answer for each
sub-query, and \(L_{\mathrm{final\_answer}}\) focuses on generating the final answer
\(A\). Only the target output tokens (sub-query, sub-answer, or final answer) are
used in the loss. The model acquires
a better grasp of the step-by-step retrieval reasoning and final answer formulation by training simultaneously on all three tasks.

\paragraph{Iterative Rejection Sampling.}
After training, the model can be used in an iterative loop. For a new query \(Q\),
the model generates a retrieval chain by sampling sub-queries and sub-answers,
and then selects the highest-likelihood chain for final answer production. This
newly generated data can be fed back into the training pipeline to further refine
the model. As a result, CoRAG leverages multi-task supervision and rejection
sampling to improve both retrieval quality and answer correctness, producing a
robust QA system that learns to construct and utilize intermediate retrieval steps.

\subsubsection{Monte Carlo Tree Search (OmegaPRM)}

Luo et al. \cite{luo2024improve} addresses the persistent challenge of enhancing multi-step reasoning particularly in mathematical problem solving by introducing an automated process supervision framework that eschews human annotation. Conventional Outcome Reward Models (ORMs) often fall short in tasks with lengthy or multi-hop reasoning chains because they do not adequately reward or penalize intermediate steps. To overcome this, the authors propose OmegaPRM, a novel divide-and-conquer style Monte Carlo Tree Search (MCTS) algorithm that efficiently collects high-quality process supervision data. OmegaPRM employs a binary search strategy to swiftly locate the first error in a chain-of-thought (CoT), thereby balancing positive and negative examples and significantly reducing computational overhead. This method enabled the generation of over 1.5 million supervision annotations, which were used to train a Process Reward Model (PRM) via pointwise soft label training objectives based on Monte Carlo estimations. When integrated with a weighted self-consistency algorithm, the approach improved the math reasoning performance of the instruction-tuned Gemini Pro model, boosting its success rate on the MATH benchmark from 51\% to 69.4\% a 36\% relative improvement. Importantly, this fully automated framework offers a cost-effective alternative to human annotation while demonstrating superior performance compared to previous methods such as PRM800K and Math-Shepherd, marking a significant step forward in developing more reliable and efficient process supervision techniques for LLM reasoning tasks.

\subsubsection{Self-taught Reasoner with Tools}

Li et al. \cite{li2025start} introduce START (Self-Taught Reasoner with Tools), an innovative framework that enhances the reasoning capabilities of large language models by integrating external computational tools into the long chain-of-thought (CoT) process. Traditional large reasoning models, such as OpenAI-o1 and DeepSeek-R1, exhibit strong reasoning abilities but are often hindered by hallucinations and inefficiencies stemming from their exclusive reliance on internal reasoning. START addresses these limitations by incorporating a self-learning framework that enables the model to perform complex computations, self-check, explore alternative methods, and self-debug through tool calls like a Python interpreter. The core innovations include the Hint-infer mechanism, which strategically injects context-specific hints (e.g., “Wait, maybe using Python here is a good idea”) to activate latent tool-using abilities without needing demonstration data, and the Hint Rejection Sampling Fine-Tuning (Hint-RFT) process, which refines reasoning trajectories via scoring, filtering, and fine-tuning. Applying to the QwQ-32B model, START demonstrates substantial improvements across various benchmarks including AMC, AIME, and coding challenges achieving competitive performance that rivals state-of-the-art models while significantly reducing debug errors and enhancing overall problem-solving accuracy.

\subsection{Test-Time Compute scaling}

Brown et al. \cite{brown2024large} explore an often-overlooked axis of scaling namely, inference compute by systematically examining how repeated sampling can enhance the performance of language models. Traditionally, LLMs make a single attempt at generating an answer, but this study demonstrates that generating multiple, diverse candidate solutions and selecting the best one can significantly improve the fraction of problems solved, a metric referred to as coverage. Empirical results indicate that coverage scales in a log-linear fashion with the number of samples, following an exponentiated power law. In domains such as coding and formal proofs where automatic verifiers like unit tests are available this approach directly translates into measurable performance gains. For instance, on the SWE-bench Lite benchmark, the DeepSeek-Coder-V2-Instruct model's performance increases from 15.9\% to 56\% when the number of samples is raised from one to 250, outstripping the single-sample state-of-the-art of 43\%. However, in domains lacking automatic verification, traditional selection strategies like majority voting and reward model scoring tend to plateau after a few hundred samples. The implementation strategy involves generating a high number of independent responses with elevated sampling temperatures for diversity, followed by selecting the best candidate based on established criteria. 

Huggingface team \cite{huggingface2023scaling} investigated dynamic inference strategies such as repeated sampling methods that allow models to “think longer” on challenging tasks like complex math problems. They systematically evaluate several search-based techniques including majority voting (self-consistency), Best-of-N sampling, beam search augmented with process reward models (PRMs), and a novel Diverse Verifier Tree Search (DVTS) designed to enhance candidate diversity under high compute budgets. Their experimental pipeline involves generating multiple candidate solutions, scoring intermediate reasoning steps with a PRM, and selecting optimal outputs based on these scores. Remarkably, the study demonstrates that smaller models (e.g., Llama 3.2 1B) can, with sufficient test-time compute, rival or even surpass the performance of larger counterparts (such as Llama 3.1 70B) on the MATH-500 benchmark, achieving up to 55\% accuracy. The findings also reveal that the effectiveness of each search strategy varies with problem difficulty beam search tends to excel on harder problems while Best-of-N sampling is better suited for simpler ones and that an adaptive, compute-optimal scaling approach can yield significant performance gains.

\subsection{SageAttention} 

SageAttention \cite{zhang2024sageattention} is a novel quantization method specifically designed to accelerate the attention mechanism in transformers a critical component that traditionally incurs an O(N²) computational cost with long sequence lengths. While existing quantization techniques predominantly optimize the linear layers, SageAttention directly targets attention by employing a 4/8-bit quantization for the Q and K matrices and FP8/16 for the P and V matrices, complemented by smoothing methods for Q and V. It dynamically adjusts quantization parameters across timesteps and layers, serving as a drop-in replacement for PyTorch’s scaled\_dot\_product\_attention. Experimental results are compelling: SageAttention achieves approximately a 3× speedup over FlashAttention2, outperforming FlashAttention2 and xformers by about 2.1× and 2.7× in operations per second, respectively, while maintaining roughly 99\% of the original model performance. Moreover, extensive evaluations demonstrate that this approach results in negligible end-to-end performance loss across a range of applications, including large language processing, image generation, and video generation.

The standard self-attention is computed as follows \cite{vaswani2017attention}:

\begin{align*}
S &= \frac{Q\,K^\top}{\sqrt{d}},
\end{align*}
where:
\begin{itemize}
    \item \(Q \in \mathbb{R}^{T \times d}\) is the query matrix,
    \item \(K \in \mathbb{R}^{T \times d}\) is the key matrix,
    \item \(d\) is the model's hidden dimension, and
    \item \(T\) is the sequence length.
\end{itemize}

The scaling factor \(\sqrt{d}\) prevents the dot products from growing too large, which would saturate the softmax function. The attention scores \(S\) are normalized using the softmax function:

\begin{align*}
P &= \mathrm{softmax}(S),
\end{align*}
where the softmax is applied row-wise so that each row of \(P \in \mathbb{R}^{T \times T}\)
sums to 1.

The final output is then:
\begin{align*}
O &= P\,V,
\end{align*}

with \(V \in \mathbb{R}^{T \times d}\) being the value matrix. Thus, each output token in \(O\) is a weighted sum of the value vectors.

FlashAttention improves efficiency by processing the attention computation in blocks. Assume that we partition the key and value matrices into blocks \(\{K_i, V_i\}\), where each block corresponds to a subset of tokens.

For a given block \(i\) and token index \(j\) within that block, we first compute the partial
logits:
\begin{align*}
l_{j}^{(i)} &= \exp\Bigl( S_{j}^{(i)} - m_{j}^{(i)} \Bigr),
\end{align*}
where:
\begin{itemize}
    \item \(S_{j}^{(i)}\) denotes the attention score for token \(j\) in block \(i\),
    \item \(m_{j}^{(i)} = \max_{k \in \text{block}} S_{k}^{(i)}\) is the maximum score in the block,
    ensuring numerical stability.
\end{itemize}

Next, the partial softmax probability for token \(j\) in block \(i\) is defined as:
\begin{align*}
p_{j}^{(i)} &= \frac{l_{j}^{(i)}}{\sum_{k \in \text{block}} l_{k}^{(i)}},
\end{align*}
which normalizes the logits within the block.

Finally, the output for token \(j\) from block \(i\) is computed as:
\begin{align*}
O_{j}^{(i)} &= p_{j}^{(i)}\,V_i,
\end{align*}
where \(V_i\) is the value matrix corresponding to block \(i\). The final output \(O\) is
obtained by aggregating the outputs \(O_{j}^{(i)}\) from all blocks. This blockwise approach avoids the need to compute and store the full attention matrix
\(P\) of size \(T \times T\), thereby reducing both memory usage and I/O overhead, which is
especially beneficial for long sequences.

\subsection{Speculative Knowledge Distillation (SKD)}

Google introduced Speculative Knowledge Distillation (SKD) \cite{xu2024speculative}, a novel approach designed to overcome the inherent limitations of conventional knowledge distillation methods namely, supervised KD and on-policy KD in bridging the gap between teacher and student models. Traditional supervised KD suffers from a distribution mismatch between static training datasets and the dynamic outputs produced during inference, while on-policy KD often relies on lower-quality, student-generated samples that can yield inaccurate teacher feedback. In contrast, SKD leverages an interleaved sampling strategy in which the student model initially generates a draft sequence of tokens, which are then evaluated against the teacher model's top-K predictions. If a student-proposed token ranks within the top 25 of the teacher’s probability distribution, it is retained; otherwise, it is replaced by the teacher’s highest probability token. This adaptive token replacement mechanism allows SKD to generate high-quality training data on-the-fly that is better aligned with the student's inference-time distribution. Evaluated across diverse text generation tasks including translation, summarization, mathematical problem-solving, and instruction following SKD consistently outperforms supervised and on-policy KD methods, enhancing speculative decoding performance in the distilled model.

\subsection{Self-Taught Reasoner (STaR)}

Zelikman et al. \cite{zelikman2022star} proposed the Self-Taught Reasoner (STaR), an innovative iterative training technique that enables language models to improve their reasoning capabilities by learning from their own generated rationales. Traditional chain-of-thought prompting requires either extensive datasets of annotated rationales or relies on few-shot inference, which can compromise accuracy. STaR circumvents these limitations through a simple yet effective loop: using a small set of few-shot examples with rationales, the model generates step-by-step explanations (rationales) and answers for a large collection of problems. When the generated answer is incorrect, the model is prompted again this time with a hint that includes the correct answer to produce a revised rationale. The model is then fine-tuned on both the initially correct rationales and these corrected examples. Repeating this process over multiple iterations (typically 30–40) enables the model to refine its reasoning skills progressively. Experimental results show that STaR significantly improves performance on various reasoning tasks, such as GSM8K and CommonsenseQA, achieving gains comparable to those from fine-tuning a model 30 times larger. Notably, on GPT-J, STaR boosts performance from 5.8\% to 10.7\%, demonstrating its potential to reduce the reliance on large, manually labeled rationale datasets while effectively enabling models to learn from their mistakes.

V-STaR \cite{hosseini2024v} represents a significant advancement in self-improvement for large language models by harnessing the overlooked value of incorrect solutions. Traditional approaches like STaR focus solely on fine-tuning models using correct self-generated solutions, discarding a wealth of information present in the erroneous outputs. In contrast, V-STaR employs a novel strategy where both correct and incorrect solutions are utilized to form preference pairs, which are then used to train a DPO-based verifier. This verifier evaluates and ranks candidate solutions during inference, selecting the most promising output. By iteratively refining both the generator and the verifier, V-STaR achieves an impressive 4\% to 17\% improvement in test accuracy on common code generation and math reasoning benchmarks with LLaMA2 models, outperforming conventional self-consistency and majority voting approaches, particularly for smaller candidate sets.

Quiet-STaR \cite{zelikman2024quiet} is a novel generalization of the Self-Taught Reasoner (STaR) approach that teaches language models to "think before they speak" by generating token-level rationales essentially internal thoughts alongside each token prediction. Instead of being limited to generating final answers, Quiet-STaR trains models to output parallel sequences of "thoughts" (delimited by learnable $<|startofthought|>$ and $<|endofthought|>$ tokens) for each token using a tokenwise parallel sampling algorithm with a specially designed attention mask. A mixing head then integrates these thought outputs to inform next-token predictions, while REINFORCE is used to optimize the generated thoughts based on how much they improve the log-likelihood of future tokens. Iteratively training on large diverse corpora (e.g., OpenWebMath, C4) leads to significant zero-shot performance gains on reasoning-heavy tasks improving GSM8K scores from 5.9\% to 10.9\% and CommonsenseQA from 36.3\% to 47.2\% for a Mistral 7B model demonstrating that embedding reflective reasoning into every token can help models more effectively predict complex, unstated logic without additional fine-tuning.

\subsection{Self-Taught Evaluators}

Meta team \cite{wang2024selftaughtevaluators} proposes a novel iterative self-improvement framework for model-based evaluation that entirely forgoes the need for human-labeled preference data. Traditionally, training evaluators whether for reward modeling in reinforcement learning or as stand-ins for human evaluators relies on extensive collections of human judgments, a process that is both costly and susceptible to obsolescence as models evolve. Instead, the authors introduce a self-taught evaluator approach where an LLM is trained using synthetic data generated from unlabeled instructions. In each iteration, the model produces contrasting outputs one high-quality and one intentionally sub-optimal followed by generating detailed reasoning traces and final judgments for these response pairs. These synthetic preference pairs are then used to iteratively fine-tune the evaluator, with each cycle leveraging the improved judgments from the previous iteration. Empirical results on RewardBench demonstrate a marked improvement: the evaluator based on Llama3-70B-Instruct sees its performance increase from 75.4\% to 88.3\% (or 88.7\% with majority voting), matching the effectiveness of top reward models trained on human annotations. This synthetic, iterative strategy not only reduces reliance on costly human data but also allows for customizable evaluation criteria, though it also highlights potential challenges such as the risk of amplifying initial model biases over successive iterations.

\subsection{Self-Enhanced Test-Time Scaling (SETS)}

Google team presents Self-Enhanced Test-Time Scaling (SETS) \cite{chen2025setsn}, a novel framework that leverages the self-verification and self-correction capabilities of advanced LLMs to overcome the diminishing returns observed in conventional test-time scaling methods. Traditional approaches, such as repeated sampling with majority voting or employing task-specific reward models, have struggled to yield proportional performance gains as computational resources increase. In contrast, SETS integrates a three-step iterative loop generating multiple solution candidates, self-verifying these outputs using the model’s inherent reasoning abilities, and then applying targeted self-correction to refine erroneous responses into a unified process. By incorporating majority voting over these refined outputs, the framework enhances accuracy and demonstrates more favorable scaling characteristics. Experimental evaluations on challenging planning and reasoning benchmarks indicate that SETS can achieve up to an 8.7\% improvement in accuracy while reducing compute requirements by 22\% compared to traditional sampling-heavy baselines. This approach underscores a promising direction for exploiting the emergent self-refinement abilities of LLMs, thereby pushing the boundaries of test-time computation in complex reasoning tasks.

\subsection{Stream of Search (SoS)}

Gandhi et al. \cite{gandhi2024stream} introduced the "Stream of Search" (SoS) approach, a training methodology designed to teach language models to perform structured search and backtracking by representing the entire search process as a continuous sequence a flattened string of actions. Unlike conventional training, which exposes models only to optimal reasoning trajectories, SoS incorporates all intermediate steps, including mistakes and corrective backtracking, thereby mimicking human problem-solving behavior. The method is demonstrated on the challenging Countdown game, where a set of input numbers must be combined through arithmetic operations to reach a target number. Using a heuristic problem solver, the authors generated 500,000 chain-of-thought/search examples; notably, only 57\% of these samples led to correct solutions, highlighting the inherent complexity of the search process. Training a transformer-based language model on this rich, diverse dataset improved search accuracy by 25\% compared to models trained solely on optimal paths. Furthermore, by applying policy improvement techniques such as Self-Taught Reasoner (STaR) and Advantage-Induced Policy Alignment (APA), the fine-tuned models achieved an additional ~6\% gain, ultimately solving 36\% of previously unsolvable problems. Overall, the study demonstrates that exposing language models to the full spectrum of search behaviors including errors enables them to learn more robustly and self-correct, thereby improving their performance on complex reasoning tasks.

\subsection{Reflective Augmentation (RefAug)}

Zhang et al. \cite{zhang2024learn} propose Reflective Augmentation (RefAug), a novel technique designed to embed problem reflection into training data for language models, enhancing their problem-solving abilities on mathematical and coding tasks. Unlike conventional data augmentation that merely increases training volume, RefAug integrates alternative reasoning approaches, analogies, and diverse perspectives alongside each problem and its solution, thereby training models to engage in both forward and reflective reasoning. Extensive experiments demonstrate that this method improves accuracy in math tasks by 6.8 points, boosts code performance by 3.5 points in Pass@1, and achieves an error correction rate of 81. 11\%, outperforming traditional chain-of-threat methods.

\begin{table*}[htbp]
\centering
\scriptsize
\caption{Challenges in Enhancing LLM Reasoning}
\label{tab:challenges}
\begin{adjustbox}{max width=\textwidth}
\begin{tabularx}{\textwidth}{@{}%
>{\raggedright\arraybackslash}p{3.0cm} % Topic
>{\raggedright\arraybackslash}p{3.5cm} % Core Issue
>{\raggedright\arraybackslash}p{3.0cm} % Proposed Approach
>{\raggedright\arraybackslash}p{3.0cm} % Observed Outcomes
>{\raggedright\arraybackslash}X@{}} % Overall Significance
\toprule
\textbf{Topic} & \textbf{Core Issue} & \textbf{Proposed Approach} & \textbf{Observed Outcomes} & \textbf{Overall Significance} \\
\midrule
Improving Reasoning Without Human Supervision & Enhancing multi-step reasoning without human annotations. & OmegaPRM \cite{luo2024improve} – a divide-and-conquer Monte Carlo Tree Search that autonomously collects process supervision data. & Boosts Gemini Pro's success on MATH500 from 51\% to 69.4\% and on GSM8K from 86.4\% to 93.6\%. & Demonstrates a scalable, cost-effective method for self-supervised reasoning improvement. \\
\midrule
Synthetic Data for Math Olympiad & LLMs show a reasoning gap in chained, multi-hop math problems. & ToRA \cite{gou2023tora} and MuMath-Code \cite{yin2024mumath} combine synthetic data generation with tool integration. & ToRA-7B scores 44.6\% on MATH; MuMath-Code-70B achieves 90.7\% on GSM8K and 55.1\% on MATH. & Highlights the potential of data augmentation and external tool use to bridge reasoning gaps. \\
\midrule
Criticality of Online Data for RLHF & Offline datasets often lack diversity for optimal RLHF convergence. & Comparing online RL methods (e.g., PPO) and introducing Hybrid Preference Optimization (HyPO) \cite{song2024importance}. & Online approaches require only partial coverage and outperform offline methods. & Emphasizes that on-policy data is crucial for effective model alignment. \\
\midrule
Multilingual RLHF Transfer & RLHF techniques predominantly focus on English, limiting performance in other languages. & Generating synthetic multilingual feedback and mixing language data during training \cite{dang2024rlhf}. & Training on English-only data yields up to 7\% improvements in other languages; multi-language data boosts win rates by up to 19\%. & Demonstrates significant cross-lingual transfer, enhancing global model alignment. \\
\midrule
DPO for LLM as a Judge & Standard reward models struggle with robust auto-evaluation. & SFR-Judges \cite{wang2024direct} integrate Direct Preference Optimization (DPO) with supervised fine-tuning loss using chain-of-thought critiques. & Achieves ~44\% performance on Alpaca Eval and ~81\% pairwise comparison accuracy. & Improves generative judge capabilities and downstream model evaluation. \\
\midrule
SFT Memorization vs. RL Generalization & SFT tends to overfit, hindering out-of-domain generalization. & Reinforcement Fine-Tuning (RFT) \cite{chu2025sftmemorizesrlgeneralizes} applied on top of an SFT baseline. & Yields up to 61\% improvement on out-of-domain tasks and +34\% on visual navigation tasks. & Underlines the need to combine SFT with RL for robust model adaptation. \\
\midrule
Impact of Structured Prompting & Rigid formats (e.g., JSON, XML) may restrict reasoning ability. & Comparing constraint decoding, format-restricting instructions, and two-step approaches \cite{tam2024let}. & Mixed outcomes: some models perform optimally with JSON; others see performance drops. & Highlights the trade-off between output consistency and cognitive flexibility. \\
\midrule
Effectiveness of Long-Context RAG & LLMs degrade when processing contexts beyond optimal lengths. & Systematic evaluation of retrieval thresholds and failure modes \cite{leng2024long,laban2024summary}. & Optimal performance at ~32k–64k tokens, with degradation and issues (e.g., repetition) beyond. & Suggests a need for improved long-context training and retrieval strategies. \\
\midrule
Role of the Reference Model in DPO & The strength of the reference model affects DPO alignment quality. & Experimenting with varied KL constraints and using stronger, compatible reference models \cite{liu2024understanding}. & Performance increases from 12.36 to 20.25 with optimal settings. & Underlines that a well-chosen reference model is vital for DPO success. \\
\midrule
Online vs. Offline Alignment & Offline methods excel in classification but lag in generative tasks. & Empirical comparison of online RLHF (e.g., PPO, RLOO) versus offline contrastive approaches \cite{tang2024understanding}. & Online methods achieve higher win rates, despite offline models performing better in pairwise tasks. & Highlights the value of on-policy sampling for generative quality and the trade-offs in alignment. \\
\midrule
Leveraging LLMs for Code Compilation & LLMs have untapped potential for code compilation and optimization tasks. & LLM Compiler \cite{cummins2024meta}: Fine-tuning Code Llama on large corpora of LLVM IR and assembly code. & Achieves 77\% of autotuning potential, 45\% disassembly accuracy, and 14\% exact match. & Offers a scalable, cost-effective foundation for compiler research and optimization. \\
\midrule
Training MoE Architectures & Efficiently training large-scale MoE models poses communication challenges. & Leveraging advanced communication libraries (e.g., DeepSpeed, MegatronLM) for MoE \cite{deepspeed-moe,megatron-lm,llm-foundry,deepep}. & Enables development of DeepSeek-R1 with efficient training (approx. \$6M cost) and optimized inference. & Advances efficient MoE parallelism, crucial for scaling large language models. \\
\bottomrule
\end{tabularx}
\end{adjustbox}
\end{table*}

\section{Challenges and Open problems}\label{sec:7}

This section presents a discussion of key challenges in advancing LLM capabilities. It highlights the potential of autonomous methods to enhance multi-step reasoning without human intervention while also noting the persistent difficulties in solving chained, multi-hop problems. It discusses the benefits and limitations of reinforcement learning from human feedback, particularly in multilingual settings, and examines the trade-offs between structured prompting and maintaining reasoning flexibility. In addition, it addresses the challenges associated with long-context retrieval and the opportunities in leveraging LLMs for code compilation and efficient Mixture-of-Experts architectures. Table \ref{tab:challenges} presents the challenges in enhancing LLM reasoning.

\subsection{Can we improve Reasoning without human supervision?}

Recent advances by Google DeepMind \cite{luo2024improve} demonstrate that autonomous approaches can effectively enhance multi-step reasoning in large language models. This work introduces OmegaPRM, a novel divide-and-conquer Monte Carlo Tree Search algorithm that autonomously collects high-quality process supervision data to train Process Reward Models (PRMs) and significantly boost reasoning performance. OmegaPRM employs a binary search strategy to identify the first error in a chain-of-thought, isolating mistakes and balancing positive and negative examples, which enables the generation of over 1.5 million supervision annotations without any human intervention. These annotations are used to train a PRM with a pointwise soft label objective, and when combined with weighted self-consistency, the method dramatically improves performance boosting the instruction-tuned Gemini Pro model's success rate on MATH500 from 51\% to 69.4\% and on GSM8K from 86.4\% to 93.6\%, while similarly enhancing Gemma2 27B's outcomes. Overall, OmegaPRM offers a cost-effective, scalable solution that reduces computational overhead and refines LLM reasoning by rewarding intermediate progress, all achieved without direct human supervision.

\textcolor{black}{This emerging paradigm opens new research directions for optimizing automated supervision signals. One promising avenue is integrating uncertainty estimation or probabilistic error localization into the reward signal design. Furthermore, combining self-supervised distillation with programmatic supervision, such as that derived from automated theorem provers or formal verifiers, could reduce reliance on manually labeled reasoning trajectories. Investigating how these methods generalize across domains and reasoning tasks (e.g., symbolic, scientific, legal) remains a key challenge for future work.}

\subsection{Will Synthetic Data Win the AI Math Olympiad?}

The study by Hosseini et al. \cite{hosseini2024not} delves into the grade-school math problem-solving abilities of LLMs by examining their performance on paired problems where the solution to the first is essential for resolving the second to uncover what the authors term a "reasoning gap." While many LLMs perform adequately on isolated math questions, their accuracy drops significantly when tasked with chained, multi-hop problems, revealing that the contextual overload and additional reasoning steps introduce challenges not apparent in standard benchmarks. Notably, this gap is more evident in smaller, cost-efficient, and math-specialized models, suggesting that even targeted fine-tuning can lead to overfitting and hinder generalization to composite tasks. The paper also highlights that improvements from instruction-tuning and code generation vary with model size, and that the reasoning shortfall is attributable not to data leakage but to inherent distractions and insufficient second-hop reasoning capabilities. Overall, the findings prompt a reevaluation of how LLMs are assessed for complex, multi-step problem solving, urging the development of more nuanced benchmarks that capture true reasoning proficiency.

Gou et al. \cite{gou2023tora} presents ToRA, a series of Tool-integrated Reasoning Agents that enhance large language models' ability to tackle complex mathematical problems by integrating natural language reasoning with external computational tools such as symbolic solvers and computation libraries. Recognizing that traditional LLMs struggle with intricate mathematical tasks due to limitations in pure language processing, the authors propose a novel training regimen that involves curating interactive tool-use trajectories from mathematical datasets, leveraging imitation learning on annotated data, and applying output space shaping to fine-tune reasoning behavior. Empirical evaluations demonstrate that ToRA models outperform existing open-source alternatives on ten mathematical reasoning benchmarks, yielding average absolute improvements between 13\% and 19\%. Notably, the ToRA-7B model achieves a score of 44.6\% on the challenging MATH dataset, surpassing the previous best open-source model by 22 percentage points, while ToRA-Code-34B becomes the first open-source model to exceed 50\% accuracy on MATH outperforming GPT-4’s chain-of-thought approach and rivaling its program-based problem solving.

Yin et al. \cite{yin2024mumath} proposes a novel approach that bridges two previously distinct research directions in mathematical reasoning for LLMs: external tool integration and data augmentation. Traditionally, tool-use LLMs enhance performance by interfacing with Python interpreters, while tool-free methods rely on enriching math datasets. To harness the benefits of both, the authors introduce MuMath-Code, which fine-tunes Llama-2 on an augmented dataset generated via multi-perspective data augmentation and synthesized code-nested solutions. The model employs a two-stage training strategy: Stage-1 involves fine-tuning on chain-of-thought data to establish robust reasoning patterns, and Stage-2 further refines the model on the newly augmented code-nested data. At inference, MuMath-Code generates executable code that interacts with an external Python interpreter to produce precise computation results. The approach is validated through extensive experiments, with the MuMath-Code-7B model achieving scores of 83.8 on GSM8K and 52.4 on MATH, and the MuMath-Code-70B model setting new state-of-the-art performance among open-source methods 90.7\% on GSM8K and 55.1\% on MATH. These results underscore the effectiveness of combining tool use with data augmentation and highlight a promising direction for future research in enhancing mathematical reasoning in LLMs.

\textcolor{black}{These findings motivate several open research directions. One important avenue involves building curriculum-based synthetic datasets that gradually increase reasoning complexity to improve generalization across math tasks. Another promising line of work is the design of adaptive tool-augmented agents capable of deciding when and how to invoke external solvers during problem solving. Further research could also investigate the cross-domain transferability of these synthetic math reasoning skills to adjacent fields such as symbolic logic, formal verification, or algorithmic coding tasks.}

\subsection{How critical is online data for RLHF?}

Song et al. \cite{song2024importance} presents a rigorous theoretical and empirical comparison between online reinforcement learning (RL) methods and offline contrastive techniques for fine-tuning large language models using human preference data. By introducing the notion of dataset coverage a measure of how well the training data represents the test distribution the authors establish that while offline methods like Direct Preference Optimization (DPO) require a stringent global coverage condition for optimal convergence, online RL methods, such as Proximal Policy Optimization (PPO), only need a weaker partial coverage condition. This theoretical distinction helps explain why online methods often outperform their offline counterparts, particularly in scenarios where the offline dataset lacks sufficient diversity. Motivated by these insights, the paper introduces Hybrid Preference Optimization (HyPO), a novel algorithm that leverages offline contrastive optimization alongside online data for KL regularization. The hybrid approach performs better than pure offline methods and retains computational and memory efficiency. I find PPO most compelling among the online RLHF methods discussed PPO, RLOO, Online-DPO, ReMax, and J-BOND given its robust theoretical foundation and proven success in dynamically generating and integrating new data during the training process.

\textcolor{black}{This line of work opens several promising research directions. One direction involves the design of adaptive data collection strategies that prioritize coverage improvement during online training. Additionally, future work could explore principled hybrid architectures that dynamically switch between offline and online regimes based on distributional drift or model uncertainty. Another area of interest is developing new preference modeling paradigms that extend beyond binary or scalar reward signals to support structured or multi-dimensional human feedback. Finally, benchmarking these approaches across diverse instruction-following and dialog tasks can help determine their real-world robustness and generalization.}

\subsection{Does RLHF transfer to different languages?}

Dang et al. \cite{dang2024rlhf} present a comprehensive study on extending preference optimization techniques for LLMs to a multilingual setting, addressing the gap in current research that predominantly focuses on English and Chinese. The authors propose a novel and scalable method to generate high-quality multilingual feedback data, which balances data coverage and facilitates cross-lingual transfer during training. By creating synthetic multilingual preference datasets using approximately 50K English prompts translated into 22 languages and designing various dataset mixtures, the study rigorously evaluates both offline (DPO) and online (RLOO) RLHF methods. Key findings reveal that even training on English-only data can yield up to 7\% improvements in other languages, while incorporating multiple languages can boost win rates by as much as 19\% on unseen languages. Notably, online RLHF (RLOO) consistently outperforms its offline counterpart by up to 10.6\%, demonstrating superior transfer capabilities. The resulting preference-trained model achieves a 54.4\% win-rate against the current state-of-the-art Aya 23 8B and outperforms several widely used models, thereby significantly advancing multilingual LLM alignment across 23 languages and impacting nearly half of the world's population.

\textcolor{black}{These results highlight a promising frontier for cross-lingual alignment via preference learning. Future research could explore more granular multilingual signal integration, such as culturally adaptive reward models or region-specific annotation schemes. Another direction involves dynamically adjusting language mixing ratios during training based on feedback quality or model confidence. In addition, aligning multilingual LLMs using code-switching inputs or zero-shot preference extrapolation may offer insights into generalized policy learning across languages and dialects. Finally, evaluating multilingual alignment performance across low-resource and typologically diverse languages remains a crucial step toward equitable LLM development.}

\subsection{Can DPO improve LLM as a Judge?}

Wang et al. \cite{wang2024direct} explore enhancing the auto-evaluation capabilities of LLMs by training them as generative judges using both positive and negative data through preference optimization. The authors introduce a novel framework SFR-Judges that leverages three distinct data collection methods, including chain-of-thought critiques, classification-based judgments, and response deduction, to generate comprehensive preference pairs across various evaluation tasks (single rating, pairwise comparison, and classification). By integrating Direct Preference Optimization (DPO) with supervised fine-tuning (SFT) loss, the generative judge model robustly outperforms strong baselines such as GPT-4o and specialized judge models, achieving superior performance on 10 out of 13 benchmarks. Notably, the approach mitigates common biases like position and length bias, adapts flexibly to diverse evaluation protocols, and provides insightful language feedback for improving downstream generative models. Empirical results further demonstrate that models evaluated with SFR-Judges not only reach a 44\%  performance on the Alpaca Eval but also achieve an average pairwise comparison accuracy of around 81\%, underscoring this method's promise in advancing LLM evaluation strategies.

\textcolor{black}{This line of research opens up important avenues for building more interpretable and unbiased evaluation frameworks. One promising direction involves extending judge models to support multi-turn or multi-agent conversations, enabling richer discourse-level assessments. Another is the incorporation of uncertainty quantification or calibration techniques to improve judge confidence and reliability. Moreover, adapting DPO-based judge models for multilingual or domain-specific evaluation tasks such as scientific reasoning or legal analysis could enhance their practical applicability. Finally, integrating generative judge outputs into closed-loop reinforcement learning setups may enable continuous model self-improvement through informed feedback.}

\subsection{SFT Memorizes and RL Generalizes}

Traditional supervised fine-tuning (SFT) has long been the default approach for specialized model tuning, but its limitations are becoming increasingly apparent. In a groundbreaking study by researchers from DeepMind, UC Berkeley, NYU, and HKU \cite{chu2025sftmemorizesrlgeneralizes}, it was demonstrated that SFT often traps models in a cycle of memorization, significantly degrading performance on out-of-domain tasks by as much as 79\% in some cases. This creates a major challenge: how can models be tuned to not only recall training data but also to adapt, reason, and generalize to varied, real-world scenarios, such as those encountered in enterprise environments?

Enter Reinforcement Fine-Tuning (RFT). Unlike SFT, RFT leverages outcome-based reward signals and reinforcement learning techniques (using PPO, for example) to encourage models to learn from their interactions with evaluation environments. In the study \cite{chu2025sftmemorizesrlgeneralizes}, a duplicate Llama 3.2 11B base was first fine-tuned with supervised methods and then refined via reinforcement learning. When evaluated on both text-based (GeneralPoints, an arithmetic reasoning card game) and visual navigation (V-IRL) tasks, RFT demonstrated dramatic improvements boosting performance on out-of-domain tasks by up to 61\% and delivering a +34\% improvement on the V-IRL task over previous state-of-the-art models.

However, the challenge is not entirely resolved. While RFT enhances generalization and adaptive reasoning, SFT still plays a crucial role by providing the structural stability necessary for effective secondary reinforcement learning. In essence, the study reveals that to unlock the potential of large language models truly, one must combine the foundational stability of SFT with the adaptive, generalizing power of RFT. This integrated approach represents a pivotal step toward building models that can excel across diverse, dynamic tasks, and it sets a new benchmark for the future of model tuning.

\textcolor{black}{Future work should explore curriculum-aware reinforcement fine-tuning that gradually shifts from memorization to generalization. Additionally, incorporating meta-learning principles into RFT could allow models to adapt to unseen domains with limited supervision quickly. Another promising direction involves developing diagnostic benchmarks to explicitly measure the trade-offs between memorization and generalization. Finally, expanding RFT to encompass multimodal environments (e.g., vision-language, embodied agents) will be crucial for evaluating its full potential in real-world reasoning settings.}

\subsection{Does structure prompting impact reasoning? }

Tam et al. \cite{tam2024let} explores a significant challenge in building real-world LLM applications: balancing structured prompting with the preservation of reasoning abilities. Researchers investigate whether imposing formats such as JSON, XML, or YAML common in many structured prompts might undermine an AI's capacity to reason effectively. The study reveals mixed outcomes by comparing various methods, including constraint decoding, format-restricting instructions, and a two-step natural language-to-format approach across diverse datasets and models. While structured formats can enhance consistency in some cases, they often lead to a decline in reasoning performance. For instance, although specific models perform optimally with JSON, others show notable drops in performance when forced into a rigid schema. This nuanced analysis underscores the complex trade-off between ensuring output uniformity and maintaining the cognitive flexibility required for advanced reasoning tasks.

\textcolor{black}{Looking ahead, future research could explore adaptive prompting mechanisms that dynamically adjust structural constraints based on task complexity or reasoning depth. Another direction involves training format-aware models that can decouple syntactic conformity from semantic reasoning, possibly through multi-task objectives. Moreover, fine-tuning or reinforcement learning strategies that explicitly reward both structural correctness and reasoning quality could help bridge the gap. Developing benchmarks that isolate format-induced reasoning degradation will also be crucial for better diagnostic insights.}

\subsection{How effective are LLMs at long-context RAG?}

Databricks Mosaic Research \cite{leng2024long} systematically investigates the performance of LLMs in long-context Retrieval Augmented Generation (RAG) settings through over 2,000 experiments on 13 open and closed models across four curated datasets. The study reveals that while retrieving additional documents generally enhances RAG performance, most models begin to degrade beyond certain context thresholds around 32k tokens for models like Llama-3.1-405B and up to 64k tokens for GPT-4. Moreover, different models exhibit distinct failure modes under long-context conditions: for example, Claude 3.5 shows a sharp increase in copyright-related failures from 3.7\% at 16k to 49.5\% at 64k, and DBRX's ability to follow instructions deteriorates from 5.2\% at 8k to 50.4\% at 32k. Some models, such as Mixtral, even generate repeated content, and many suffer from a "lost in the middle" effect, where information in the central portions of long texts is not effectively utilized. These findings suggest that the optimal context length is highly model- and task-dependent, and that the lack of long-context post-training may be a key factor in the observed performance limitations.

Laban et al. \cite{laban2024summary} introduces the "Summary of a Haystack" (SummHay) benchmark, a novel framework designed to evaluate long-context understanding in LLMs and retrieval-augmented generation (RAG) systems. As modern LLMs can handle millions of tokens, traditional tasks like Needle-in-a-Haystack fail to capture the complexity of extended information extraction. In response, the authors synthesize "haystacks" of documents where key insights are deliberately repeated, challenging models to generate summaries that not only capture these insights (Coverage) but also accurately cite the corresponding source documents (Citation). The evaluation protocol is highly reproducible, automatically scoring summaries on these two critical dimensions. A comprehensive analysis across 10 LLMs and 50 RAG systems reveals that even with Oracle signals for document relevance, current systems lag human performance (56\%) by more than 10 points on a joint score. Notably, models such as Google DeepMind Gemini 1.5 pro outperform others with and without RAG while long-context LLMs like GPT-4o and Claude 3 Opus score below 20\% when not aided by a retriever. Furthermore, smaller models integrated with RAG, such as Claude 3 Haiku or Gemini 1.5 Flash, sometimes outperform larger LLMs, underscoring the importance of effective retrieval strategies in handling long-context tasks. Overall, SummHay provides a challenging yet precise benchmark to drive future improvements in long-context information extraction and enterprise RAG systems.

\textcolor{black}{Future research could explore hybrid training approaches incorporating long-context conditioning during pre-training and fine-tuning stages. Developing retrieval-aware attention mechanisms or memory-efficient compression strategies could mitigate middle-content loss and repetition. Additionally, co-training retrievers with decoders using joint objectives may yield more coherent and well-grounded summaries. Finally, benchmarks like SummHay could be extended to include multilingual and multimodal contexts, further testing model generalization under real-world complexity.}

\subsection{How important is the reference model in DPO?}

Liu et al. \cite{liu2024understanding} investigates a critical yet underexplored aspect of Direct Preference Optimization (DPO) for instruction fine-tuning of large language models, focusing on the role of the reference model or policy. The authors examine three key research questions: the optimal strength of the KL-divergence constraint (controlled by $\beta$), the necessity of a reference model in DPO, and whether a stronger, compatible reference model can enhance performance. Their experiments reveal that reducing the KL-divergence constraint improves DPO performance up to an optimal point (with scores increasing from 12.36 to 16.25), beyond which performance degrades. Moreover, using a stronger reference model such as one that is already DPO-ed or aligned further boosts performance (from 16.25 to 20.25), whereas employing an incompatible reference model (e.g., Llama 3) yields suboptimal results. Overall, the study highlights the confounding role of the reference policy in DPO and provides valuable insights for best practices, as demonstrated by an improved Mistral 7B AlpacaEval2 score from 12.36 to 20.25.

\textcolor{black}{Building on these insights, future research could investigate adaptive or dynamic reference policies that evolve during training rather than remaining static. Another promising direction is exploring ensemble or interpolated reference models to stabilize preference gradients and reduce brittleness. Additionally, understanding how reference model compatibility impacts multilingual, multimodal, or tool-augmented tasks could further generalize DPO best practices. Finally, theoretical analysis of DPO convergence under varying reference policy conditions would help ground empirical findings in more rigorous guarantees.}

\subsection{Online and offline alignment algorithms}

Tang et al. \cite{tang2024understanding} provides an in-depth empirical comparison between online and offline reinforcement learning from human feedback (RLHF) methods for large language model alignment. While offline approaches rely on fixed datasets for preference tuning, online RLHF continuously generates new training data from the evolving policy, leading to a significant performance edge. Using Identity Preference Optimization (IPO) as a common framework, the authors demonstrate that online methods yield higher win rates, mainly due to more diverse and effective on-policy data coverage. Intriguingly, the study reveals a trade-off: offline-trained policies excel in pairwise classification but underperform in generative tasks, whereas online-trained models, despite being less adept at classification, generate superior outputs. This performance gap persists across both contrastive and non-contrastive loss functions and remains unmitigated by merely scaling up model size. Overall, the findings underscore the pivotal role of on-policy sampling in RLHF and highlight fundamental challenges in offline alignment algorithms, suggesting that iterative approaches to make offline datasets more on-policy could enhance performance.

\textcolor{black}{These insights suggest several avenues for future exploration. One promising direction involves hybrid alignment frameworks that combine offline pre-training with targeted online refinement cycles to bridge the generative-classification performance gap. Another area of interest is the development of proxy metrics or alignment validators to better guide on-policy sampling strategies. Additionally, advancing offline data augmentation methods, such as imitation distillation of on-policy behavior or synthetic preference mining, could help make static datasets more reflective of real-time interaction patterns. Finally, benchmarking alignment techniques under multi-agent, multilingual, or adversarial feedback settings could further elucidate the generalization limits of current alignment algorithms.}

\subsection{Leveraging LLMs for Code Compilation and Optimization}

Meta team \cite{cummins2024meta} introduces the Large Language Model Compiler (LLM Compiler), a robust suite of pre-trained models tailored for code and compiler optimization a domain that has been largely overlooked despite LLMs' impressive performance in software engineering. Built on the foundation of Code Llama, the LLM Compiler is available in 7B and 13B parameter sizes and has been trained on an extensive corpus of 546 billion tokens of LLVM intermediate representations (IRs) and assembly code from x86\_64, ARM, and CUDA. Moreover, through instruction fine-tuning on an additional 164 billion tokens, the models have learned to interpret compiler behavior effectively. Empirical evaluations reveal that the fine-tuned versions can achieve 77\% of the optimization potential of an autotuning search and attain a 45\% disassembly round trip accuracy with a 14\% exact match rate. Released under a bespoke commercial license, this work offers a scalable and cost-effective foundation for both academic research and industrial applications in compiler optimization.

\textcolor{black}{Future research could investigate integrating LLM-based compilers with traditional optimization pipelines to create hybrid systems that combine statistical reasoning with symbolic guarantees. Another promising direction involves training multilingual or cross-architecture compiler models that generalize across IRs and assembly code from various platforms. Additionally, fine-tuning LLM compilers using reinforcement learning with performance-based rewards could further push the boundaries of learned optimization. Finally, benchmarking compiler-oriented LLMs on real-world codebases and resource-constrained edge platforms would offer deeper insights into their practical viability and robustness.}

\subsection{Training MoE architecture}

Recent developments in LLMs have increasingly incorporated the Mixture-of-Experts (MoE) paradigm, which extends conventional transformer architectures by replicating feedforward blocks into multiple specialized "experts." A dynamic routing mechanism in this framework allocates input tokens to select experts, thereby significantly increasing model capacity (ranging from 20B to 600B parameters) without a proportional increase in per-token computational operations. This architectural modification theoretically enhances model intelligence while preserving latency, albeit at the cost of elevated memory usage. The inherent challenges of MoE, particularly the need for efficient multi-GPU parallelization and optimized inter-device communication, have been addressed by only a limited number of codebases, including DeepSpeed \cite{deepspeed-moe}, MegatronLM \cite{megatron-lm}, and MosaicML's LLM Foundry \cite{llm-foundry}. In this context, the recent open-source release from DeepSeek \cite{deepep}, which includes a state-of-the-art communication and orchestration library with advanced FP8 support, represents a significant contribution to the field. This release underpins the development of DeepSeek-R1 a state-of-the-art MoE model noted for its impressive performance, training efficiency (with an estimated training cost of \$6M), and optimized inference capabilities and provides valuable insights into efficient MoE parallelism. 

\textcolor{black}{Future work could explore intelligent expert routing strategies beyond token-level gating to consider task context, domain specialization, or user intent. Another research direction involves hierarchical or sparse MoE configurations to reduce memory and communication bottlenecks further. Co-optimizing model architecture and training pipelines through neural architecture search (NAS) or reinforcement learning may yield more scalable MoE designs. Finally, extending MoE capabilities to multilingual and multimodal inputs could unlock new frontiers in generalized reasoning and efficient LLM deployment.}

\section{Conclusion}
\label{sec:8}

Recent advances in reasoning LLMs, exemplified by models such as DeepSeek-R1, OpenAI’s o1 \& o3, and GPT-4o have underscored the value of intermediate multistep processing to solve complex problems ranging from advanced math and coding challenges to intricate puzzles. By generating intermediate reasoning steps, these systems provide transparency into their thought processes, leading to more accurate and reliable outcomes than direct-response language models.

To build and refine these reasoning models, researchers have explored various strategies, including inference-time scaling, pure reinforcement learning (as seen in DeepSeek-R1-Zero), combined supervised fine-tuning with reinforcement learning, and pure supervised fine-tuning with distillation. Models like Qwen-32B and Llama-based variants have also been adapted using these techniques, demonstrating that emergent reasoning capabilities can be effectively nurtured through a blend of reinforcement learning and fine-tuning. These approaches enhance performance and help mitigate issues such as overthinking and excessive verbosity during inference.

In this paper, we presented a comprehensive review of the advancements in large language models and their training methodologies that have emerged between 2023 and 2025. Our analysis spanned a diverse array of state-of-the-art LLMs. It provided an in-depth comparison of their performance across multiple benchmarks, offering valuable insights into their evolving capabilities. We examined innovative training strategies, including a mixture of experts, retrieval-augmented generation, and reinforcement learning, as well as architectural improvements that have significantly enhanced model performance.

Overall, our study highlights the remarkable progress achieved in the field and outlines the critical hurdles that future research must address. By shedding light on these aspects, we hope to inspire further innovations that will continue to push the boundaries of what large language models can achieve in both academic and real-world applications. The trend toward domain- and application-specific optimization is expected to intensify. As evidenced by models like DeepSeek-R1-Distill, Sky-T1, and TinyZero, the integration of these strategies promises to deliver specialized reasoning systems that balance high performance with computational efficiency.

\section*{Declaration of competing interest}
The authors declare that they have no known competing
financial interests or personal relationships that could have
appeared to influence the work reported in this paper.

\bibliographystyle{IEEEtran}
\bibliography{bibliography} 

\end{document}